%% file: main.tex
\documentclass[10pt,twocolumn,letterpaper]{article}
\usepackage[pagenumbers]{cvpr}

\input{tex/plots}

\usepackage[export]{adjustbox}

\input{tex/abbrev}
\input{tex/macro}
\usepackage[inkscapelatex=false]{svg}
\usepackage{colortbl}
\usepackage{multirow}
\usepackage{tcolorbox}
\usepackage{pifont}%

\definecolor{cvprblue}{rgb}{0.21,0.49,0.74}
\usepackage[pagebackref,breaklinks,colorlinks,allcolors=cvprblue]{hyperref}

\title{Advancing Semantic Future Prediction through\\ Multimodal Visual Sequence Transformers}

\author{Efstathios Karypidis$^{1,3}$ \hspace{0.5em}
Ioannis~Kakogeorgiou$^{1}$ \hspace{0.5em} Spyros~Gidaris$^{2}$ \hspace{0.5em}  Nikos~Komodakis$^{1,4,5}$ \vspace{0.5em} 
\\
$^1$Archimedes, Athena Research Center, Greece \hspace{1.0em} $^2$valeo.ai \\
$^3$National Technical University of Athens \hspace{1.0em} $^4$University of Crete \hspace{1.0em} $^5$IACM-Forth
}

\begin{document}

\input{sec/teaser}

\maketitle
\input{sec/0_Abstract}    
\input{sec/1_Introduction}
\input{sec/2_Related}
\input{sec/3_Methodology}
\input{sec/4_Experiments}
\input{sec/5_Conclusion}

\input{sec/acknowledgements}

{
    \small
    \bibliographystyle{ieeenat_fullname}
    \bibliography{main}
}

\clearpage
\appendix
\input{sec/supplementary}

\end{document}

%% file: tex/plots.tex
\usepackage{tikz}
\usetikzlibrary{arrows.meta,shapes,calc,matrix,fit,positioning,backgrounds,decorations.markings}
\usepackage{pgfplots}
\usepackage{pgfplotstable}
\pgfplotsset{compat=1.9}
\usepackage{xstring}

\usepgfplotslibrary{external}

\IfBeginWith*{\jobname}{fig/extern/}{\finalcopy}{}

\tikzstyle{every picture}+=[
	remember picture,
	every text node part/.style={align=center},
	every matrix/.append style={ampersand replacement=\&},
]
\tikzstyle{tight} = [inner sep=0pt,outer sep=0pt]
\tikzstyle{node}  = [draw,circle,tight,minimum size=12pt,anchor=center]
\tikzstyle{op}    = [draw,circle,tight]
\tikzstyle{dot}   = [fill,draw,circle,inner sep=1pt,outer sep=0]
\tikzstyle{pt}    = [fill,draw,circle,inner sep=1.5pt,outer sep=.2pt]
\tikzstyle{box}   = [draw,rectangle,inner sep=3pt]
\tikzstyle{high}  = [black!60]
\tikzstyle{group} = [high,box,opacity=.5]
\tikzstyle{dim1}  = [fill opacity=.3,text opacity=1]
\tikzstyle{dim2}  = [fill opacity=.5,text opacity=1]
\tikzstyle{dim3}  = [fill opacity=.7,text opacity=1]
\tikzstyle{rectc} = [tight,transform shape]
\tikzstyle{rect}  = [rectc,anchor=south west]

\tikzset{every mark/.append style={solid}}
\pgfplotsset{
	grid=both, width=\columnwidth, try min ticks=5,
	every axis/.append style={font=\small},
	every axis plot/.append style={thick,mark=none,mark size=1.8,tension=0.18},
	legend cell align=left, legend style={fill opacity=0.8},
	xticklabel={\pgfmathprintnumber[assume math mode=true]{\tick}},
	yticklabel={\pgfmathprintnumber[assume math mode=true]{\tick}},
	nodes near coords math/.style={
		nodes near coords={\pgfmathprintnumber[assume math mode=true]{\pgfplotspointmeta}},
	},
}

\pgfplotsset{
	dash/.style={mark=o,dashed,opacity=0.6},
	dott/.style={mark=o,dotted,opacity=0.6},
	nolim/.style={enlargelimits=false},
	plain/.style={every axis plot/.append style={},nolim,grid=none},
}

\pgfdeclarelayer{bg4}
\pgfdeclarelayer{bg3}
\pgfdeclarelayer{bg2}
\pgfdeclarelayer{bg1}
\pgfdeclarelayer{fg1}
\pgfdeclarelayer{fg2}
\pgfdeclarelayer{fg3}
\pgfdeclarelayer{fg4}
\pgfsetlayers{bg4,bg3,bg2,bg1,main,fg1,fg2,fg3,fg4}

\pgfkeys{/tikz/.cd, aspect/.store in=\aspect, aspect=1}
\pgfkeys{/tikz/.cd, depth/.store in=\depth, depth=.5}
\pgfkeys{/tikz/.cd, stepx/.store in=\step, stepx=1}

\tikzstyle{geom} = [line join=bevel,aspect=1,depth=.5,z={(\depth*\aspect,\depth)}]
\tikzstyle{wire} = [geom,draw,thick]

\def\cx[#1,#2,#3]{#1}
\def\cy[#1,#2,#3]{#2}
\def\cz[#1,#2,#3]{#3}
\def\ex[#1,#2,#3]{#1,0,0}
\def\ey[#1,#2,#3]{0,#2,0}
\def\ez[#1,#2,#3]{0,0,#3}



%% file: tex/abbrev.tex
\newcommand{\Th}[1]{\textsc{#1}}
\newcommand{\mr}[2]{\multirow{#1}{*}{#2}}
\newcommand{\mc}[2]{\multicolumn{#1}{c}{#2}}

\newcommand{\red}[1]{{\textcolor{red}{#1}}}

\newcommand{\citeme}[1]{\red{[XX]}}
\newcommand{\refme}[1]{\red{(XX)}}

\newcommand{\miou}{\Th{mIoU}\xspace}%

\makeatletter
\newcommand*\bdot{\mathpalette\bdot@{.7}}
\newcommand*\bdot@[2]{\mathbin{\vcenter{\hbox{\scalebox{#2}{$\m@th#1\bullet$}}}}}
\makeatother

\makeatletter
\DeclareRobustCommand\onedot{\futurelet\@let@token\@onedot}
\def\@onedot{\ifx\@let@token.\else.\null\fi\xspace}

\def\ie{\emph{i.e}\onedot}

\def\etal{\emph{et al}\onedot}
\makeatother

%% file: tex/macro.tex
\definecolor{DarkenedMagenta}{RGB}{225,0,167}

\newcommand{\ours}{\texttt{FUTURIST}\xspace}

\definecolor{ForestGreen}{RGB}{34,139,34}
\definecolor{Orange}{RGB}{1.0, 0.55, 0.0}
\definecolor{Purple}{RGB}{0.58, 0.44, 0.86}
\definecolor{VibrantMagenta}{RGB}{255,0,197}

\definecolor{TableColor}{rgb}{0.984, 0.922, 0.843}

\newcommand{\parag}[1]{\smallskip\noindent\textbf{#1}\enspace}

\newenvironment{customlegend}[1][]{%
    \begingroup
    \csname pgfplots@init@cleared@structures\endcsname
    \pgfplotsset{#1}%
}{%
    \csname pgfplots@createlegend\endcsname
    \endgroup
}%

\def\addlegendimage{\csname pgfplots@addlegendimage\endcsname}

\newcommand{\supplautoref}[1]{\hyperref[#1]{Suppl.~\autoref{#1}}}
\newcommand{\supplsec}[1]{\hyperref[#1]{Suppl. Sec.~\ref{#1}}}
\newcommand{\reffigures}[1]{\hyperref[#1]{Figures~\ref{#1}}}

%% file: sec/teaser.tex

\newcommand{\teaserfig}[1]{\includegraphics[trim={0cm 0cm 0cm 0cm},width=1.0\textwidth,valign=c]{#1}}

\makeatletter
\apptocmd\@maketitle{{\teaser{}}}{}{}
\makeatother

\newcommand{\teaser}{%
\vspace{-10pt}
\centering

\teaserfig{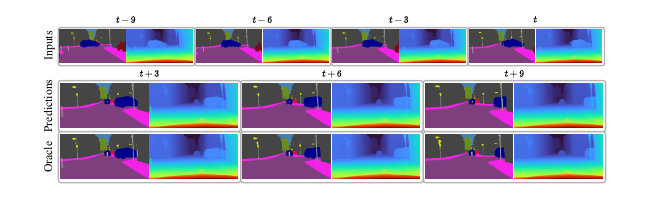}\\

\vspace{-6pt}
\captionof{figure}{Our framework predicts future semantic segmentation and depth maps using a multimodal transformer architecture. Leveraging masked visual modeling and cross-modal fusion, it excels in future semantic prediction, achieving state-of-the-art results in both tasks.}
\label{fig:teaser}

\par\vspace{10pt}
}

%% file: sec/0_Abstract.tex
\begin{abstract}
\vspace{-20pt}

Semantic future prediction is important for autonomous systems navigating dynamic environments. This paper introduces \ours, a method for multimodal future semantic prediction that uses a unified and efficient visual sequence transformer architecture. Our approach incorporates a multimodal masked visual modeling objective and a novel masking mechanism designed for multimodal training. This allows the model to effectively integrate visible information from various modalities, improving prediction accuracy. Additionally, we propose a VAE-free hierarchical tokenization process, which reduces computational complexity, streamlines the training pipeline, and enables end-to-end training with high-resolution, multimodal inputs. We validate \ours on the Cityscapes dataset, demonstrating state-of-the-art performance in future semantic segmentation for both short- and mid-term forecasting. Project page and code at  \href{https://futurist-cvpr2025.github.io/}{https://futurist-cvpr2025.github.io/}.

\end{abstract}

%% file: sec/1_Introduction.tex
\section{Introduction}
\label{sec:intro}
Future prediction is crucial for autonomous agents, enabling them to navigate complex, dynamic environments and make informed real-time decisions. By anticipating potential scenarios and future outcomes, these systems can effectively plan their actions, adapt to changing conditions, and ensure safe and efficient operation \cite{DosovitskiyK17}. 

Recent research has increasingly focused on predicting raw RGB values for future video frames \cite{srivastava2015unsupervised,mathieuCL15,gupta2023maskvit, yu2023magvit, gao2024vista,yang2024generalized, hu2023gaia,yu2024language,gupta2025photorealistic, ho2022video}. Although these methods can create realistic and diverse scenarios for simulation and testing, forecasting RGB frames remains highly challenging due to the data’s inherent complexity and high-dimensional nature. An alternative approach is to predict semantic modalities, which offers several advantages \cite{luc2017predicting,chiu2020segmenting, vsaric2019single,nabavi2018future,liu2023meta}. First, forecasting semantic information simplifies the prediction process by avoiding the need to replicate low-level details that are often less critical for decision-making. Second, semantic modalities provide information that is directly relevant to an agent’s tasks. For instance, when navigating around or avoiding a moving obstacle, predicting depth information is far more useful than modeling the scene’s precise visual appearance.

Motivated by these observations, we propose a novel framework for future semantic prediction. Our approach is built on a multimodal visual sequence transformer architecture that is optimized for the joint prediction of multiple modalities and trained using a masked visual modeling objective. It accepts inputs from various modalities across past frames and predicts the same set of modalities for future frames. This multimodal capability is crucial for practical applications, as different modalities often offer complementary insights that, when combined, enhance the agent's overall understanding of its environment. Moreover, using multiple semantic modalities has a synergistic effect: they can mutually improve the accuracy of future predictions due to the natural correlations between them, which transformers are well-suited to capture. Our experiments confirm the efficacy of this synergistic approach, which is also one of the contributions of this work.

An important consideration in developing a multimodal video transformer is to make high-resolution images efficient for transformer processing while ensuring the easy integration of multiple modalities~\cite{Girdhar_2023_CVPR, jaegle2022perceiver, mizrahi2024_4m, bachmann2024_4m, Sun_2024_CVPR,bai2024sequential, lu2024unified,bachmann2022multimae, xiao2024florence, hu2021unit}. A seemingly obvious approach, inspired by recent transformer-based video generative methods, might involve using Variational Autoencoder (VAE)-based embedding techniques~\cite{yu2023magvit, yan2021videogpt,tschannen2023givt,rakhimov2020latent,yu2024language,kondratyuk2024videopoet}. However, this approach comes with significant drawbacks, including the need to train separate VAEs for each modality, making it impractical for multimodal setups. Moreover, VAE-based methods face inherent challenges, such as difficulties in stabilizing training~\cite{esser2021taming, razavi2019generating}, which further complicates the integration of diverse modalities. These limitations underscore the need for a more efficient solution to handle multimodal integration effectively.
Thus, we adopt a VAE-free embedding strategy, employing a hierarchical embedding technique that progresses from pixel-level to patch-level embeddings for each modality. This end-to-end approach simplifies the training pipeline and eliminates the dependency on pre-trained VAEs. To further improve the efficiency and effectiveness of our transformer training, we introduce a cross-modality fusion mechanism within our embedding process. This technique reduces the token sequence length by merging tokens from different modalities that share the same spatial-temporal location, producing a more compact and computationally efficient representation.

A key aspect of our approach is that our multimodal transformer can be trained without ground truth label maps for any modality. Instead, we demonstrate that using pseudo-labels generated by off-the-shelf foundation models \cite{strudel2021segmenter, yang2024depth} is sufficient. Remarkably, even under this relaxed supervision, we demonstrate that our multimodal visual sequence transformer generalizes well and achieves state-of-the-art performance.

In summary, our contributions are the following:
\begin{itemize}
    \item 
    We introduce \ours, a method for future multimodal semantic prediction using a simple, efficient, unified transformer architecture that leverages a masked visual modeling task and and offers easy adaptation to various modalities.
    To our knowledge, this is the first work to employ a transformer architecture for multimodal semantic future prediction, addressing a gap in prior research.
    \item 
    We propose a simple hierarchical tokenization process integrated directly into our multimodal transformer, eliminating the need for costly VAE-based tokenization methods and enabling end-to-end training.
    \item We develop a specifically adapted masking mechanism for multimodal training, enabling the model to efficiently leverage visible information across modalities.
    \item Finally, we demonstrate the effectiveness of our unified approach by applying it to future semantic segmentation and depth prediction (see \autoref{fig:teaser}), achieving 
    state-of-the-art results in both domains, highlighting the synergistic benefits of leveraging multimodal information for future prediction. 
\end{itemize}

%% file: sec/2_Related.tex
\section{Related Work}
\label{sec:related}

\begin{figure*}[t]
\centering
\includegraphics[trim={0cm 0cm 0cm 0cm},width=.95\linewidth]{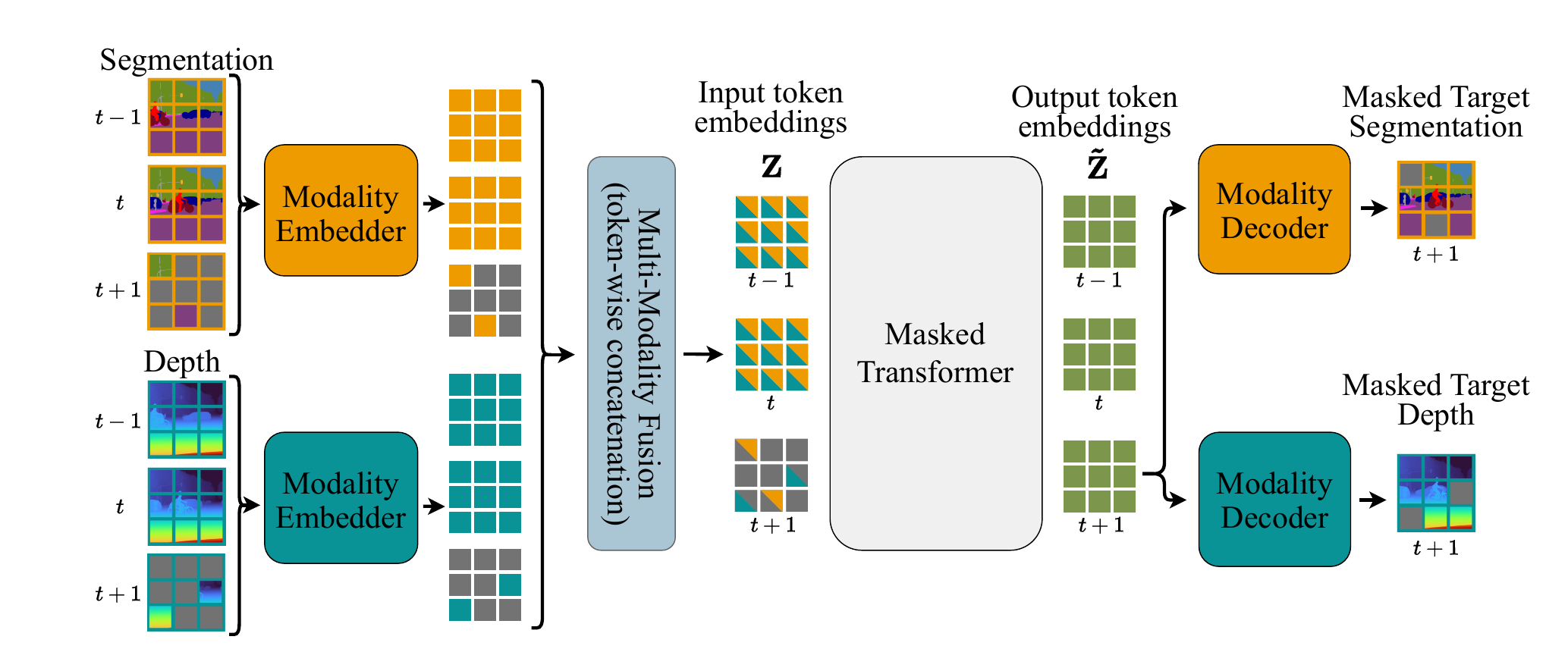}
\caption{\textbf{End-to-End Trainable Multimodal Visual Sequence Transformer.} Our framework employs modality-specific embedders to map inputs into tokens. Then, these token embeddings are fused through token-wise concatenation to form the combined input embedding $\mathbf{Z}$. Next, a masked transformer processes $\mathbf{Z}$, capturing spatiotemporal dependencies between modalities, and outputs $\tilde{\mathbf{Z}}$. Finally, modality-specific decoders produce future frame predictions for each modality based on $\tilde{\mathbf{Z}}$, enabling efficient and accurate multimodal semantic future prediction within a unified architecture. Tokens are shown in 2D (not flattened) for visualization purposes.}
\label{fig:overview}
\end{figure*}

\parag{Future Video Forecasting.}
Future video forecasting and generation have garnered significant attention \cite{oprea2020review} in the computer vision and machine learning communities due to their potential applications in various domains, including autonomous driving and robotics. Early approaches to this task often relied on recurrent neural networks (RNNs) such as Convolutional Long Short Term Memory (Conv-LSTM) architectures \cite{nabavi2018future, gao2022simvp, wu2021motionrnn,Xu_2018_CVPR,wang2018eidetic,lee2021video,castrejon2019improved} to capture temporal dependencies in video sequences. 
However, these methods frequently struggled with long-term predictions and maintaining visual consistency and realism. Other works have leveraged generative adversarial networks (GANs) and variational autoencoders (VAEs) \cite{vondrick2016generating,lee2018stochastic,castrejon2019improved,yan2021videogpt,babaeizadeh2018stochastic} to produce more realistic and diverse future frames. More recently, techniques leveraging diffusion models \cite{ho2022imagen,ho2022video,gao2024vista}, demonstrate impressive visual results capturing both spatial and temporal relationships. Finally, inspired from their success in natural language models, some works extend the ideas of auto-regressive and masked modeling objectives to images \cite{chang2022maskgit,chen2020generative} and videos \cite{Weissenborn2020Scaling,gupta2023maskvit,yu2023magvit,yu2024language} and employ pure transformer models in conjunction with a tokenization module~\cite{razavi2019generating,van2017neural,esser2021taming}. In our work, we propose a method that directly predicts semantic modalities for future frames without the need for complex tokenization modules, offering a simpler and more efficient approach to future forecasting. 

\parag{Multimodal Learning with Transformers.} 
To effectively process multiple dense visual modalities and enable robust cross-modal interactions, several multimodal transformer approaches have been proposed. Some approaches employ flexible architectures that handle arbitrary input and output modalities. Notably, Perceiver IO~\cite{jaegle2022perceiver} and ImageBind~\cite{Girdhar_2023_CVPR} integrate embeddings from multiple modalities into a shared representation space, enabling unified processing of diverse data types within a single framework. Other works extend masked autoencoders to multimodal and multitask settings, such as MultiMAE~\cite{bachmann2022multimae}, which introduces a masked transformer for joint pretraining across modalities.  Building upon this, models for massively multimodal masked modeling and any-to-any vision tasks have been introduced \cite{mizrahi2024_4m, bachmann2024_4m}.

To further unify representations across various tasks, transformer architectures have been leveraged to integrate multiple modalities within a single framework. For instance, Xiao \etal\cite{xiao2024florence} focused on combining vision and language data, while Lu \etal\cite{lu2024unified} have integrated vision, language, audio, and action data. Additionally, generative transformer-based multimodal models have demonstrated the ability to perform in-context learning across modalities, improving adaptability and performance in diverse tasks~\cite{Sun_2024_CVPR}. Moreover, sequential modeling techniques have been applied to large multimodal transformers, enabling scalable training and enhancing performance in multimodal tasks~\cite{bai2024sequential}. Furthermore, UniT~\cite{hu2021unit} introduces a unified transformer framework for multimodal multitask learning, effectively processing images, text, and videos across various tasks within a single model. Our work introduces a VAE-free multimodal visual sequence transformer for semantic future prediction. The model employs early multimodality fusion through token concatenation and enables end-to-end training without complex tokenization modules.

\parag{Semantic Future Prediction.}
Future segmentation and depth prediction aim to use past frames to predict the segmentation masks and depth maps of future frames. In semantic segmentation, Luc \etal\cite{luc2017predicting} introduced the first method for predicting future semantics by exploring different input configurations. They experimented with using previous RGB frames, past segmentation maps, or a combination of both. Their results showed that a two-phase approach—predicting future frame pixel values and then using these predictions to generate segmentation masks—was less effective. We further validate this finding by applying recent diffusion-based, state-of-the-art driving world models for future RGB frame generation.

To better capture the temporal nature of videos, several approaches have been proposed. Focusing on the architecture, \cite{nabavi2018future, sun2019predicting} integrated ConvLSTM modules, while \cite{chen2019multi} introduced an attention mechanism. \cite{chiu2020segmenting} adopted a teacher-student framework, where the student performs the forecasting task, and the teacher provides semantic guidance through knowledge distillation. Another promising direction involves flow-based forecasting \cite{jin2017predicting, terwilliger2019recurrent, saric2020warp, vsaric2021dense, ciamarra2022forecasting}, where the prediction model uses optical flow information. Jin \etal\cite{jin2017predicting} developed a method that predicts both optical flow and future semantic segmentations by allowing the segmentation model to leverage motion features from a flow-predicting network. Terwilliger \etal\cite{terwilliger2019recurrent} further improved this by using a convolutional LSTM to predict optical flow and then warp the last segmentation mask. In a similar approach, Ciamarra \etal\cite{ciamarra2022forecasting} applied optical flow warping to instance segmentation forecasting.

In recent work, \cite{saric2020warp, vsaric2021dense} proposed the F2MF model, which extends the F2F (Feature-to-Feature) approach \cite{luc2018predicting}, where features are regressed from observed frames, by warping deep features to generate future semantic segmentation. For instance and panoptic segmentation forecasting, several methods \cite{graber2021psf,vip-deeplab,graber2022joint} have been proposed that extend beyond semantic boundaries to predict individual object instances. For future depth prediction, Liu \etal\cite{liu2023meta, nag2022far} tackled the problem of predicting depth maps for unobserved frames. Depth prediction was also explored in \cite{hu2020probabilistic, qi20193d, ciamarra2024flodcast} as part of multimodal prediction, where depth was combined with other modalities like optical flow, semantic segmentation, and scene understanding to enhance the accuracy and robustness of future scene interpretations.

In our work, we also focus on multimodal semantic future prediction. However, unlike prior approaches, we use a multimodal visual sequence transformer rather than convolutional or recurrent neural networks. We treat the next-frame prediction task as a masked visual modeling problem \cite{chang2022maskgit, gupta2023maskvit}. By using transformer architectures \cite{vaswani2017attention} for multimodal semantic prediction, we can better model long-range dependencies in the data. 

%% file: sec/3_Methodology.tex
\section{Multimodal Visual Sequence Transformer}
\label{sec:method}

We propose \ours, a flexible framework for semantic future prediction that processes video sequences in a spatially dense, multimodal way. Our framework uses a video sequence of $N_c$ frames from timesteps $\{t - N_c + 1, \ldots, t\}$ as context for making predictions. These input frames may represent any semantic modality, such as semantic segmentation or depth maps, or multiple modalities simultaneously. The objective is to predict future frames for the input modalities in a frame-wise autoregressive manner, with a focus on the short-term (${t+1}$) and mid-term (${t+3}$) future, consistent with previous studies \cite{luc2017predicting, luc2018predicting, chiu2020segmenting, lin2021predictive}.

To accomplish this, we use a multimodal visual sequence transformer and approach 
the next-frame prediction task as a masked visual modeling problem \cite{chang2022maskgit, gupta2023maskvit}. The transformer processes a sequence of $N = N_c + N_p$ frames, comprising both context frames ($N_c$) and target future frames ($N_p$). Future frames are either fully or partially masked at the input, with the transformer's task being to predict the missing values. Typically, we set $N_p = 1$, predicting one future frame at a time. An overview of this method is shown in \autoref{fig:overview}.

While our framework can model individual modalities like semantic segmentation or depth maps, it is particularly effective in jointly modeling these visual modalities. Moreover, it is adaptable and can incorporate other modalities such as instance or panoptic segmentation and optical flow.

\paragraph{Pseudo-Labeled Multimodal Training Dataset.}
Manual annotation of video sequences for semantic segmentation or depth maps is labor-intensive.  To address this, we generate these maps using off-the-shelf pretrained models. Specifically, we use Segmenter \cite{strudel2021segmenter}, a well-established transformer model for semantic segmentation, along with DepthAnythingV2 \cite{yang2024depth}, a foundation model that offers improved depth maps compared to its predecessor \cite{yang2024depth_v1}.

\subsection{VAE-free Multimodal Embedder} \label{sec:embedder}

\paragraph{Integrated Per-Modality Tokenization Process.} For each modality $\mathcal{I}$, let $\mathbf{X}_\mathcal{I} \in \mathcal{L}_\mathcal{I}^{N \times H \times W}$ represent a video sequence of $N$ frames, where $\mathcal{L}_\mathcal{I}$ is the set of possible discrete  pixel values for modality $\mathcal{I}$. For instance, $\mathcal{L}_\mathcal{I}$ could be semantic labels for segmentation or discrete depth values for depth prediction. To make high-resolution images manageable for transformer processing, most generative transformer methods pre-train VAE-based tokenizers that convert high-resolution images to a reduced set of token embeddings, $L$, where $L \ll H \cdot W$.

\begin{figure}[t]
\centering
\includegraphics[trim={0cm 0cm 0cm 0cm},width=1.0\linewidth]{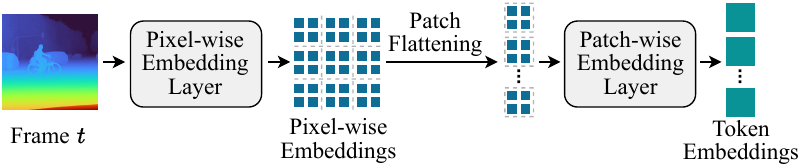}
\caption{
\textbf{Modality Embedder.} Our method embeds per-pixel values and aggregates them into patch tokens.} 
\label{fig:hiera_embedder}
\end{figure}

Instead of using pre-trained VAE-based tokenizers, we propose a simple two-stage hierarchical tokenization process, illustrated in \autoref{fig:hiera_embedder}, that is integrated directly into the video transformer model. First, our tokenizer performs per-pixel embedding, converting $\mathbf{X}_\mathcal{I}$ into continuous per-pixel embeddings $\mathbf{X}^E_{\mathcal{I}} \in \mathbb{R}^{N \times H \times W \times C_\mathcal{I}}$ using a trainable embedding matrix $W_\mathcal{I} \in \mathbb{R}^{|\mathcal{L}_\mathcal{I}| \times C_\mathcal{I}}$. To reduce memory usage and the number of parameters, we use low-dimensional embeddings (e.g., $C_\mathcal{I} = 10$). Next, we convert $\mathbf{X}^E_{\mathcal{I}}$ into patch token embeddings by dividing it into $P \times P$ patches, flattening them, and applying a linear projection with parameters $W^{'}_\mathcal{I} \in \mathbb{R}^{(P^2 \cdot C_\mathcal{I}) \times d_\mathcal{I}}$. This results in the final token embeddings $\mathbf{Z}_\mathcal{I} \in \mathbb{R}^{(N \cdot L) \times d_\mathcal{I}}$, where $L = \frac{H \cdot W}{P^2}$ and $d_\mathcal{I}$ is the embedding dimension for modality $\mathcal{I}$. By default, we use $P = 16$ as patch size. By using low-dimensional embeddings in the first stage, we ensure that the number of parameters $(P^2 \cdot C_\mathcal{I}) \times d_\mathcal{I}$ in $W^{'}_\mathcal{I}$ remains manageable.

Our approach simplifies tokenization compared to other video generative transformers that often rely on complex VAE-based tokenizers, with either continuous \cite{tschannen2023givt, li2024autoregressive} or discrete tokenization strategies \cite{yu2023magvit, yu2024language}, such as VQ-VAE \cite{van2017neural, esser2021taming, razavi2019generating}. In contrast, our VAE-free method integrates tokenization directly into the video transformer model. This integration streamlines the training process by enabling end-to-end training of the entire video generation pipeline and eliminates the need for training separate tokenization models for each modality. This reduces computational overhead and results in a more efficient and scalable framework for future semantic prediction.

\paragraph{Efficient Multimodality Fusion.}  
We aim to leverage the transformer's ability to capture dependencies across modalities while keeping computation manageable when processing all $K$ available modalities. Concatenating the $K$ per-modality video sequences into a single long sequence of length $K \cdot N \cdot L$ would slow down processing due to the quadratic complexity of self-attention. Instead, we adopt an early fusion approach to create a combined embedding $\mathbf{Z} \in \mathbb{R}^{(N \cdot L) \times d}$, where $d$ is the transformer's hidden dimension, by merging the per-modality embeddings $\{\mathbf{Z}_{\mathcal{I}_i}\}_{i=1}^K$ along the embedding dimension.

Specifically, we explored two simple early-fusion strategies: (i) \textsc{Concat}, which concatenates embeddings across the embedding dimension, \ie, $\mathbf{Z} = \mathrm{concat}(\mathbf{Z}_{\mathcal{I}_1}, \mathbf{Z}_{\mathcal{I}_2}, \ldots, \mathbf{Z}_{\mathcal{I}_K}) \in \mathbb{R}^{(N \cdot L) \times d}$. Here, embedding dimensions are chosen such that $\sum_{i=1}^K d_{\mathcal{I}_i} = d$ (e.g., for segmentation $\mathcal{S}$ and depth $\mathcal{D}$, we set $d_\mathcal{S} = d_\mathcal{D} = \frac{d}{2}$). (ii) \textsc{Add}, where embeddings are summed, \ie, $\mathbf{Z} = \sum_{i=1}^K \mathbf{Z}_{\mathcal{I}_i}$. Here, the per-modality embedding dimensions are set to $d$. Our experiments show that \textsc{Concat} gives better results, and we use it by default.

\begin{figure}[t]
\centering
\includegraphics[trim={0cm 0cm 0cm 0cm},width=1.0\linewidth]{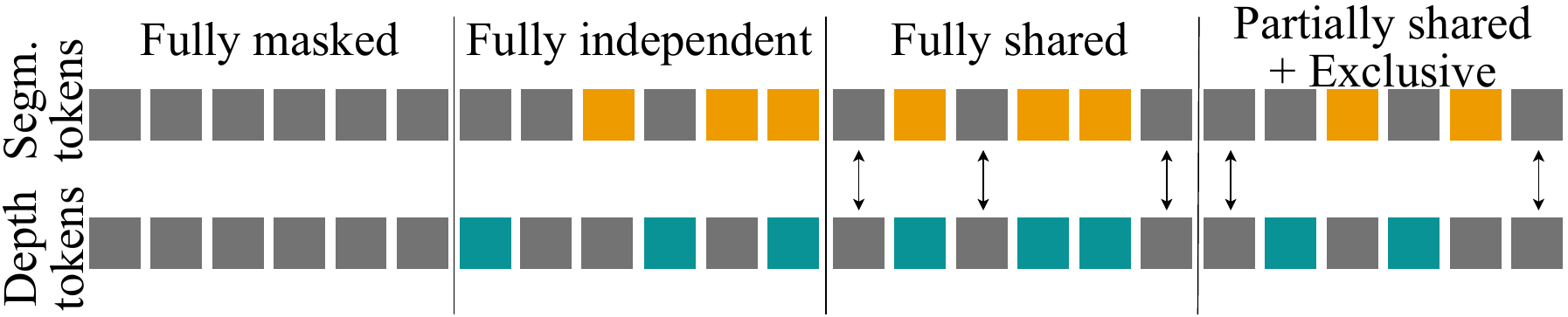}
\caption{\textbf{Masking strategies.} We explored four strategies for masking future-frame tokens across modalities: Fully Masked (all tokens masked), Fully Independent (independently masked tokens per modality), Fully Shared (shared mask across modalities), and Partially Shared + Exclusive (shared mask for some tokens, with others masked exclusively per modality).}
\label{fig:masking_strategies}
\end{figure}

\subsection{Multimodal Masking Strategies} \label{sec:masking}

During training, we keep all tokens from the $N_c$ context (past) frames intact and randomly mask a subset of tokens from the $N_p$ future frames. Let $\mathbf{M}_{\mathcal{I}} \in \{0, 1\}^{N_p \cdot L}$ be a binary mask indicating which patch tokens in the $N_p$ future frames are masked (1) or visible (0) for each modality $\mathcal{I}$. To create a mask $\mathbf{M}$, we first sample a masking ratio $r \in (0,1)$ (selected uniformly) and adjust it with a scheduling function $\gamma(r)$. The final number of masked tokens for the future frames is $\lfloor \gamma(r) \cdot N_p \cdot L \rfloor$.

To enhance masked visual modeling, we allow masking to occur across different modalities for each future frame token. We explored the following multimodal masking strategies, illustrated in \autoref{fig:masking_strategies}:

\begin{itemize} 
\item \textbf{Fully-independent:} Each modality $\mathcal{I}$ generates its own independent mask. We explore two variants: one with independently chosen masking ratios $r$ per modality and another with a shared masking ratio $r$ across modalities. 
\item \textbf{Fully-shared:} All modalities share a single, randomly-sampled, mask $\mathbf{M}$, meaning the same tokens are masked for each modality. 
\item \textbf{Partially-shared + Exclusive:} First, a common initial mask $\mathbf{M}$ is sampled for all modalities. Then, to create per-modality masks $\mathbf{M}_{\mathcal{I}}$, each future frame token marked as visible in $\mathbf{M}$ is set to be visible in only one of the $K$ modalities, while remaining masked in the others. 
\item \textbf{Fully-masked:} All future-frame tokens are masked simultaneously across all modalities. This baseline strategy differs from the others in that it is not random, masking all tokens in the future frames without sampling a subset. 
\end{itemize}

For each modality $\mathcal{I}$, the future-frame tokens in $\mathbf{Z}_\mathcal{I}$ indicated as masked in $\mathbf{M}_{\mathcal{I}}$ are replaced with a modality-specific learnable mask embedding $\textsc{[mask]}_\mathcal{I} \in \mathbb{R}^{d_{\mathcal{I}}}$.

\paragraph{Inference Time.} At this phase, all future-frame tokens are treated as masked tokens and are predicted with single step.

\subsection{Spatio-Temporal Transformer with Decomposed Attention} \label{sec:transformer}

After the per-modality masking process, the transformer processes the fused multimodal token embeddings $\mathbf{Z}$ across all frames, treating future-frame tokens as masked during inference. To begin, we add learnable position embeddings to the token embeddings $\mathbf{Z}$, which are then passed through alternating layers of bidirectional self-attention and MLPs, producing the output token embeddings $\mathbf{\tilde{Z}}$.

To address the quadratic complexity of Multi-Head Self Attention (MSA) with respect to the number of tokens \cite{vaswani2017attention}, 
we decompose attention into temporal and spatial components, similar to \cite{arnab2021vivit, gupta2023maskvit}.
Temporal attention is applied to tokens at the same spatial position across frames, while spatial attention is applied to tokens within the same frame. Both the MSAs and MLPs are implemented using Pre-LN \cite{xiong2020layer}.

\subsection{Multimodal Decoder} \label{sec:decoder}

After the final transformer layer, we use $K$ prediction heads, one for each modality, to decode the transformer's output token embeddings $\mathbf{\tilde{Z}}$ into $K$ sets of pixel-wise predictions.

\paragraph{Per-Modality Decoder}
For each modality $\mathcal{I}$, decoding from the transformer’s output embeddings $\mathbf{\tilde{Z}}$ to pixel-wise predictions follows a two-stage approach that mirrors the hierarchical structure of the per-modality embedder. First, a linear layer projects the token embeddings $\mathbf{\tilde{Z}} \in \mathbb{R}^{(N \cdot L) \times d}$ from $d$ dimensions to $P^2 \cdot C_{\mathcal{I}}$ dimensions, producing intermediate embeddings $\mathbf{\tilde{X}}^{E}_{\mathcal{I}} \in \mathbb{R}^{(N \cdot L) \times (P^2 \cdot C_{\mathcal{I}})}$. 

Next, we reshape $\mathbf{\tilde{X}}^{E}_{\mathcal{I}}$ to dimensions $N \times H \times W \times C_\mathcal{I}$ and apply a weight matrix $\tilde{W}_\mathcal{I} \in \mathbb{R}^{C_\mathcal{I} \times |\mathcal{L}_\mathcal{I}|}$ followed by a softmax operation, yielding the final per-pixel predictions $\mathbf{\tilde{X}}_{\mathcal{I}}\in \mathbb{R}^{N \times H \times W \times |\mathcal{L}_\mathcal{I}|}$. Following common practices in language modeling, we tie $\tilde{W}_\mathcal{I}$ to the transpose of the embedding matrix $W_\mathcal{I}\in \mathbb{R}^{|\mathcal{L}_\mathcal{I}| \times C_\mathcal{I}}$ from the embedder, i.e., $\tilde{W}_\mathcal{I} = W^\top_\mathcal{I}$.

Like the embedder, our decoding approach avoids VAEs and is fully integrated within the transformer, enabling end-to-end training of the full video prediction pipeline.
 
\subsection{End-to-End Training for Masked Multimodal Visual Modeling} \label{sec:training}

To effectively learn the multimodal next-frame prediction task, we introduce a novel masked visual modeling objective tailored for multimodal data. Building on traditional masked visual modeling methods designed for single-modality data \cite{bao2021beit, he2022masked, chang2022maskgit, gupta2023maskvit}, our method integrates multiple modalities into a unified, end-to-end trainable framework.

\paragraph{Per-Modality Masked Modeling Loss.}  
To simplify the loss description, we define $\mathbf{\tilde{X}}^{F}_{\mathcal{I}} \in \mathbb{R}^{(N_p \cdot L) \times P^2 \times |\mathcal{L}_\mathcal{I}|}$ and $\mathbf{X}^{F}_{\mathcal{I}} \in \mathcal{L}_\mathcal{I}^{(N_p \cdot L) \times P^2}$ as the per-modality model predictions and targets for the $N_p$ future frames. These are reshaped so that the first two dimensions represent the number of tokens ($N_p \cdot L$) and the number of pixels ($P^2$) within each token, respectively. The per-modality loss $\mathcal{L}_{\mathcal{I}}(\mathbf{X}_{\mathcal{I}}, \mathbf{M}_{\mathcal{I}})$ is computed as the cross-entropy loss between the predictions and the targets for the future-frame tokens that are masked:
\begin{equation} \label{eq:modality_objective}
\mathcal{L}_{\mathcal{I}}(\mathbf{X}_{\mathcal{I}}, \mathbf{M}_{\mathcal{I}}) = \sum_{n}^{N_p \cdot L} \mathbf{M}_{\mathcal{I}}(n) \sum_{p}^{P^2}  \mathcal{L}_{CE}(\mathbf{\tilde{X}}^{F}_{\mathcal{I}} (n,p), \mathbf{X}^{F}_{\mathcal{I}} (n,p))
\end{equation}
where $\mathcal{L}_{CE}(\cdot, \cdot)$ is the cross-entropy loss function, $(n,p)$ indexes the values in $\mathbf{\tilde{X}}^{F}_{\mathcal{I}}$ and $\mathbf{X}^{F}_{\mathcal{I}}$ corresponding to the $p$-th pixel within the $n$-th token, and $(n)$ indexes the masking value of the $n$-th token in $\mathbf{M}_{\mathcal{I}}$.

\paragraph{End-to-End Multimodal Training.}
Our multimodal video transformer is trained end-to-end using the following multimodal masked visual modeling objective:
\begin{equation} \label{eq:total_objective}
\mathcal{L}_{\mathrm{MVM}} = \underset{\{\mathbf{X}_{\mathcal{I}_i}, \mathbf{M}_{\mathcal{I}_i}\}_{i=1}^K \sim \mathcal{D}}{\mathbb{E}} \sum_{i=1}^K w_{\mathcal{I}_i} \mathcal{L}_{\mathcal{I}}(\mathbf{X}_{\mathcal{I}_i}, \mathbf{M}_{\mathcal{I}_i}),
\end{equation}
where $w_{\mathcal{I}_i}$ is the weight coefficient for the loss of the $\mathcal{I}_i$ modality. 
Unlike most transformer-based generative models that rely on pre-trained tokenizers, our framework trains all modules—the embedders (\autoref{sec:embedder}), the masked transformer backbone (\autoref{sec:transformer}), and the decoder heads (\autoref{sec:decoder})—together using the objective defined in \autoref{eq:total_objective}. This end-to-end approach ensures that the entire model is optimized jointly, making our multimodal visual modeling framework both simple and effective.

%% file: sec/4_Experiments.tex
\section{Experiments}
\label{sec:experiments}

\subsection{Experimental Protocol}

\parag{Data.} 
To evaluate our method, we use the Cityscapes dataset \cite{Cordts_2016_CVPR}, which contains video sequences of a car driving in urban environments. The dataset includes 2,975 training sequences, 500 validation sequences, and 1,525 test sequences, each with 30 frames at 16 fps and a resolution of 1024 × 2048 pixels. The 20th frame in each sequence has semantic segmentation annotations for 19 classes. We train on the training split and evaluate on the validation split.

\parag{Implementation details.} For our models we set \( C_\mathcal{I} = 10 \) for the pixel-wise embeddings. We use patch size of \( P = 16 \), which on frames of 256 × 512 resolution produces \( 16 \times 32 = 512 \) patch tokens per frame. The transformer, built upon the implementation of \cite{besnier2023pytorch}, has 12 layers with a hidden dimension of \( d = 1536 \). We train both multimodal models (segmentation and depth modalities) and single-modality variations for these two tasks. Unless stated otherwise, we use the \textsc{Concat} fusion strategy, the ``Partially-shared + Exclusive" multimodal masking strategy, and a sequence length \( N = 5 \) (with \( N_c = 4 \) context frames and \( N_p = 1 \) future frame). For end-to-end training, we use the Adam optimizer \cite{adamopt} with momentum parameters \( \beta_1 = 0.9 \), \( \beta_2 = 0.99 \), and a learning rate of \( 1.6 \times 10^{-4} \) with cosine annealing. Training is conducted on 8 A100 40Gb GPUs with an effective batch size of 64. The longest training run was 3200 epochs (256,512 iterations) over 32 hours. For most experiments, we use 800 epochs (approximately 8 hours).

\parag{Evaluation Metrics.} 
To assess our method's performance in predicting future semantic segmentation and depth maps, we evaluate on full-resolution using the following evaluation metrics.

For semantic segmentation, we apply two metrics based on mean Intersection over Union (mIoU) between predicted and ground truth segmentations. The first metric, mIoU (ALL), includes all available semantic classes. The second metric, MO-mIoU (MO), evaluates only classes representing movable objects, including person, rider, car, truck, bus, train, motorcycle, and bicycle.

For depth prediction, we use two standard evaluation metrics \cite{Godard_2017_CVPR, liu2023meta, depth_est}. The first is the mean Absolute Relative Error (AbsRel), defined as \(\frac{1}{M} \sum_{i=1}^M \frac{|a_i - b_i|}{b_i} \), multiplied by 100 for readability, where \( a_i \) and \( b_i \) represent the ground truth and predicted disparities at pixel \( i \), respectively, and \( M \) is the total number of pixels. We also evaluate depth accuracy with a threshold metric, \( \delta_{1} \), which measures the percentage of pixels where \(\max\left(\frac{a_i}{b_i}, \frac{b_i}{a_i}\right)\) falls below 1.25.

\parag{Evaluation Scenarios.}
Following prior work \cite{luc2017predicting, nabavi2018future}, we consider two evaluation scenarios: \emph{short-term prediction} (3 frames ahead, ~0.18s) and \emph{mid-term prediction} (9 frames ahead, ~0.54s). In both cases, the 20th frame is the forecasting target. 
We temporally subsample sequences by a factor of 3 before processing them with our model. This means that, for \emph{short-term prediction}, a model with a context length of four frames (i.e., $N_c=4$ and $N_p=1$) uses frames 8, 11, 14, and 17 as context to predict frame 20. For \emph{mid-term prediction}, the model uses frames 2, 5, 8, and 11 as context and predicts frame 20 auto-regressively (first frame 14, then 17, and finally frame 20).

We calculate segmentation metrics using the ground truth annotations for the 20th frame in Cityscapes. For depth, since Cityscapes lacks true depth annotations, we use pseudo-annotations from DepthAnythingV2 for evaluation.

\subsection{Quantitative Results}
Here, we quantitatively evaluate our model, \ours, on predicting future semantic segmentation and depth maps.

\parag{Baselines.} 
We compare our method against three baselines. The first is the \emph{Oracle} baseline, which applies the image segmentation model (Segmenter \cite{strudel2021segmenter}) or the monocular depth prediction model (DepthAnythingV2 \cite{yang2024depth}) directly on the target future frame. 
Since these models are the same ones used to compute segmentation and depth maps for the context frames, the Oracle baselines serve as an upper bound for our method's performance.

The second baseline, \emph{Copy-Last}, simply copies the segmentation or depth maps from the last available context frame to the target future frame, providing a lower bound for our method's performance.

The third baseline uses the state-of-the-art driving world model, \emph{VISTA}~\cite{gao2024vista},  which generates future RGB frames conditioned on three context frames through a video diffusion model. VISTA is a large-scale model with 2.5 billion parameters, trained on 1,740 hours of driving videos from YouTube. In this baseline, we finetune VISTA on Cityscapes and use it (with action-free conditioning) to generate future RGB frames, then apply the Segmenter and DepthAnythingV2 models to these frames to produce future-frame segmentation and depth maps. This comparison allows us to assess our method, which directly predicts future segmentation and depth maps, against the state-of-the-art in generative-based future prediction. More details regarding Vista finetuning in  \hyperref[sec:vista_ft_comp]{Appendix~\ref*{sec:vista_ft_comp}}. 

\input{sec/tab_comparison_segmentation}
\input{sec/tab_modality_comparison}

\parag{Comparison with Prior Work on Segmentation Forecasting.}
In \autoref{tab:comparison_with_sota}, we compare our model with previous semantic future prediction methods on the Cityscapes dataset. We report results for both the default multimodal version (using segmentation and depth modalities) and a segmentation-only variant. Both versions outperform prior methods in short- and mid-term predictions, with the multimodal version achieving the best performance and setting a new state-of-the-art. Our model shows significant improvements over the previous state-of-the-art method, PFA \cite{lin2021predictive}, especially for movable object (MO) classes, which are typically more difficult to predict.

\input{sec/fig_model_scaling}

\parag{Comparison with VISTA.}
We also provide results for the VISTA-based baseline in \autoref{tab:comparison_with_sota}. VISTA performs notably worse, even lower than the Copy-Last baseline. As shown in \autoref{fig:qualitative-results}, this is due to VISTA’s inability to maintain the semantic structure of the scene in future frames. Despite generating visually realistic frames, VISTA's future predictions often suffer from issues such as object disappearance and artifacts that lead to inaccurate results from the oracle segmentation and depth models. These limitations highlight the advantage of our approach, which models future predictions directly in semantic space. Complete quantitative results, including depth metrics, are provided in \hyperref[tab:comparison_with_vista]{Appendix Tab. ~\ref*{tab:comparison_with_vista}}.

\parag{Multi-modal vs Single-modality models.}
In \autoref{tab:modal-comparison}, we compare our model's default multimodal version (segmentation plus depth) to single-modality versions using only segmentation or only depth. Results for both segmentation and depth forecasting show that the multimodal model outperforms the single-modality versions, demonstrating that it captures synergistic effects between the two modalities, enabling more accurate future predictions.

\input{sec/tab_fusion_strategies}
\input{sec/tab_masking_strategies}
\input{sec/tab_sequence_length}

\parag{Scaling Training Epochs.} 
Building on research into scaling laws for transformer-based architectures \cite{kaplan2020scaling, zhai2022scaling}, we investigate how our model’s performance scales with the number of training epochs. As shown in \autoref{fig:ScalingSegm}, extending the training duration results in significant improvements in both segmentation and depth prediction. These results suggest that our model can continue to enhance its predictive capabilities with more training, even when trained on limited datasets like Cityscapes.

\parag{Modality Fusion Strategies}
In \autoref{tab:fusion-strategies}, we compare the \textsc{ADD} and \textsc{Concat} strategies for multimodal fusion. 
Both strategies are effective, with \textsc{Concat} yielding slightly better results, making it our default choice.

\begin{figure}[t!]
    \centering
    \include{sec/fig_qualitative_results}
    \vspace{-30pt}
    \caption{
\textbf{Qualitative comparison of future semantic segmentation and depth prediction.} Oracle results are derived from Segmenter~\cite{strudel2021segmenter} (segmentation) and DepthAnythingV2~\cite{yang2024depth} (depth). We compare against the state-of-the-art method VISTA~\cite{gao2024vista}, which generates future RGB frames via video diffusion, followed by Segmenter and DepthAnythingV2 for predictions.}
    \label{fig:qualitative-results}
    \vspace{-10pt}
\end{figure}
\begin{figure}[t!]
    \centering
    \include{sec/fig_qualitative_rollout}
    \vspace{-10pt}
    \caption{\textbf{Long-term semantic segmentation and depth predictions.} Our method uses autoregressive rollouts to predict future semantic and depth maps over long horizons. Starting from four context frames (\(X_{t-9}\) to \(X_t\)), it generates coherent predictions up to 48 frames (2.88 seconds) with a frame-wise step of 3, accurately capturing both static elements and dynamic changes. Due to space constraints, we present only frames \(X_{t+3}\), \(X_{t+18}\), and \(X_{t+48}\).
    }
    \label{fig:qualitative-rollout}
    \vspace{-10pt}
\end{figure}

\parag{Masking Strategies.}
In \autoref{tab:masking-strategies}, we compare various multimodal training masking strategies discussed in \autoref{sec:masking}. The \textbf{Partially-shared + Exclusive} strategy achieves the best results across all metrics. This strategy restricts future-frame masking during training such that each future-frame token is either fully masked (both modalities are hidden) or partially masked (only one modality is visible). Importantly, no future-frame token has both modalities visible at the same time. We believe this approach helps the model better learn the synergistic effects between the two modalities. In contrast, the \textbf{Fully-Masked} strategy yields the worst results, despite matching the masking approach used during inference. We hypothesize this is because the strategy is too aggressive and less effective in promoting the model's learning of interactions between the modalities.

\parag{Sequence Length Impact.}
In \autoref{tab:Ablation-SeqLen} we compare models with sequence lengths $N$ of 4, 5, and 7 frames (with $N_p=1$). While all models perform similarly, the 7-frame model achieves slightly better results for mid-term segmentation and all depth prediction metrics. However, we use the 5-frame context by default in this paper due to the higher computational cost of the 7-frame model.

\parag{Additional Experiments.} In the \autoref{sec:addresults}, we present additional experiments validating our design choices. First, in \hyperref[tab:comparison_with_vqvae]{Appendix Tab.~\ref*{tab:comparison_with_vqvae}} we demonstrate that our VAE-free tokenization approach significantly outperforms alternative tokenization methods using pretrained and finetuned VQ-VAE models. Second, in \hyperref[tab:sep-tokens-comparison]{Appendix Tab.~\ref*{tab:sep-tokens-comparison}} we examine our multi-modal fusion strategy against a separable tokens approach, showing that under equal compute budgets, our method achieves superior performance on most metrics.

\subsection{Qualitative Results}

We present qualitative results for our method in \autoref{fig:teaser} and \autoref{fig:qualitative-results} on semantic segmentation and depth prediction forecasting, comparing it to the Oracle baseline. Additionally, \autoref{fig:qualitative-results} includes a comparison with the VISTA baseline. Our method effectively forecasts the scene's semantic structure, producing future predictions closely aligned with the Oracle and outperforming the VISTA baseline.

In \autoref{fig:qualitative-rollout}, we show long-horizon rollouts of the semantic segmentation and depth maps generated by consecutive frame-wise autoregressive predictions from our model. Starting from four context frames (\(X_{t-9}\) to \(X_t\)), it produces consistent predictions up to 48 frames ahead, accurately capturing both static elements and dynamic changes. Additional visualizations are provided in ~\autoref{sec:visualizations}.

%% file: sec/tab_comparison_segmentation.tex
\begin{table}
\footnotesize
\centering
\setlength{\tabcolsep}{4.5pt}
\resizebox{0.8\columnwidth}{!}{%
\begin{tabular}{lcccc} \toprule
\mr{2}{\Th{Method}} & \multicolumn{2}{c}{ \Th{Short-term} } & \multicolumn{2}{c}{ \Th{Mid-term} } \\ \cmidrule(r){2-3} \cmidrule(l){4-5}
 & \Th{ALL} & \Th{MO} & \Th{ALL} & \Th{MO} \\
\midrule 
3Dconv-F2F~\cite{chiu2020segmenting} & 57.0 & - & 40.8 & - \\
Dil10-S2S~\cite{luc2017predicting} & 59.4 & 55.3 & 47.8 & 40.8 \\
F2F~\cite{luc2018predicting} & - & 61.2 & - & 41.2 \\
DeformF2F~\cite{vsaric2019single} & 65.5 & 63.8 & 53.6 & 49.9 \\
LSTM M2M~\cite{terwilliger2019recurrent} & 67.1 & 65.1 & 51.5 & 46.3 \\
F2MF~\cite{saric2020warp} & 69.6 & 67.7 & 57.9 & 54.6 \\
PFA~\cite{lin2021predictive}  & 71.1 & 69.2 & 60.3 & 56.7 \\
\midrule 
Oracle~\cite{strudel2021segmenter} & 78.6 & 80.8 & 78.6 & 80.6 \\
Copy-Last & 55.5 & 52.7 & 40.5 & 32.2 \\
Vista Fine-tuned \cite{gao2024vista} & 65.7 & 64.1 & 53.2 & 49.1 \\
\midrule
\ours Segm.-Only (ours) & \underline{73.4} & \underline{74.2} & \underline{62.4} & \underline{60.5} \\
\ours (ours) & \textbf{73.9} & \textbf{74.9} & \textbf{62.7} & \textbf{61.2} \\

\bottomrule
\end{tabular}%
}
\caption{\textbf{Comparison results with future semantic segmentation methods on Cityscapes validation set using \miou metric.} \Th{ALL}: mIoU of all classes. \Th{MO}: mIoU of movable objects.}
\label{tab:comparison_with_sota}

\end{table}

%% file: sec/tab_modality_comparison.tex
\begin{table}[t!]
\small
\centering
\setlength{\tabcolsep}{3.0pt}
\resizebox{\columnwidth}{!}{%
\begin{tabular}{lcccccccc}
\toprule
& \multicolumn{4}{c}{\Th{Segmentation}} & \multicolumn{4}{c}{\Th{Depth}} \\
\cmidrule(r){2-5} \cmidrule(l){6-9}
\mr{2}{\shortstack{\Th{Training} \\ \Th{Modality}}} & \multicolumn{2}{c}{\Th{Short-term}} & \multicolumn{2}{c}{\Th{Mid-term}} & \multicolumn{2}{c}{\Th{Short-term}} & \multicolumn{2}{c}{\Th{Mid-term}} \\
\cmidrule(r{0.3em}){2-3} \cmidrule(l{0.3em}r){4-5} \cmidrule(lr{0.3em}){6-7} \cmidrule(l{0.3em}){8-9}
& \Th{ALL} & \Th{MO} & \Th{ALL} & \Th{MO} & $\delta_{1}$ & \Th{AbsRel} & $\delta_{1}$ & \Th{AbsRel} \\
\midrule
\Th{Segmentation} & 72.5 & 73.3 & 61.1 & 58.6 & - & - & - & - \\
\Th{Depth} & - & - & - & - & 95.4 & 5.923 & 90.6 & 10.194 \\
\Th{Multimodal} & \textbf{72.9} & \textbf{73.8} & \textbf{61.6} & \textbf{59.9} & \textbf{95.8} & \textbf{5.606} & \textbf{91.5} & \textbf{9.490} \\
\bottomrule
\end{tabular}%
}
\caption{\textbf{Multi-modal vs. single-modality approaches.} Trained for 800 epochs. \Th{AbsRel} is multiplied by 100 for readability.}
\label{tab:modal-comparison}
\end{table}

%% file: sec/fig_model_scaling.tex
\begin{figure}[t!]
    \centering
    \begin{tabular}{cc}
    \multicolumn{2}{c}
    {
    \centering
    \begin{tikzpicture}
        \centering
        \begin{customlegend}[legend columns=3, legend style={align=center, column sep=.7ex, nodes={scale=0.9, transform shape}, draw=white!90!black}, legend entries={ \small Segm.-ALL, \small Segm.-MO, \small Depth}]
            \addlegendimage{color=blue, mark=*, line width=1.2pt}
            \addlegendimage{color=orange, mark=square*, line width=1.2pt}
            \addlegendimage{color=gray, mark=triangle*, dashed, line width=1.2pt}
        \end{customlegend}
    \end{tikzpicture}
    }
    \\    
        \centering
        \begin{tikzpicture}
            \tikzstyle{every node}=[font=\scriptsize]
            \begin{axis}[
                width=0.48\linewidth,
                height=0.48\linewidth, 
                xlabel={Epochs},
                ylabel={\miou},
                ymin=71.75, ymax=76, 
                xlabel shift=-3 pt,
                ylabel shift=-3 pt,
                font=\footnotesize,
                xmode=log,
                log basis x=2, 
                xmin=400, xmax=3200,
                xtick={400, 800, 1600, 3200},
                xticklabels={400, 800, 1600, 3200}, 
                grid=both,
                legend cell align={left},
                axis y line*=left,
                axis x line*=bottom,
                tick label style={font=\footnotesize}, 
                label style={font=\footnotesize}, 
                title={(a) Short-term},
                title style={font=\footnotesize},
                title style={yshift=-25.ex,},
            ]
            \addplot[blue, mark=*] coordinates {
                (400, 72.08)
                (800, 72.94)
                (1600, 73.45)
                (3200, 73.92)
            };
            \addplot[orange, mark=square*] coordinates {
                (400, 72.83)
                (800, 73.79)
                (1600, 74.35)
                (3200, 74.86)
            };        
            \end{axis}
            \begin{axis}[
                width=0.48\linewidth,
                height=0.48\linewidth, 
                xlabel={},
                axis y line*=right,
                axis x line=none,
                ymin=5.15, ymax=6,
                font=\footnotesize,
                xmode=log,
                log basis x=2, 
                xmin=400, xmax=3200,
                xtick=\empty,
                yticklabel style={align=right, font=\footnotesize}, 
                grid=none,
            ]
            \addplot[gray, mark=triangle*, dashed] coordinates {
                (400, 5.886)
                (800, 5.60)
                (1600, 5.44)
                (3200, 5.384)
            };             
            \end{axis}
        \end{tikzpicture}
    &
    \hfill
        \centering
        \begin{tikzpicture}
            \tikzstyle{every node}=[font=\scriptsize]
            \begin{axis}[
                width=0.48\linewidth,
                height=0.48\linewidth, 
                xlabel={Epochs},
                xlabel shift=-3 pt,
                ymin=57.70, ymax=64, 
                font=\footnotesize,
                xmode=log,
                log basis x=2, 
                xmin=400, xmax=3200,
                xtick={400, 800, 1600, 3200},
                xticklabels={400, 800, 1600, 3200}, 
                grid=both,
                legend cell align={left},
                axis y line*=left,
                axis x line*=bottom,
                tick label style={font=\footnotesize}, 
                label style={font=\footnotesize}, 
                title={(b) Mid-term},
                title style={font=\footnotesize},
                title style={yshift=-25.ex,},
            ]
            \addplot[blue, mark=*] coordinates {
                (400, 60.54)
                (800, 61.61)
                (1600, 62.10)
                (3200, 62.69)
            };
            \addplot[orange, mark=square*] coordinates {
                (400, 58.42)
                (800, 59.85)
                (1600, 60.41)
                (3200, 61.16)
            };
            \end{axis}
            \begin{axis}[
                width=0.48\linewidth,
                height=0.48\linewidth, 
                xlabel={},
                ylabel={AbsRel},
                ylabel shift=-3 pt,
                axis y line*=right,
                axis x line=none,
                ymin=8.95, ymax=10,
                font=\footnotesize,
                xmode=log,
                log basis x=2, 
                xmin=400, xmax=3200,
                xtick=\empty,
                yticklabel style={align=right, font=\footnotesize}, 
                grid=none,
                legend entries={},
            ]
            \addplot[gray, mark=triangle*, dashed] coordinates {
                (400, 9.843)
                (800, 9.490)
                (1600, 9.09)
                (3200, 9.11)
            };          
            \end{axis}
        \end{tikzpicture}
    \end{tabular}
    \caption{\textbf{Scaling Training Epochs}. Model performance for segmentation forecasting (all classes and movable objects) and depth forecasting, as a function of the number of training epochs. Results for (a) short-term prediction and (b) mid-term prediction.}
    \label{fig:ScalingSegm}
\end{figure}

%% file: sec/tab_fusion_strategies.tex
\begin{table}[t!]
\small
\centering
\setlength{\tabcolsep}{3.0pt}
\resizebox{\columnwidth}{!}{%
\begin{tabular}{lcccccccc}
\toprule
& \multicolumn{4}{c}{\Th{Segmentation}} & \multicolumn{4}{c}{\Th{Depth}} \\
\cmidrule(r){2-5} \cmidrule(l){6-9}
\mr{2}{\shortstack{\Th{Modality Fusion} \\ \Th{Strategy}}} & \multicolumn{2}{c}{\Th{Short-term}} & \multicolumn{2}{c}{\Th{Mid-term}} & \multicolumn{2}{c}{\Th{Short-term}} & \multicolumn{2}{c}{\Th{Mid-term}} \\
\cmidrule(r{0.3em}){2-3} \cmidrule(l{0.3em}r){4-5} \cmidrule(lr{0.3em}){6-7} \cmidrule(l{0.3em}){8-9} 
& \Th{ALL} & \Th{MO} & \Th{ALL} & \Th{MO} & $\delta_{1}$ & \Th{AbsRel} &  $\delta_{1}$ & \Th{AbsRel} \\
\midrule
ADD & 72.2 & 72.8 & 60.7 & 58.5 & 95.5 & 5.793 & 91.0 & 9.763 \\
CONCAT & \textbf{72.5} & \textbf{73.3} & \textbf{60.9} & \textbf{58.9} & \textbf{95.7} & \textbf{5.745} & \textbf{91.1} & \textbf{9.634} \\
\bottomrule
\end{tabular}%
}
\caption{\textbf{Comparison of modality fusion strategies}. All models were trained for 800 epochs with sequence length 5.}
\label{tab:fusion-strategies}
\end{table}

%% file: sec/tab_masking_strategies.tex
\begin{table}[t!]
\small
\centering
\setlength{\tabcolsep}{3.0pt}
\resizebox{\columnwidth}{!}{%
\begin{tabular}{lcccccccc}
\toprule
& \multicolumn{4}{c}{\Th{Segmentation}} & \multicolumn{4}{c}{\Th{Depth}} \\
\cmidrule(r){2-5} \cmidrule(l){6-9}
\mr{2}{\shortstack{\Th{Masking} \\ \Th{Strategy}}} & \multicolumn{2}{c}{\Th{Short-term}} & \multicolumn{2}{c}{\Th{Mid-term}} & \multicolumn{2}{c}{\Th{Short-term}} & \multicolumn{2}{c}{\Th{Mid-term}} \\
\cmidrule(r{0.3em}){2-3} \cmidrule(l{0.3em}r){4-5} \cmidrule(lr{0.3em}){6-7} \cmidrule(l{0.3em}){8-9} 
& \Th{ALL} & \Th{MO} & \Th{ALL} & \Th{MO} & $\delta_{1}$ & \Th{AbsRel} &  $\delta_{1}$ & \Th{AbsRel} \\
\midrule
Indep. (Same r) & 72.5 & 73.3 & 60.9 & 58.9 & 95.7 & 5.745 & 91.1 & 9.634 \\
Indep. (Diff. r) & 71.9 & 72.8 & 59.9 & 57.9 & 95.5 & 5.862 & 90.7 & 9.888 \\
Fully Shared & 72.5 & 73.3 & 61.2 & 59.2 & 95.6 & 5.731 & 91.2 & 9.586 \\
Part. Shared+Excl. & \textbf{72.9} & \textbf{73.8} & \textbf{61.6} & \textbf{59.9} & \textbf{95.8} & \textbf{5.606} & \textbf{91.5} & \textbf{9.491} \\
Fully Masked & 71.2 & 71.6 & 59.5 & 56.9 & 95.3 & 6.139 & 90.6 & 10.122 \\
\bottomrule
\end{tabular}%
}
\caption{\textbf{Comparison of multimodal masking training strategies}. Models trained for 800 epochs. During inference, next-frame tokens are masked and predicted with a single inference step.
}
\label{tab:masking-strategies}
\end{table}

%% file: sec/tab_sequence_length.tex
\begin{table}[t!]
\small
\centering
\setlength{\tabcolsep}{3.0pt}
\resizebox{\columnwidth}{!}{%
\begin{tabular}{lcccccccc}
\toprule
& \multicolumn{4}{c}{\Th{Segmentation}} & \multicolumn{4}{c}{\Th{Depth}} \\
\cmidrule(r){2-5} \cmidrule(l){6-9}
\mr{2}{\shortstack{\Th{Sequence} \\ \Th{Length}}} & \multicolumn{2}{c}{\Th{Short-term}} & \multicolumn{2}{c}{\Th{Mid-term}} & \multicolumn{2}{c}{\Th{Short-term}} & \multicolumn{2}{c}{\Th{Mid-term}} \\
\cmidrule(r{0.3em}){2-3} \cmidrule(l{0.3em}r){4-5} \cmidrule(lr{0.3em}){6-7} \cmidrule(l{0.3em}){8-9}
& \Th{ALL} & \Th{MO} & \Th{ALL} & \Th{MO} & $\delta_{1}$ & \Th{AbsRel} & $\delta_{1}$ & \Th{AbsRel} \\
\midrule
3 & 72.3 & 73.2 & 60.5 & 58.9 & 95.5 & 5.883 & 91.0 & 9.835 \\
4 & \textbf{73.0} & \textbf{74.0} & 61.5 & 59.9 & 95.7 & 5.597 & 91.4 & 9.507 \\
5 & 72.9 & 73.8 & 61.6 & 59.9 & \textbf{95.8} & 5.606 & 91.5 & 9.490 \\
7 & 72.9 & 73.8 & \textbf{62.2} & \textbf{60.6} & \textbf{95.8} & \textbf{5.594} & \textbf{91.7} & \textbf{9.309} \\
\bottomrule
\end{tabular}%
}
\vspace{-10pt}
\caption{\textbf{Sequence Length Impact}. Models trained for 800 epochs with sequence lengths \( N\) = 4, 5, and 7, with \( N_p = 1 \).}
\label{tab:Ablation-SeqLen}
\vspace{-2pt}
\end{table}

%% file: sec/fig_qualitative_results.tex
{
\footnotesize
\centering
\newcommand{\resultsfig}[1]{\includegraphics[width=0.113\textwidth,valign=c]{#1}}
\setlength{\tabcolsep}{1pt}
\begin{tabular}{@{}ccccc@{}}

    & \mc{2}{\textbf{Scene Munster (0)}}
    & \mc{2}{\textbf{Scene  Munster (160)}} \\

    & Future Frame
    & Vista (Output)
    & Future Frame
    & Vista (Output)
    \\

    &
    \resultsfig{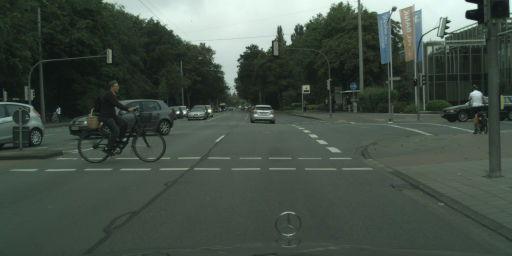} & 
    \resultsfig{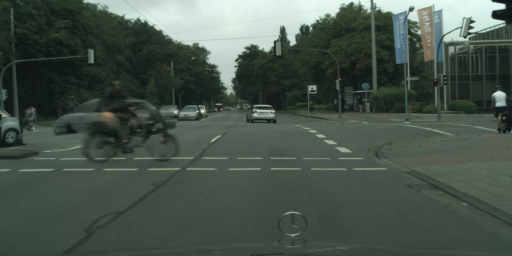} & 
    \resultsfig{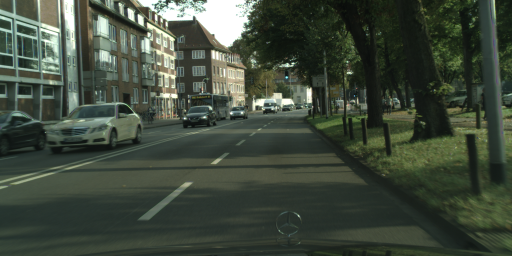} &
    \resultsfig{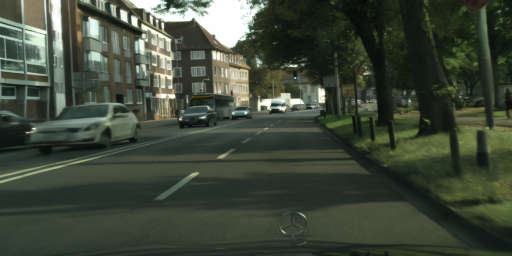} \\

    \mc{5}{\vspace{-1.6ex}}\\

    & Segmentation 
    & Depth
    & Segmentation  
    & Depth
    \\
    
    \raisebox{-.45\height}{\rotatebox{90}{\centering Oracle}}
    &
    \resultsfig{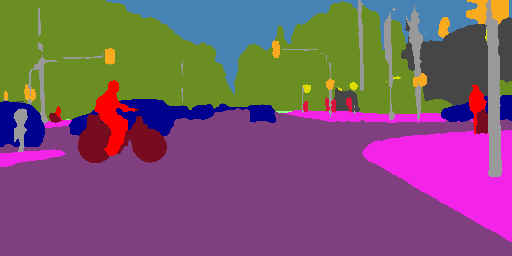} & 
    \resultsfig{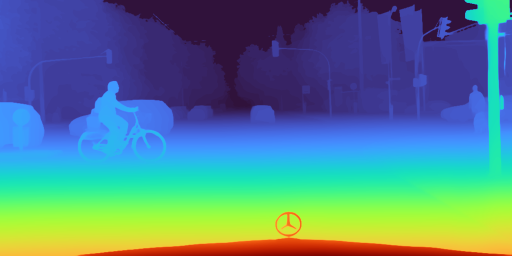} & 
    \resultsfig{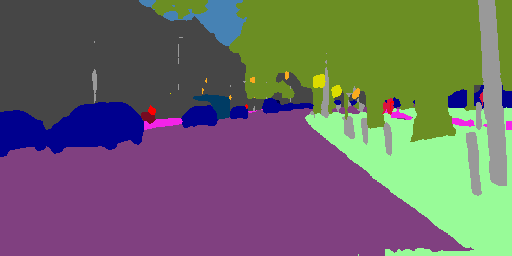} &
    \resultsfig{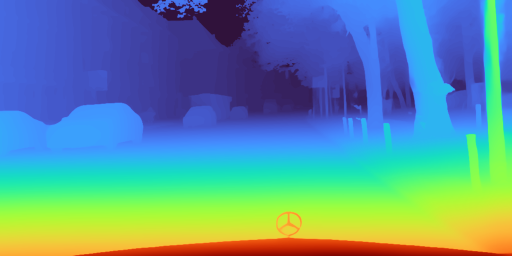} \\

    \mc{5}{\vspace{-2.1ex}}\\

    \raisebox{-.45\height}{\rotatebox{90}{\centering Vista}}
    &
    \resultsfig{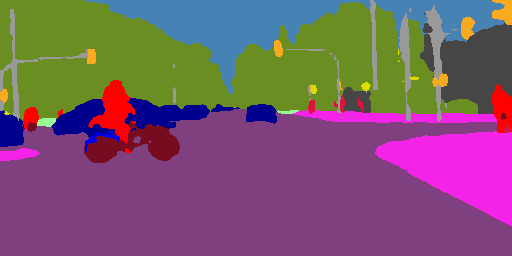} & 
    \resultsfig{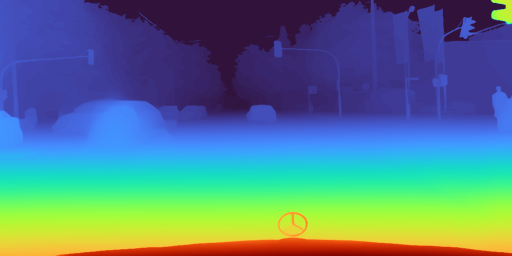} & 
    \resultsfig{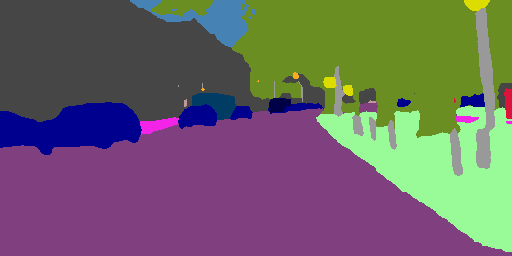} & 
    \resultsfig{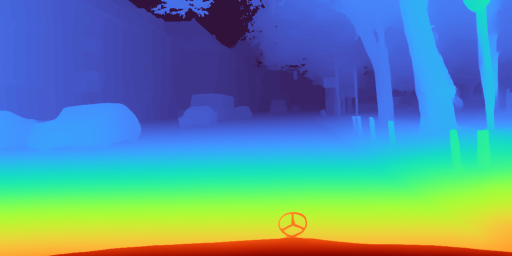}\\

    \mc{5}{\vspace{-2.1ex}}\\

    \raisebox{-.45\height}{\rotatebox{90}{\centering Ours}}
    &
    \resultsfig{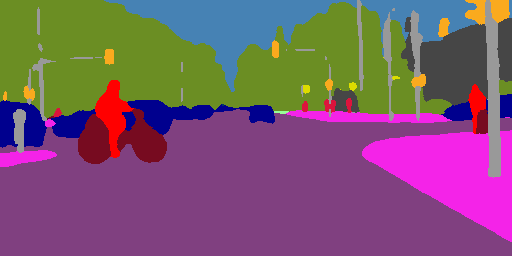} & 
    \resultsfig{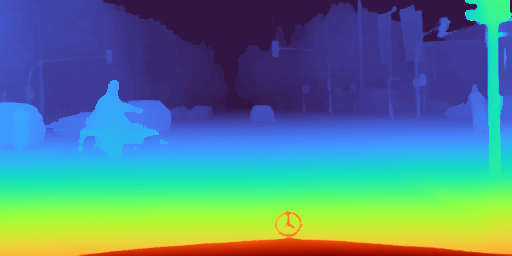} & 
    \resultsfig{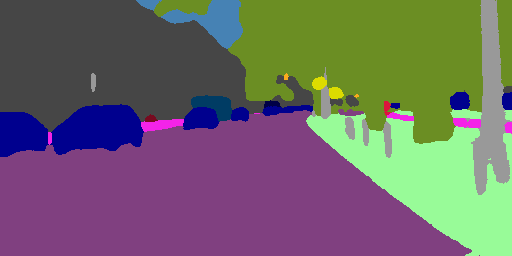} &
    \resultsfig{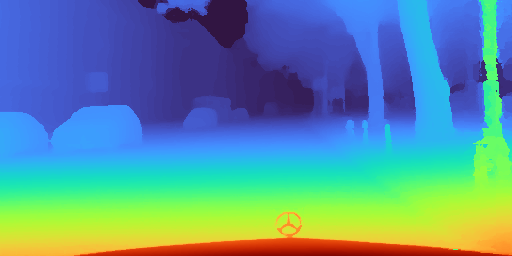} \\

    &
    \mc{4}{(a) Short-Term
    }\\

    \mc{5}{\vspace{-1.6ex}}\\

    & Future Frame
    & Vista (Output)
    & Future Frame
    & Vista (Output)
    \\

    &
    \resultsfig{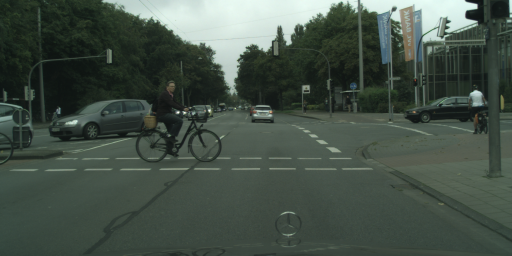} & 
    \resultsfig{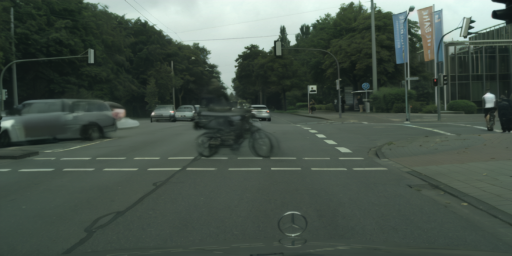} & 
    \resultsfig{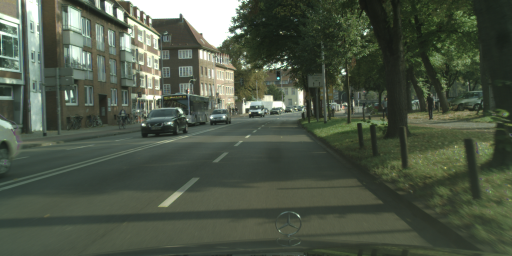} &
    \resultsfig{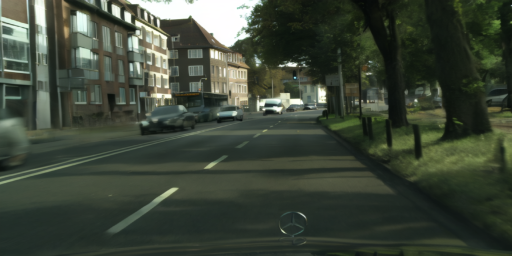} \\

    \mc{5}{\vspace{-1.6ex}}\\

    & Segmentation 
    & Depth
    & Segmentation  
    & Depth
    \\
    
    \raisebox{-.45\height}{\rotatebox{90}{\centering Oracle}}
    &
    \resultsfig{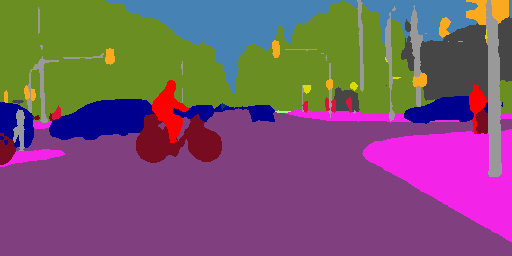} & 
    \resultsfig{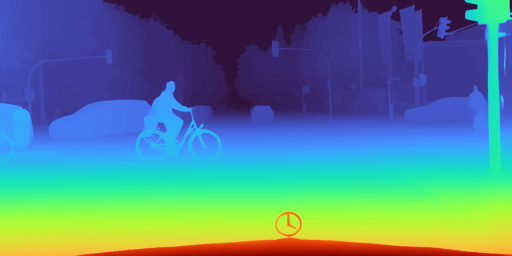} & 
    \resultsfig{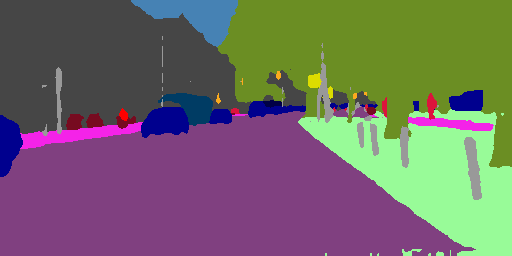} &
    \resultsfig{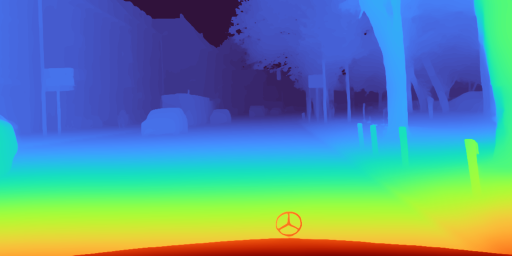} \\

    \mc{5}{\vspace{-1.6ex}}\\

    \raisebox{-.45\height}{\rotatebox{90}{\centering Vista}}
    &
    \resultsfig{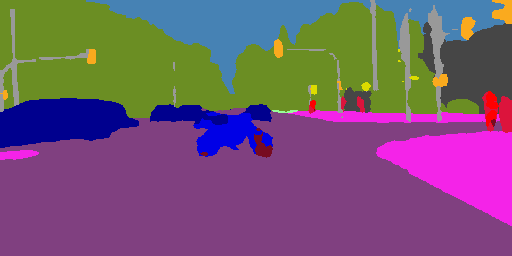} & 
    \resultsfig{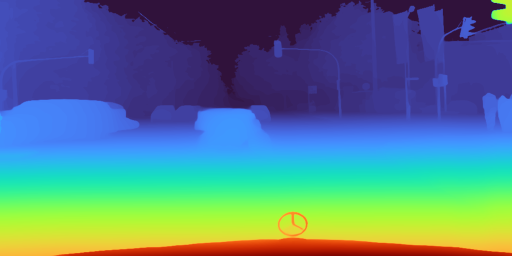} & 
    \resultsfig{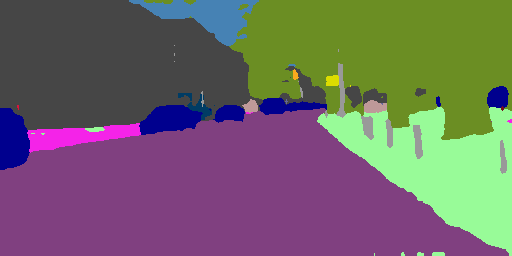} & 
    \resultsfig{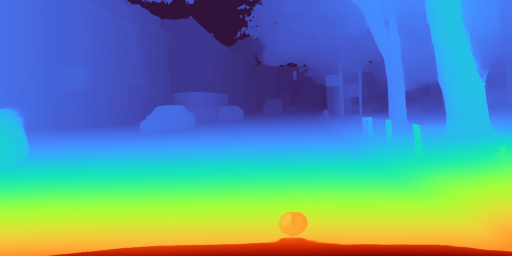}\\

    \mc{5}{\vspace{-1.6ex}}\\

    \raisebox{-.45\height}{\rotatebox{90}{\centering Ours}}
    &
\resultsfig{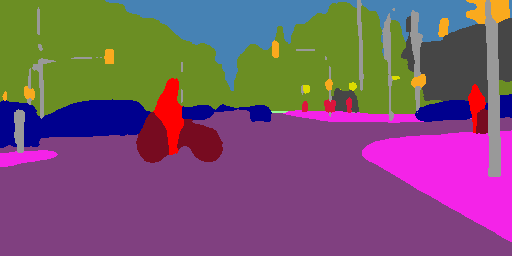} & 
    \resultsfig{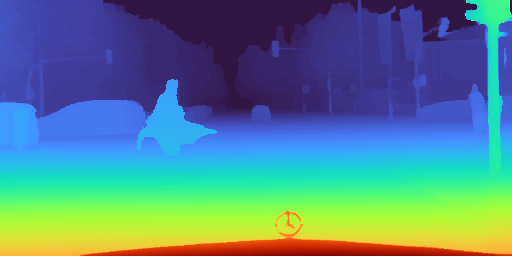} & 
    \resultsfig{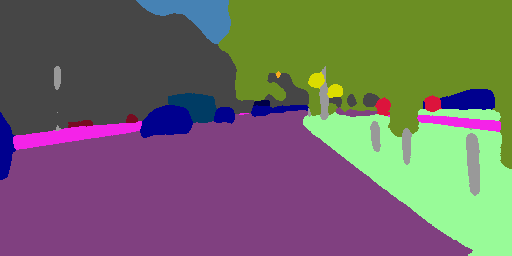} &
    \resultsfig{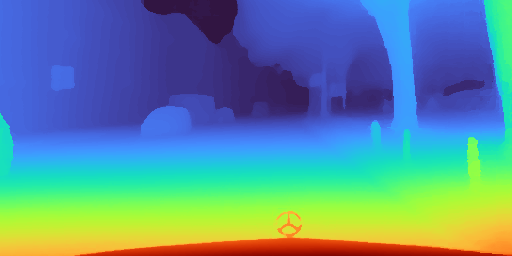} \\

    &
    \mc{4}{(b) Mid-Term}\\ 

\end{tabular}
}

%% file: sec/fig_qualitative_rollout.tex
{
\footnotesize
\centering
\newcommand{\resultsfignew}[1]{\includegraphics[width=0.11\textwidth,valign=c]{#1}}
\setlength{\tabcolsep}{1pt}
\begin{tabular}{@{}ccccc@{}}
    
    \multirow{3}{*}{\raisebox{-75pt}{\rotatebox{90}{Context Frames}}} &
    $X_{t-9}$ & $X_{t-6}$& $X_{t-3}$ & $X_{t}$ \\
    
    &
    \resultsfignew{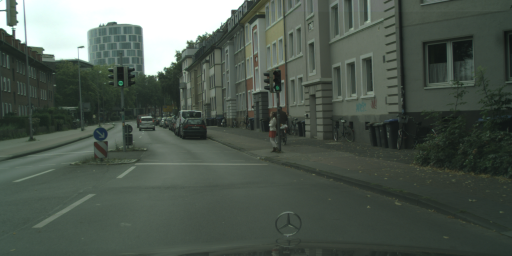} & 
    \resultsfignew{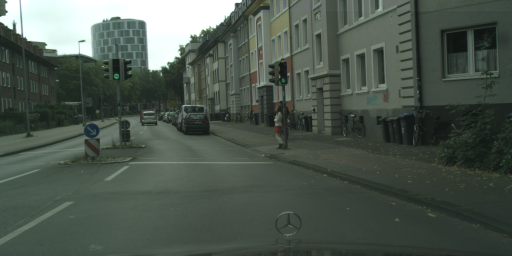} & 
    \resultsfignew{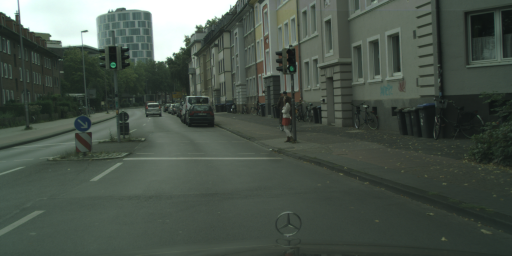} & 
    \resultsfignew{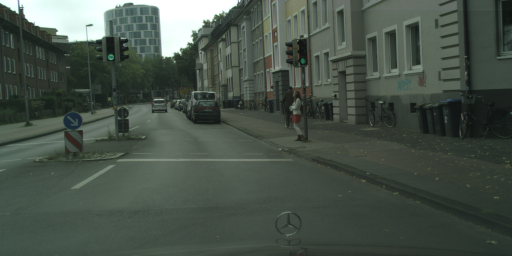}  \\

    \mc{5}{\vspace{-2.0ex}}\\
    &
    \resultsfignew{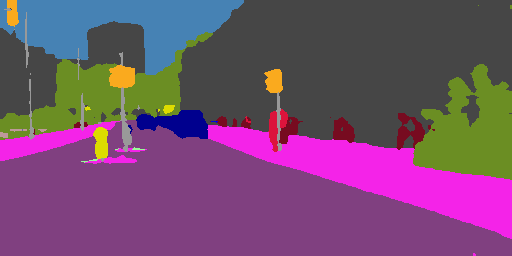} & 
    \resultsfignew{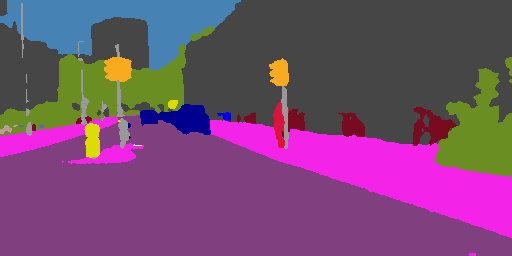} & 
    \resultsfignew{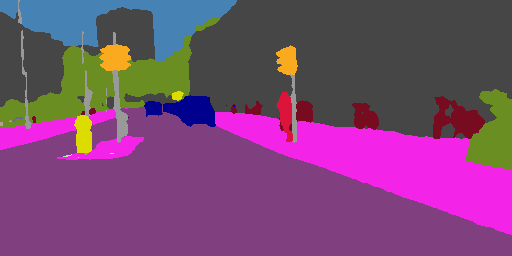} & 
    \resultsfignew{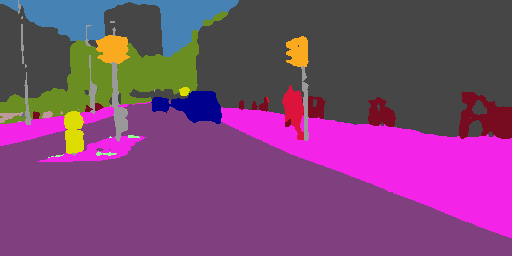}  \\

    \mc{5}{\vspace{-2.0ex}}\\

     &
    \resultsfignew{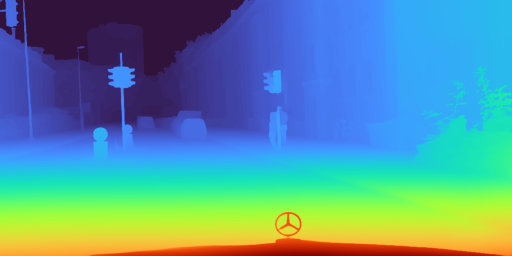} & 
    \resultsfignew{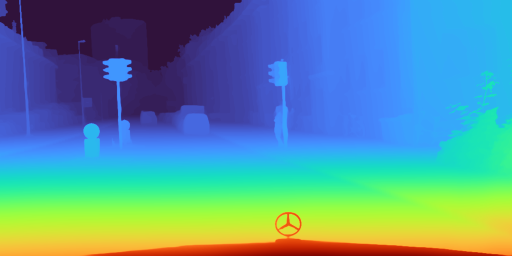} & 
    \resultsfignew{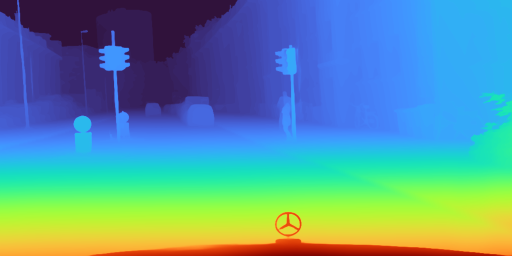} & 
    \resultsfignew{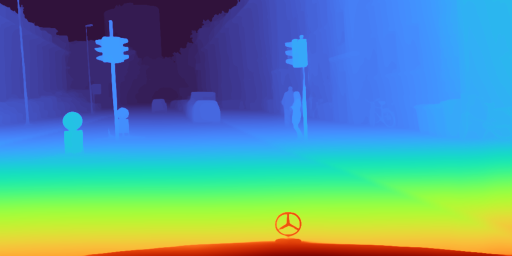}  \\

    \mc{5}{\vspace{-1.5ex}}\\

    \multirow{3}{*}{\raisebox{-60pt}{\rotatebox{90}{Predicted Frames}}}
    
    &
    $X_{t+3} (0.18s)$ & $X_{t+18} (1.08s)$& $X_{t+33} (1.98s)$ & $X_{t+48} (2.88s)$ \\

    &
    \resultsfignew{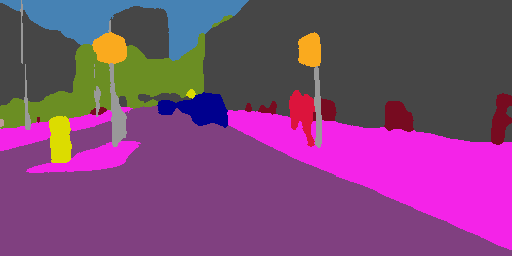} & 
    \resultsfignew{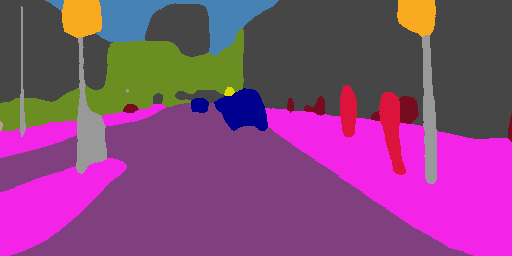} & 
    \resultsfignew{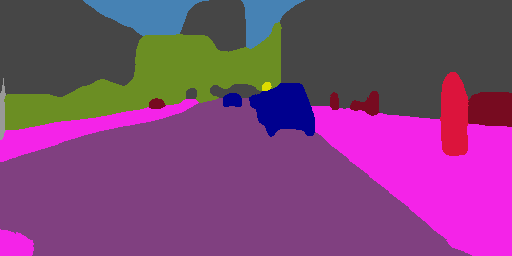} & 
    \resultsfignew{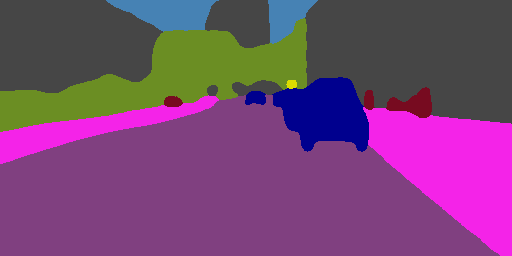}  \\

    \mc{5}{\vspace{-2.0ex}}\\
    &
\resultsfignew{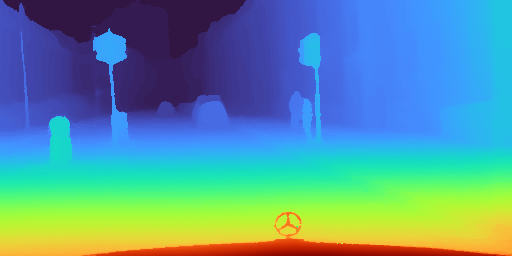} & 
    \resultsfignew{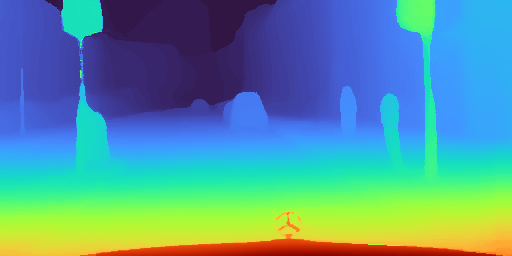} & 
    \resultsfignew{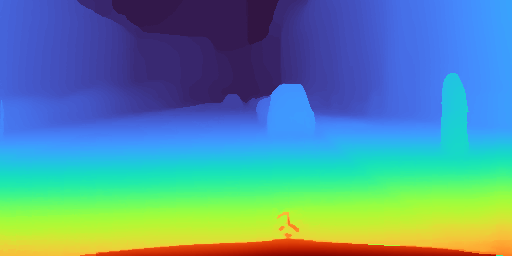} & 
    \resultsfignew{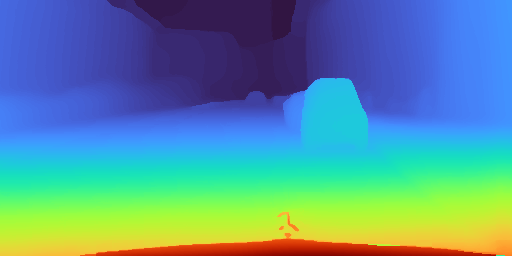}  \\
    
\end{tabular}
    }

%% file: sec/5_Conclusion.tex
\section{Conclusion}
\label{sec:conclusion}

This work advances semantic future prediction with \ours, a multimodal visual sequence transformer featuring: a VAE-free hierarchical tokenization strategy for efficient end-to-end multimodal training; a novel multimodal masked visual modeling objective; and an efficient cross-modality fusion mechanism that improves per-modality predictions by learning cross-modal synergies.

We evaluated \ours on the Cityscapes dataset for semantic segmentation and depth prediction. Our results show accurate short-term and mid-term predictions, outperforming single-modality models and setting a new state-of-the-art in future semantic segmentation. These findings demonstrate \ours's effectiveness in extending masked transformers for multimodal future prediction. We discuss the limitations of our method and directions for future work in \autoref{sec:limitations}.

%% file: sec/acknowledgements.tex
\paragraph{Acknowledgements}

This work has been partially supported by project MIS 5154714 of the National Recovery and Resilience Plan Greece 2.0 funded by the European Union under the NextGenerationEU Program. 

Hardware resources were granted with the support of GRNET. Also, this work was performed using HPC resources from GENCI-IDRIS (Grants 2023-A0141014182, 2023-AD011012884R2, and 2024-AD011012884R3).

%% file: sec/supplementary.tex
\section{Additional Results}
\label{sec:addresults}

\input{sec/supp_tab_modality_comparison}
\input{sec/supp_tab_vaes}
\input{sec/supp_tab_septokens_modal_1536}

\subsection{Comparison with VISTA}
\label{sec:vista_ft_comp}
In \autoref{tab:comparison_with_vista}, we compare our \ours approach against the VISTA~\cite{gao2024vista} baseline on both the segmentation and depth forecasting tasks. We conducted evaluations using a version fine-tuned on Cityscapes for 10 epochs to address potential performance issues arising from domain shift. LORA Finetuning required 8 GPUs, with a batch size of 1 per GPU, utilizing approximately $80\times8=640$GB of VRAM. VISTA finetuned still fell far behind \ours.

\subsection{Tokenization: Comparing Our VAE-Free Approach to VQ-VAE}

In \autoref{tab:comparison_with_vqvae}, we compare our approach to variations that replace the proposed VAE-free tokenization process (described in Sec. 3.1) with opensource large-scale pretrained discrete tokenizers such as VQ-VAE model from DALL-E~\cite{dalle2021} or LDM~\cite{rombach2022high}. For this comparison, we rendered the segmentation or depth modalities as RGB images, by applying the cityscapes colormap for segmentation and replicating values across three channels for depth and fed them into the VQ-VAE encoder for tokenization. The results show that using VQ-VAE leads to significantly worse performance compared to our VAE-free approach. 

The main reason for this gap is that the reconstruction process in VQ-VAE significantly degrades the oracle performance, particularly for the segmentation modality. In fact, segmentation results after VQ-VAE reconstruction are worse than our predicted segmentation results for short-term predictions. Moreover, we also fine-tuned LDM’s VQ-VAE \cite{rombach2022high} (since DALL-E does not offers training code and recipe) on segmentation and depth maps, improving both Oracle reconstruction (\Th{Oracle after LDM's VQ-VAE-FT Reconstruction}) and future segmentation prediction (\Th{VQ-VAE-FT from LDM}), though depth prediction performance slightly declined. Still, our VAE-free method is more effective and efficient: fewer parameters (465M vs 558M total), faster training (8h vs 22h at 800 epochs), and lower inference time per sequence (52ms vs 274ms).

The key takeaway is that while VQ-VAE tokenizers are essential for generative models focused on image or video generation, they are likely unnecessary for semantic modalities like those considered here. Our VAE-free approach not only simplifies the training pipeline but also achieves superior performance.

\subsection{Impact of Our Multi-Modal Fusion Strategy}

In \autoref{tab:sep-tokens-comparison}, we compare our approach to a variation that keeps tokens from the two modalities separate, instead of using our multi-modality early fusion strategy, which concatenates them along the embedding dimension. When trained for the same number of epochs, the separable tokens approach shows a slight improvement in segmentation performance, while our method performs better on most depth metrics. However, our approach is significantly more efficient, requiring half the training time and GPU memory. Furthermore, when trained for twice as many epochs—matching the total compute budget of the separable tokens approach—our method outperforms it in six out of eight metrics. This demonstrates that under the same compute budget, our approach delivers superior results.

\section{Additional Qualitative Results} \label{sec:visualizations}

\begin{figure*}[t!]
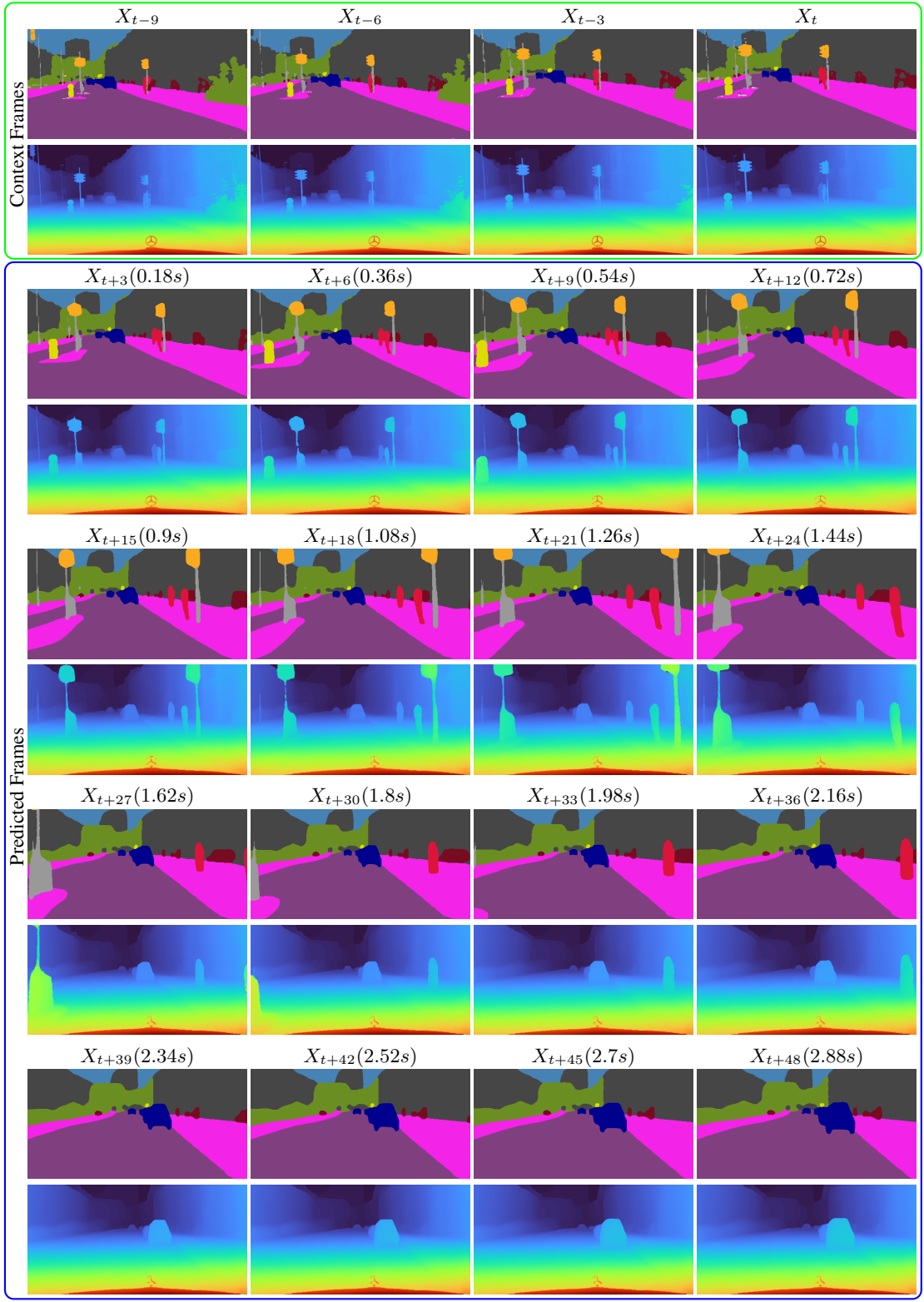

    \centering
    \include{sec/supp_fig_qualitative_rollout_munster_000023_000019}
    \vspace{-20pt}
    \caption{\textbf{Long-term semantic segmentation and depth predictions for Scene: Munster (23).
    } The model effectively captures temporal coherence in this complex scene with numerous static and moving objects.}
    \label{fig:qualitative-rollout-s1}
\end{figure*}

\begin{figure*}[t!]
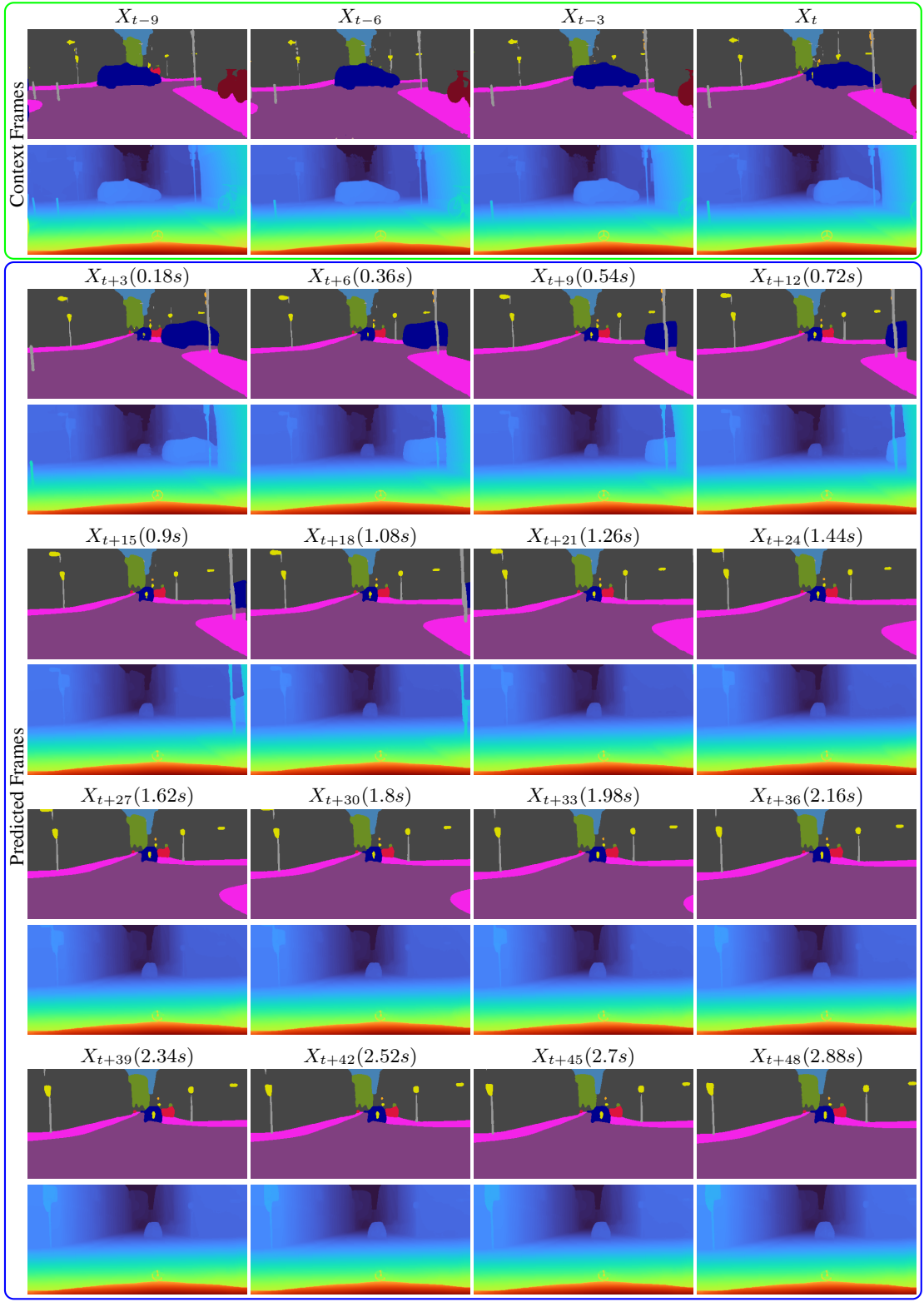

    \centering
    \include{sec/supp_fig_qualitative_rollout_frankfurt_000000_000294}
    \vspace{-20pt}
    \caption{\textbf{Long-term semantic segmentation and depth predictions for Scene: Frankfurt $\textbf{(0\_[275-304])}$.} Our approach accurately reflects the motion of static objects due to ego vehicle movement and predicts the motion of a car moving perpendicular to it.}
    \label{fig:qualitative-rollout-s2}
\end{figure*}

\begin{figure*}[t!]
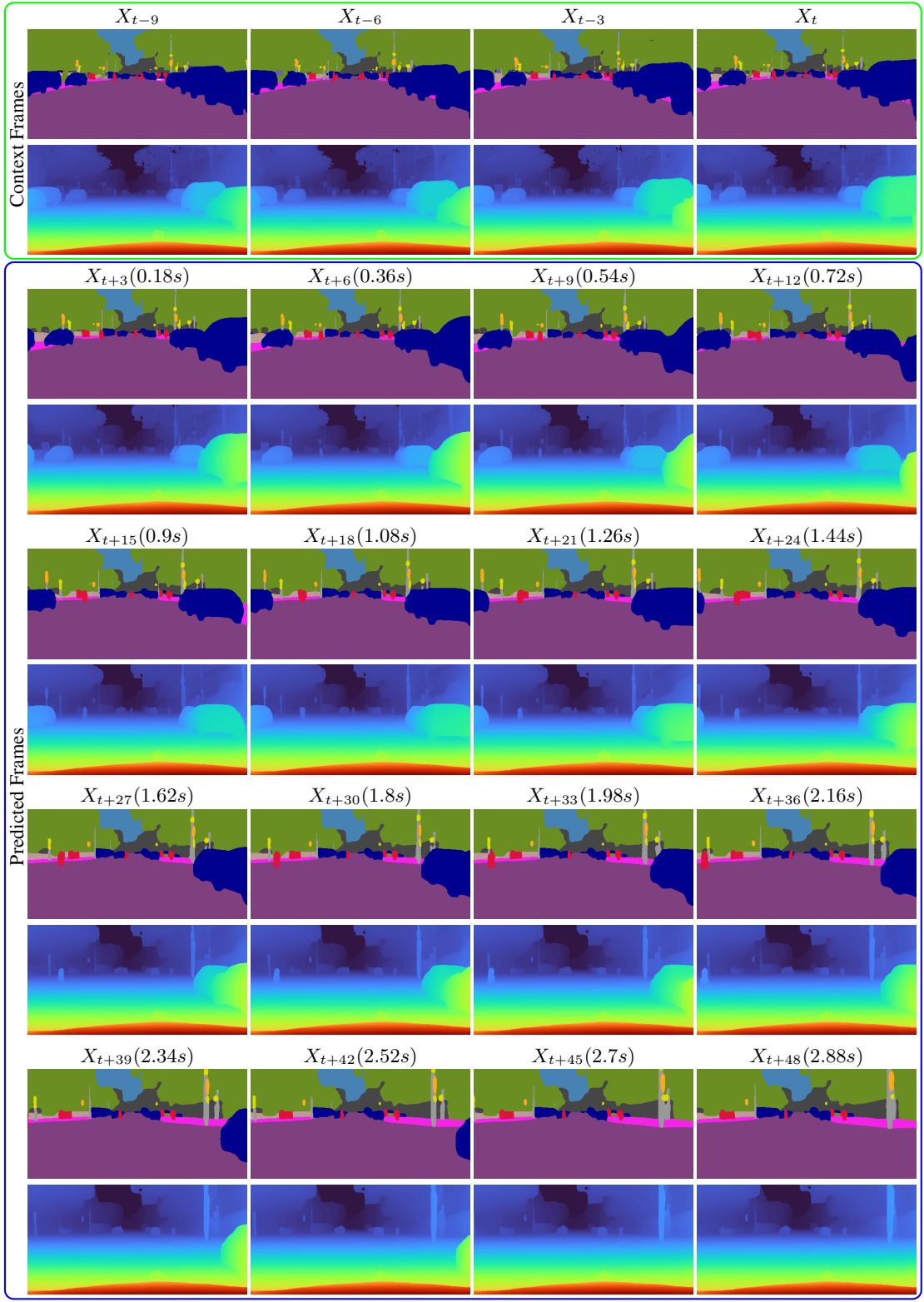

    \centering
    \include{sec/supp_fig_qualitative_rollout_frankfurt_000000_001236}
    \vspace{-20pt}
    \caption{\textbf{Long-term semantic segmentation and depth predictions for Scene: Frankfurt $\textbf{(0\_[1217-1246])}.$ 
    } Our model accurately preserves the relationships between static and dynamic elements and the motion of the ego vehicle.}
    \label{fig:qualitative-rollout-s3}
\end{figure*}

\begin{figure*}[t!]
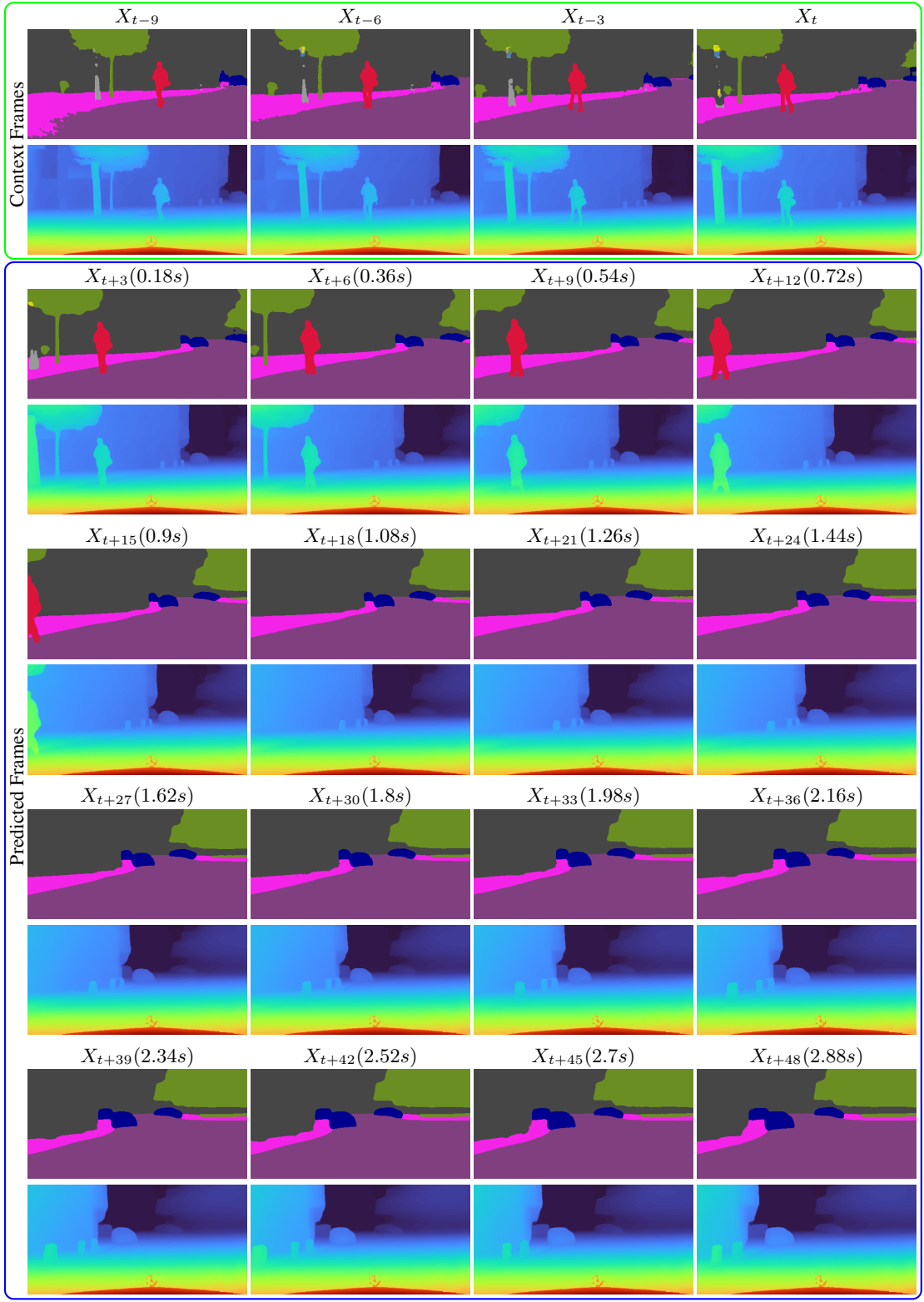

    \centering
    \include{sec/supp_fig_qualitative_rollout_lindau_000037_000019}
    \vspace{-20pt}
    \caption{\textbf{Long-term semantic segmentation and depth predictions for Scene: Lindau (37).
    } This figure demonstrates our model's ability to anticipate the completion of a right turn.}
    \label{fig:qualitative-rollout-s4}
\end{figure*}

\begin{figure*}[t!]
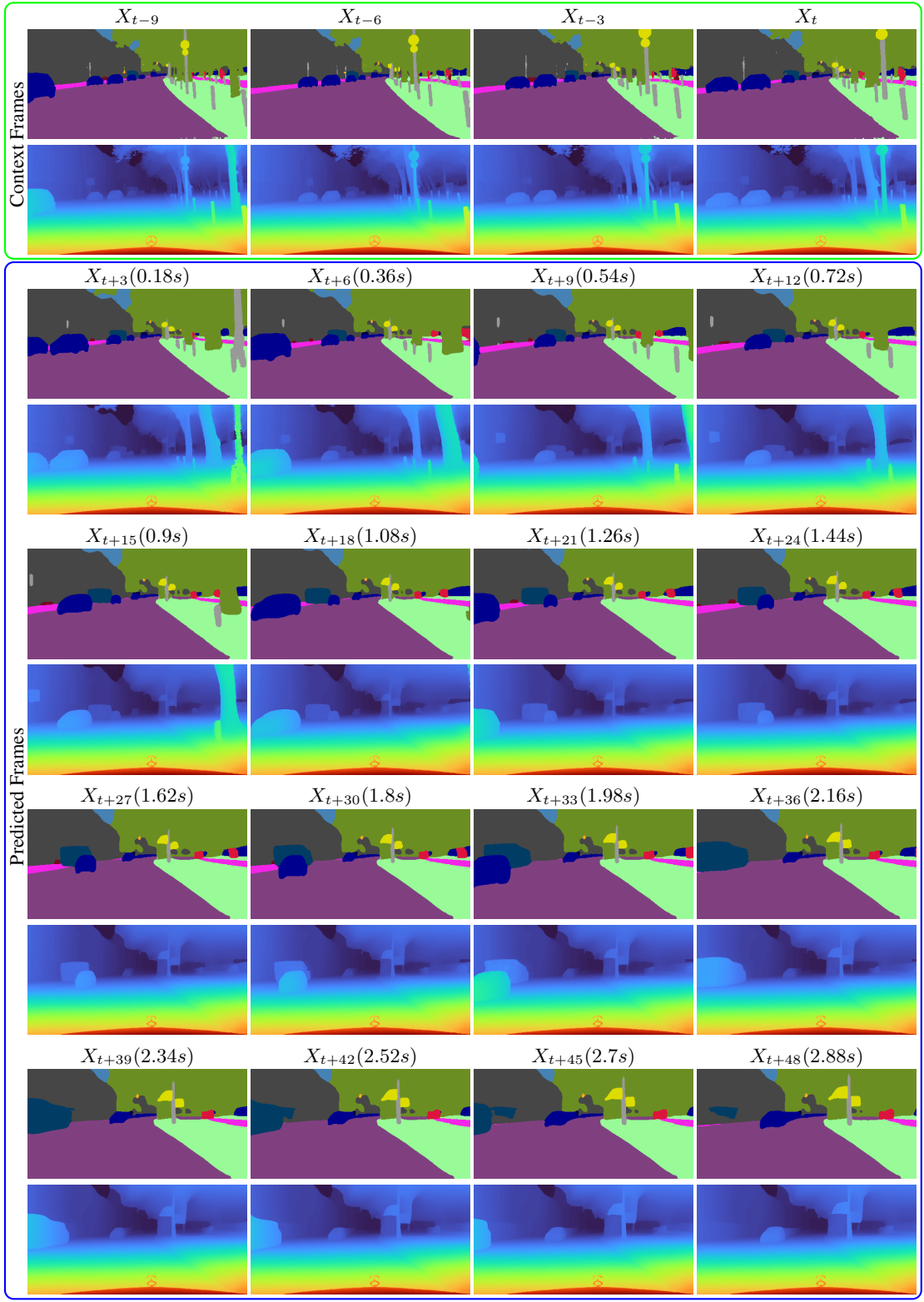

    \centering
    \include{sec/supp_fig_qualitative_rollout_munster_000160_000019}
    \vspace{-20pt}
    \caption{\textbf{Long-term semantic segmentation and depth predictions for Scene: Munster (160).
    } Despite precise short-term predictions, car masks become elongated towards the sequence's end, indicating a need for future adjustments.}
    \label{fig:qualitative-rollout-s5}
\end{figure*}

In ~\reffigures{fig:qualitative-rollout-s1}, \ref{fig:qualitative-rollout-s2}, \ref{fig:qualitative-rollout-s3}, \ref{fig:qualitative-rollout-s4} and \ref{fig:qualitative-rollout-s5}, 
we present additional qualitative results using \ours for forecasting semantic segmentation and depth maps over extended time horizons. 
These results were generated through autoregressive rollouts. 
Starting with a sequence of four context frames (\(X_{t-9}\) to \(X_t\)), the model predicts up to 48 future frames, corresponding to 2.88 seconds, with a frame interval of 3. 

The examples in ~\reffigures{fig:qualitative-rollout-s1}, \ref{fig:qualitative-rollout-s2}, \ref{fig:qualitative-rollout-s3}, and \ref{fig:qualitative-rollout-s4} demonstrate that the model effectively preserves temporal coherence, maintaining consistent relationships between static and dynamic elements over time. The depth maps transition smoothly between frames, aligning well with changes in the scene. For instance, \autoref{fig:qualitative-rollout-s1} and \autoref{fig:qualitative-rollout-s3} show complex scenes with numerous static and moving objects. Here, the model captures the ego vehicle's motion accurately, enabling precise predictions. In \autoref{fig:qualitative-rollout-s2}, the model successfully predicts the motion of a car crossing perpendicularly to the ego vehicle, while \autoref{fig:qualitative-rollout-s4} highlights the model's ability to anticipate the completion of a right turn.

However, as shown in \autoref{fig:qualitative-rollout-s5}, the quality of predictions degrades toward the end of the rollout, particularly in the final four frames. The car masks become elongated instead of approaching the ego vehicle as expected. This degradation likely stems from a mismatch between training and inference: during training, the model uses teacher forcing with oracle-provided context frames, whereas at inference, it relies on its own noisier predictions from previous steps. As discussed in \autoref{sec:limitations}, future work should aim to address this issue.

\section{Limitations and Future Work} \label{sec:limitations}

While \ours offers a simple and scalable approach to multi-modal semantic future prediction and demonstrates clear improvements over prior methods, several areas warrant further exploration to unlock its full potential.

Our work primarily focuses on short- and mid-term future predictions (0.18 seconds and 0.54 seconds, respectively), where our method excels. Although our model can be applied autoregressively for longer time horizons—generating mostly coherent predictions that capture both static and dynamic elements (as shown in \autoref{sec:visualizations})—further work is needed to improve robustness in these scenarios. This could involve addressing the challenges posed by noisy previous-step predictions (not currently considered due to teacher-forcing during training) or introducing stochasticity to better handle the inherent uncertainty of long-term predictions.

Another promising direction is extending \ours to instance and panoptic segmentation tasks. Our approach could be adapted by encoding instance information through pixel-wise offsets to instance centers \cite{PersonLab_pap, vip-deeplab, prisadnikov2024simple} rather than using instance IDs directly. This would allow each pixel within an instance mask to be represented by its x and y offsets to the corresponding instance center. These offsets could be processed as an additional modality with a dedicated embedding layer. During inference, instance masks could be recovered from these predicted offsets using a Hough transform-like approach and combined with semantic predictions to generate panoptic segmentation. This extension would enhance the applicability of our method to scenarios requiring instance-level understanding.

Currently, our method does not include action conditioning, which limits controllability. Incorporating sequences of control actions as an additional data modality would transform \ours into a world model capable of predicting outcomes based on specific actions. This could increase its utility in autonomous driving and open up opportunities for applications in robotics and embodied AI.

Finally, we have not fully explored the scaling behavior of our approach, including model size, training data, and the number of modalities. However, even with training limited to Cityscapes, our approach delivers strong results, highlighting its significant potential for scalability and broader generalization in future work. Additionally, our approach relies on pre-trained perception models to produce semantic modalities (e.g., segmentation and depth) from RGB images. While this dependency requires robust models for diverse scenes, the rapid advancements in foundational perception models and their increasing availability through open releases ensure that our method remains adaptable and continues to benefit from progress in the broader computer vision community. Moreover, operating on semantic modalities rather than RGB images makes it easier to exploit synthetic data generated by simulators, as these data do not need to be photorealistic.

%% file: sec/supp_tab_modality_comparison.tex
\begin{table*}[!h]
\small
\centering
\setlength{\tabcolsep}{3.0pt}
\begin{tabular}{lcccccccc}
\toprule
& \multicolumn{4}{c}{\Th{Segmentation}} & \multicolumn{4}{c}{\Th{Depth}} \\
\cmidrule(r){2-5} \cmidrule(l){6-9}
\mr{2}{\Th{Method}} & \multicolumn{2}{c}{\Th{Short-term}} & \multicolumn{2}{c}{\Th{Mid-term}} & \multicolumn{2}{c}{\Th{Short-term}} & \multicolumn{2}{c}{\Th{Mid-term}} \\
\cmidrule(r{0.3em}){2-3} \cmidrule(l{0.3em}r){4-5} \cmidrule(lr{0.3em}){6-7} \cmidrule(l{0.3em}){8-9}
& \Th{ALL}$\uparrow$ & \Th{MO}$\uparrow$ & \Th{ALL}$\uparrow$ & \Th{MO}$\uparrow$ & $\delta_{1}$$\uparrow$ & \Th{AbsRel}$\downarrow$ & $\delta_{1}$$\uparrow$ & \Th{AbsRel}$\downarrow$ \\
\midrule
\Th{Copy Last} & 55.5 & 52.7 & 40.5 & 32.2 & 90.5 & 10.780 & 82.2 & 18.345 \\

\Th{VISTA Fine-tuned} & 65.7 & 64.1 & 53.2 & 49.1 & 93.4 & \,\,\,8.134 & 87.4 & 12.923 \\

\ours & \textbf{73.9} & \textbf{74.9} & \textbf{62.7} & \textbf{61.2} & \textbf{96.0} & \textbf{\,\,\,5.384} & \textbf{91.9} & \textbf{\,\,\,9.111} \\

\bottomrule
\end{tabular}%
\caption{\textbf{Comparison with VISTA on semantic segmentation and depth forecasting.} Our \ours model was trained for 3200 epochs. \Th{AbsRel} is multiplied by 100 for readability.}
\label{tab:comparison_with_vista}
\end{table*}

%% file: sec/supp_tab_vaes.tex
\begin{table*}[!h]
\small
\centering
\setlength{\tabcolsep}{4pt}
\begin{tabular}{lcccccccc}
\toprule
& \multicolumn{4}{c}{\Th{Segmentation}} & \multicolumn{4}{c}{\Th{Depth}} \\
\cmidrule(r){2-5} \cmidrule(l){6-9}
\mr{2}{\Th{Tokenizer}} & \multicolumn{2}{c}{\Th{Short-term}} & \multicolumn{2}{c}{\Th{Mid-term}} & \multicolumn{2}{c}{\Th{Short-term}} & \multicolumn{2}{c}{\Th{Mid-term}} \\
\cmidrule(r{0.3em}){2-3} \cmidrule(l{0.3em}r){4-5} \cmidrule(lr{0.3em}){6-7} \cmidrule(l{0.3em}){8-9}
& \Th{ALL}$\uparrow$ & \Th{MO}$\uparrow$ & \Th{ALL}$\uparrow$ & \Th{MO}$\uparrow$ & $\delta_{1}$$\uparrow$ & \Th{AbsRel}$\downarrow$ & $\delta_{1}$$\uparrow$ & \Th{AbsRel}$\downarrow$ \\
\midrule
\Th{Oracle}~\cite{strudel2021segmenter} & 78.6 & 80.8 & 78.6 & 80.6 & - & - & - & - \\
\Th{Oracle after DALL-E's VQ-VAE Reconstruction} & 71.5 & 72.2 & 71.5 & 72.2 & 98.6 & 2.263 & 98.6 & \,\,\,2.263 \\
\Th{Oracle after LDM's VQ-VAE Reconstruction} & 72.1 & 71.9 & 72.1 & 71.9 & 97.3 & 4.665 & 97.3 & \,\,\,4.665 \\
\Th{Oracle after LDM's VQ-VAE-FT Reconstruction} & 77.1 & 78.9 & 77.1 & 78.9 & 98.7 & 2.776 & 98.7 & \,\,\, 2.776  \\
\midrule
\Th{VQ-VAE from DALL-E}~\cite{dalle2021} & 65.0 & 62.8 & 54.4 & 48.4 & 94.5 & 6.643 & 88.3 & 10.855 \\
\Th{VQ-VAE from LDM}~\cite{rombach2022high} & 60.7 & 55.7 & 51.5 & 44.6 & 91.8 & 9.032 & 84.9 & 13.804 \\
\Th{VQ-VAE-FT from LDM} & 69.2 & 68.9 & 57.2 & 53.1 & 90.7 & 8.912 & 82.2 &  14.068 \\ 
Our VAE-free hierarchical tokenization & \textbf{72.9} & \textbf{73.8} & \textbf{61.6} & \textbf{59.9} & \textbf{95.8} & \textbf{5.606} & \textbf{91.5} & \textbf{\,\,\,9.490} \\

\bottomrule
\end{tabular}%
\caption{\textbf{Tokenization: comparing our VAE-Free approach to VQ-VAE.} 
All models are trained for 800 epochs. \Th{AbsRel} is multiplied by 100 for readability. 
We do not report results for the Oracle baseline in depth forecasting because, unlike segmentation where metrics are based on ground truth from the dataset, there is no ground truth for depth. Instead, we use the Oracle (DepthAnything \cite{yang2024depth}) to generate pseudo-ground truth for comparison. As a result, the Oracle is expected to achieve 100\% accuracy and 0 error in depth prediction.}
\label{tab:comparison_with_vqvae}
\end{table*}

%% file: sec/supp_tab_septokens_modal_1536.tex
\begin{table*}[!h]
\small
\centering
\setlength{\tabcolsep}{3.0pt}
\begin{tabular}{lccccccccc}
\toprule
& \multicolumn{4}{c}{\Th{Segmentation}} & \multicolumn{4}{c}{\Th{Depth}} & \multicolumn{1}{c}{\Th{Training}} \\
\cmidrule(r){2-5} \cmidrule(l){6-9} 
\mr{2}{\Th{Approach}} & \multicolumn{2}{c}{\Th{Short-term}} & \multicolumn{2}{c}{\Th{Mid-term}} & \multicolumn{2}{c}{\Th{Short-term}} & \multicolumn{2}{c}{\Th{Mid-term}} & \Th{Time} \\
\cmidrule(r{0.3em}){2-3} \cmidrule(l{0.3em}r){4-5} \cmidrule(lr{0.3em}){6-7} \cmidrule(l{0.3em}r){8-9} \cmidrule(l{0.3em}r){10-10}
& \Th{ALL}$\uparrow$ & \Th{MO}$\uparrow$ & \Th{ALL}$\uparrow$ & \Th{MO}$\uparrow$ & $\delta_{1}$$\uparrow$ & \Th{AbsRel}$\downarrow$ & $\delta_{1}$$\uparrow$ & \Th{AbsRel}$\downarrow$ & \Th{Hours}$\downarrow$ \\
\midrule
\Th{Separate Tokens -- 800 epochs} & 73.3 & 74.3 & \textbf{62.2} & \textbf{60.6} & 95.8 & 5.622 & 91.5 & 9.442 & 18\\
\Th{Our multi-modal fusion -- 800 epochs} & 72.9 & 73.8 & 61.6 & 59.9 & 95.8 & 5.606 & 91.5 & 9.490 & \,\,\,9   \\
\Th{Our multi-modal fusion -- 1600 epochs} & \textbf{73.4} & \textbf{74.4} & 62.1 & 60.4 & \textbf{95.9} & \textbf{5.444} & \textbf{91.7} & \textbf{9.092} & 18 \\
\bottomrule
\end{tabular}%
\caption{\textbf{Impact of our multi-modal token fusion strategy.} \Th{AbsRel} is multiplied by 100 for readability. Training time is computed in hours.}
\label{tab:sep-tokens-comparison}
\end{table*}

%% file: sec/supp_fig_qualitative_rollout_munster_000023_000019.tex
{
\small
\centering
\newcommand{\resultsfignew}[1]{\includegraphics[width=0.2\textwidth,valign=c]{#1}}
\setlength{\tabcolsep}{1pt}

\pgfsetlayers{background,main}

\begin{tikzpicture}
    \node (table) {
\begin{tabular}{@{}ccccc@{}}
    
    \multirow{3}{*}{\raisebox{-84pt}{\rotatebox{90}{Context Frames}}} &
    $X_{t-9}$ & $X_{t-6}$& $X_{t-3}$ & $X_{t}$ \\

    &
    \resultsfignew{Figs/qualitative_results/rollout/munster_000023_000007_leftImg8bit.png} & 
    \resultsfignew{Figs/qualitative_results/rollout/munster_000023_000010_leftImg8bit.png} & 
    \resultsfignew{Figs/qualitative_results/rollout/munster_000023_000013_leftImg8bit.png} & 
    \resultsfignew{Figs/qualitative_results/rollout/munster_000023_000016_leftImg8bit.png}  \\

    \mc{5}{\vspace{-2.0ex}}\\

     &
    \resultsfignew{Figs/qualitative_results/rollout/munster_000023_000007_leftImg8bit_depth_colored.png} & 
    \resultsfignew{Figs/qualitative_results/rollout/munster_000023_000010_leftImg8bit_depth_colored.png} & 
    \resultsfignew{Figs/qualitative_results/rollout/munster_000023_000013_leftImg8bit_depth_colored.png} & 
    \resultsfignew{Figs/qualitative_results/rollout/munster_000023_000016_leftImg8bit_depth_colored.png}  \\

    \mc{5}{\vspace{-1.5ex}}\\
    
    &
    $X_{t+3} (0.18s)$ & $X_{t+6} (0.36s)$& $X_{t+9} (0.54s)$ & $X_{t+12} (0.72s)$ \\

    &
    \resultsfignew{Figs/qualitative_results/rollout/munster_000023_000019_pred_segm_tplus3.png} & 
    \resultsfignew{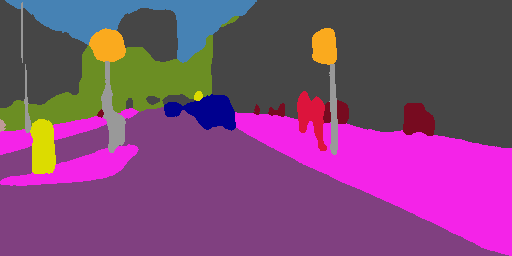} & 
    \resultsfignew{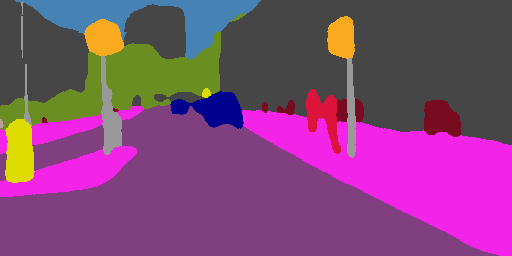} & 
    \resultsfignew{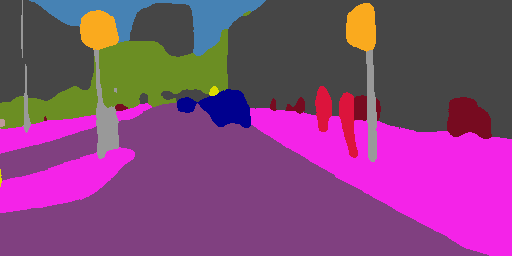}  \\

    \mc{5}{\vspace{-2.0ex}}\\
    &
\resultsfignew{Figs/qualitative_results/rollout/munster_000023_000019_pred_depth_tplus3.png} & 
    \resultsfignew{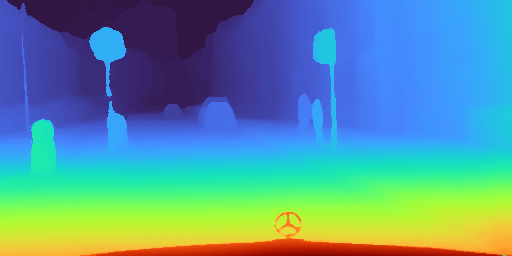} & 
    \resultsfignew{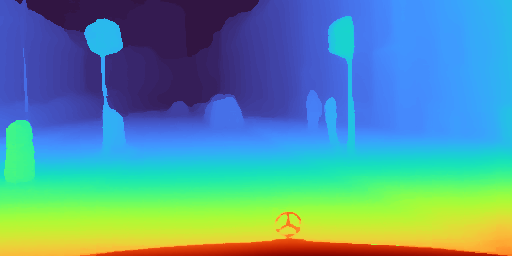} & 
    \resultsfignew{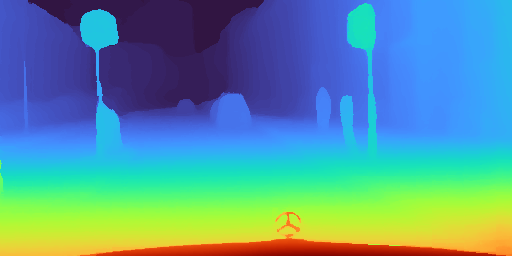}  \\

    \mc{5}{\vspace{-1.5ex}}\\
    
    \multirow{3}{*}{\raisebox{-147pt}{\rotatebox{90}{Predicted Frames}}} &
    $X_{t+15} (0.9s)$ & $X_{t+18}(1.08s)$& $X_{t+21} (1.26s)$ & $X_{t+24} (1.44s)$ \\
       &
    \resultsfignew{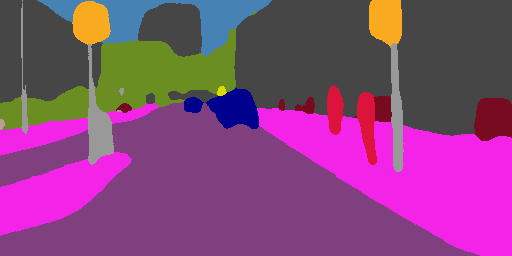} & 
    \resultsfignew{Figs/qualitative_results/rollout/munster_000023_000019_pred_segm_tplus18.png} & 
    \resultsfignew{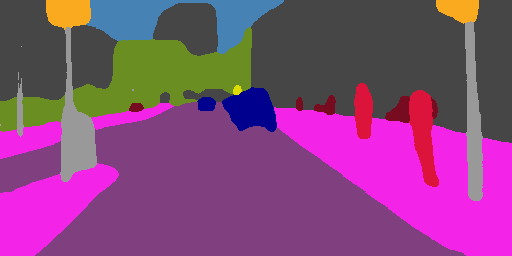} & 
    \resultsfignew{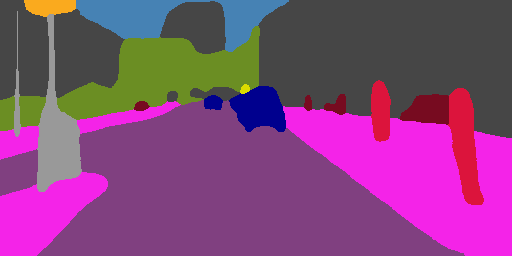}  \\

    \mc{5}{\vspace{-2.0ex}}\\
    &
\resultsfignew{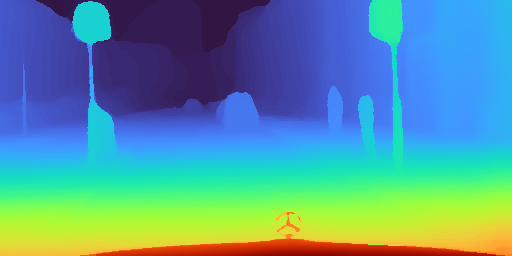} & 
    \resultsfignew{Figs/qualitative_results/rollout/munster_000023_000019_pred_depth_tplus18.png} & 
    \resultsfignew{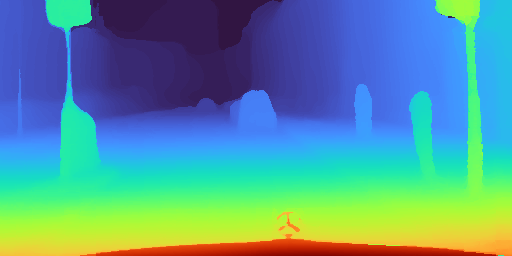} & 
    \resultsfignew{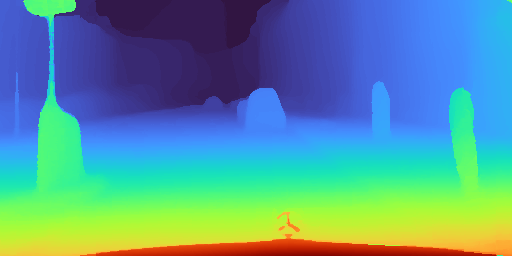}  \\

    \mc{5}{\vspace{-1.5ex}}\\
    &
     $X_{t+27}  (1.62s)$ & $X_{t+30}  (1.8s)$& $X_{t+33}  (1.98s)$ & $X_{t+36}  (2.16s)$ \\

    &
    \resultsfignew{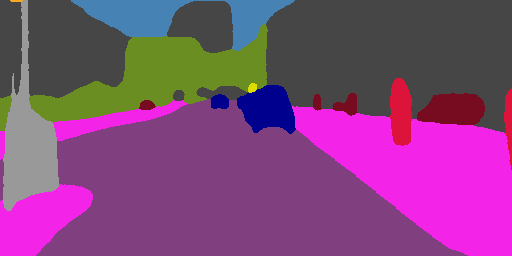} & 
    \resultsfignew{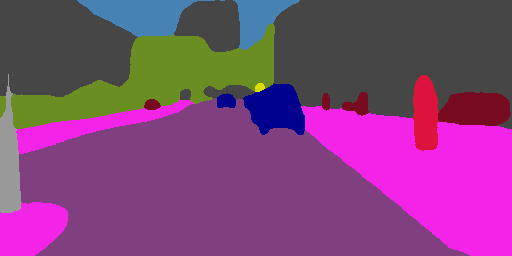} & 
    \resultsfignew{Figs/qualitative_results/rollout/munster_000023_000019_pred_segm_tplus33.png} & 
    \resultsfignew{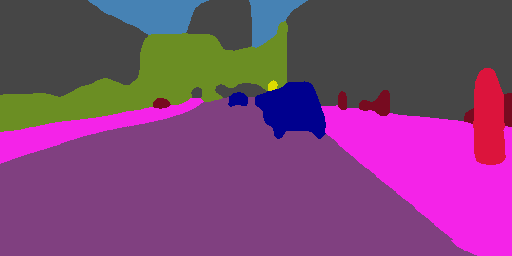}  \\

    \mc{5}{\vspace{-2.0ex}}\\
    
    &
    \resultsfignew{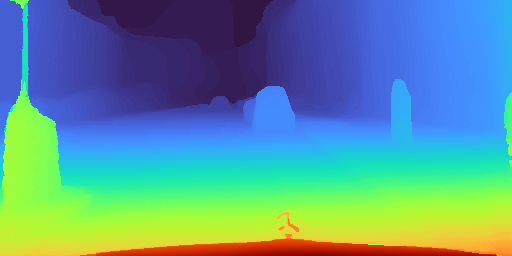} & 
    \resultsfignew{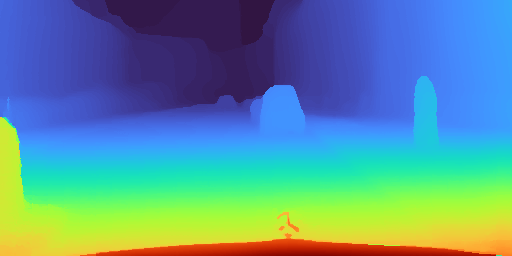} & 
    \resultsfignew{Figs/qualitative_results/rollout/munster_000023_000019_pred_depth_tplus33.png} & 
    \resultsfignew{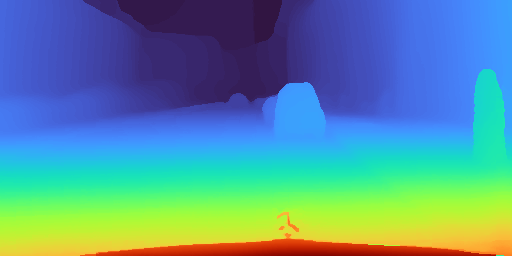}  \\

    \mc{5}{\vspace{-1.5ex}}\\
    &
     $X_{t+39} (2.34s)$ & $X_{t+42} (2.52s)$& $X_{t+45} (2.7s)$ & $X_{t+48} (2.88s)$ \\

     &
\resultsfignew{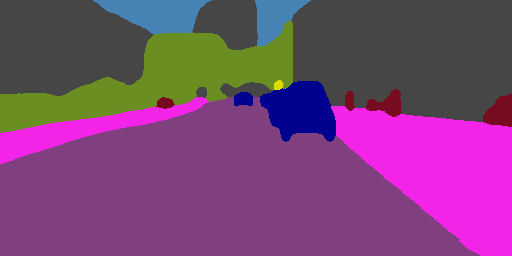} & 
    \resultsfignew{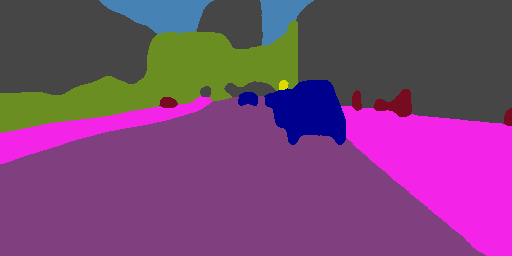} & 
    \resultsfignew{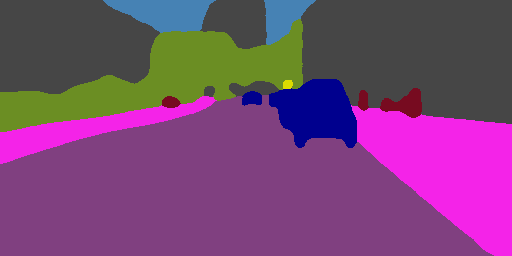} &
    \resultsfignew{Figs/qualitative_results/rollout/munster_000023_000019_pred_segm_tplus48.png}\\

    \mc{5}{\vspace{-2.0ex}}\\
     &
\resultsfignew{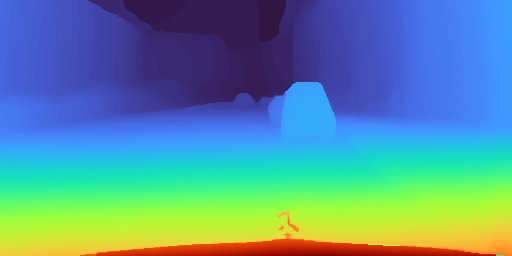} & 
    \resultsfignew{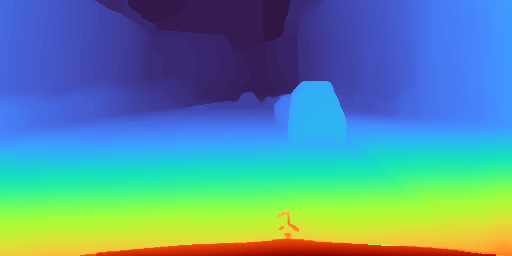} & 
    \resultsfignew{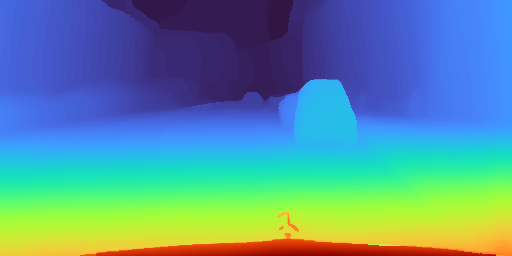} &
    \resultsfignew{Figs/qualitative_results/rollout/munster_000023_000019_pred_depth_tplus48.png}\\
    
\end{tabular}
    };

    \begin{scope}[on background layer]
        \draw[green, thick, rounded corners] 
            ([xshift=1pt, yshift=-2pt]table.north west) 
            rectangle ([xshift=-3pt, yshift=474pt]table.south east);

        \draw[blue, thick, rounded corners] 
            ([xshift=1pt, yshift=-120pt]table.north west) 
            rectangle ([xshift=-3pt, yshift=1pt]table.south east);
    \end{scope}

\end{tikzpicture}
}

%% file: sec/supp_fig_qualitative_rollout_frankfurt_000000_000294.tex
{
\small
\centering
\newcommand{\resultsfignew}[1]{\includegraphics[width=0.2\textwidth,valign=c]{#1}}
\setlength{\tabcolsep}{1pt}

\pgfsetlayers{background,main}

\begin{tikzpicture}
    \node (table) {
\begin{tabular}{@{}ccccc@{}}
    
    \multirow{3}{*}{\raisebox{-84pt}{\rotatebox{90}{Context Frames}}} &
    $X_{t-9}$ & $X_{t-6}$& $X_{t-3}$ & $X_{t}$ \\

    &
    \resultsfignew{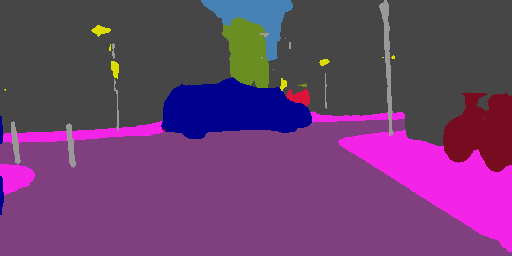} & 
    \resultsfignew{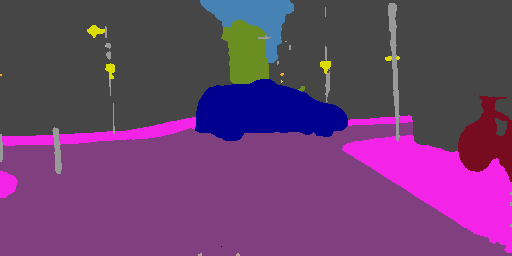} & 
    \resultsfignew{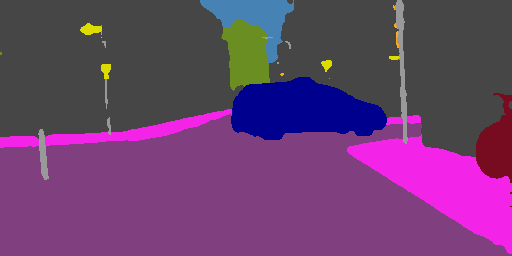} & 
    \resultsfignew{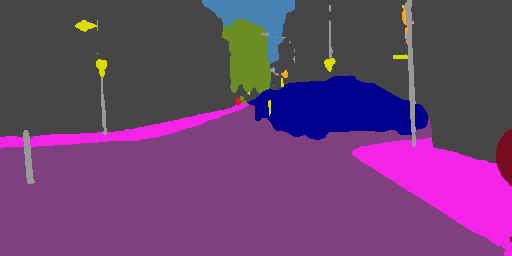}  \\

    \mc{5}{\vspace{-2.0ex}}\\

     &
    \resultsfignew{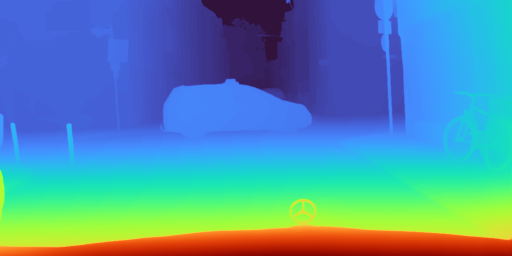} & 
    \resultsfignew{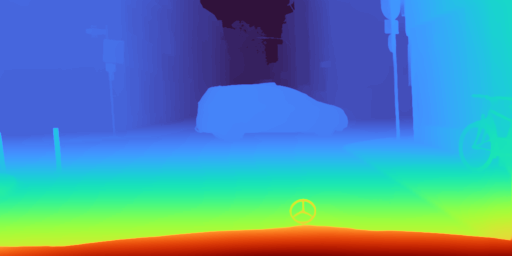} & 
    \resultsfignew{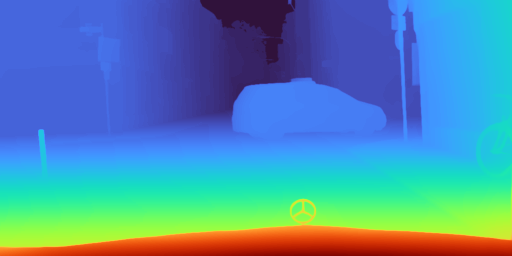} & 
    \resultsfignew{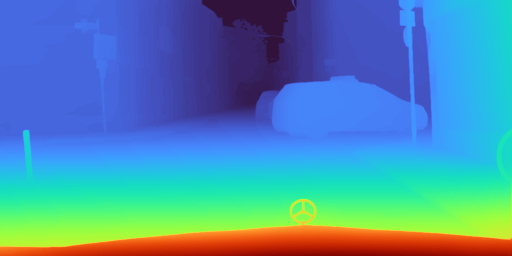}  \\

    \mc{5}{\vspace{-1.5ex}}\\
    
    &
    $X_{t+3} (0.18s)$ & $X_{t+6} (0.36s)$& $X_{t+9} (0.54s)$ & $X_{t+12} (0.72s)$ \\

    &
    \resultsfignew{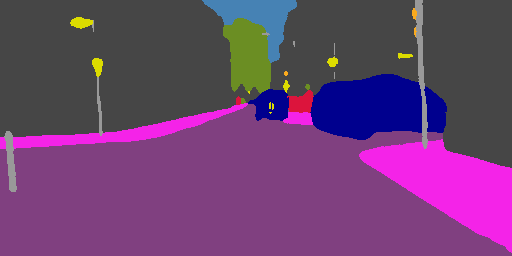} & 
    \resultsfignew{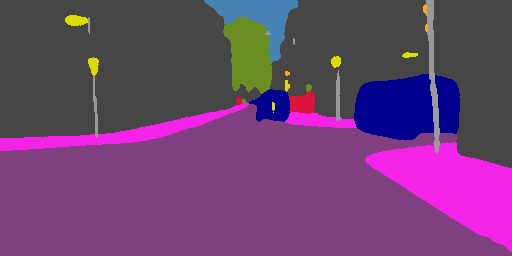} & 
    \resultsfignew{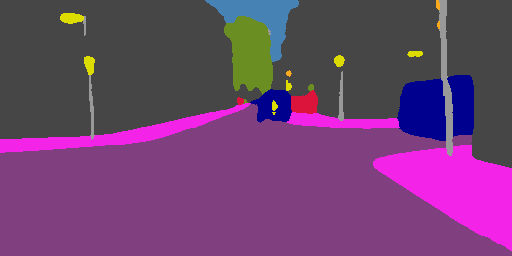} & 
    \resultsfignew{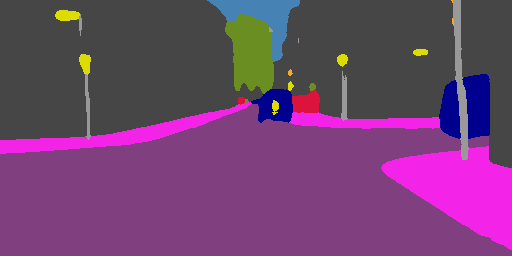}  \\

    \mc{5}{\vspace{-2.0ex}}\\
    &
    \resultsfignew{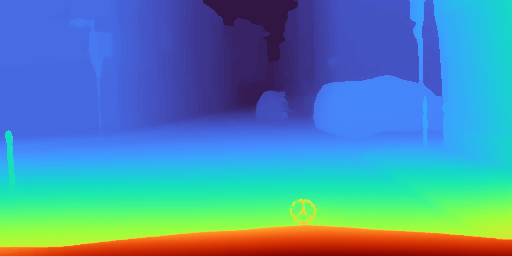} & 
    \resultsfignew{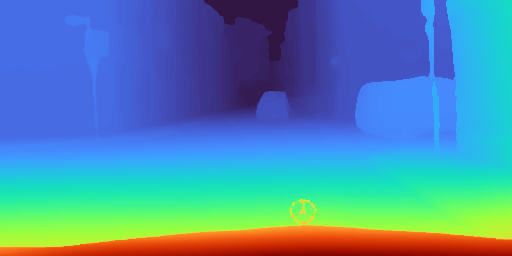} & 
    \resultsfignew{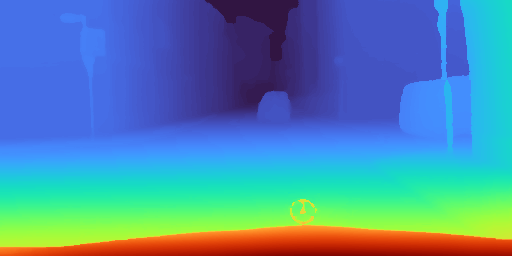} & 
    \resultsfignew{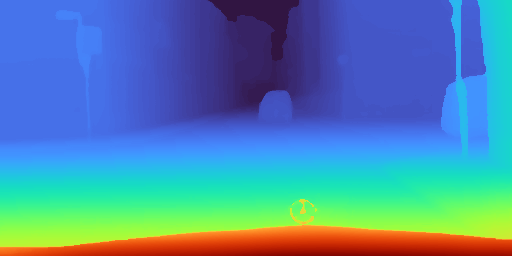}   \\

    \mc{5}{\vspace{-1.5ex}}\\
    
    \multirow{3}{*}{\raisebox{-147pt}{\rotatebox{90}{Predicted Frames}}} &
    $X_{t+15} (0.9s)$ & $X_{t+18}(1.08s)$& $X_{t+21} (1.26s)$ & $X_{t+24} (1.44s)$ \\
       &
    \resultsfignew{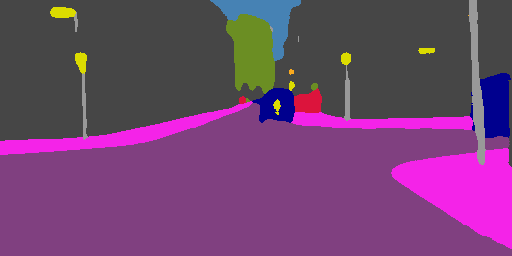} & 
    \resultsfignew{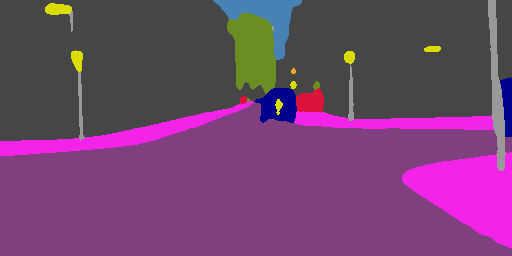} & 
    \resultsfignew{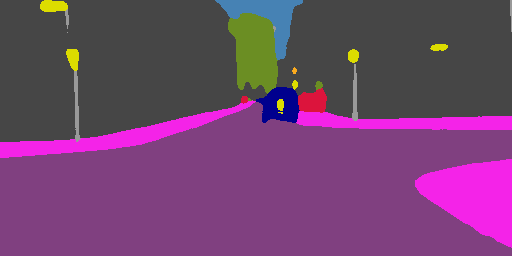} & 
    \resultsfignew{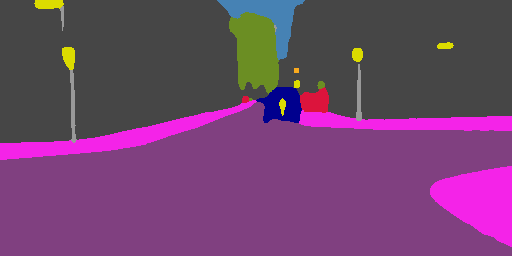}   \\

    \mc{5}{\vspace{-2.0ex}}\\
    &
    \resultsfignew{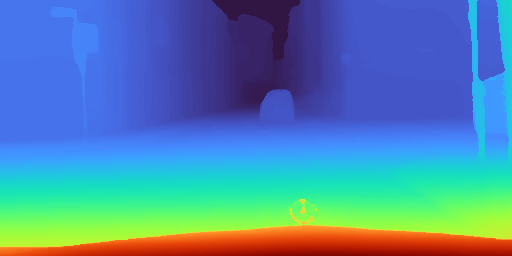} & 
    \resultsfignew{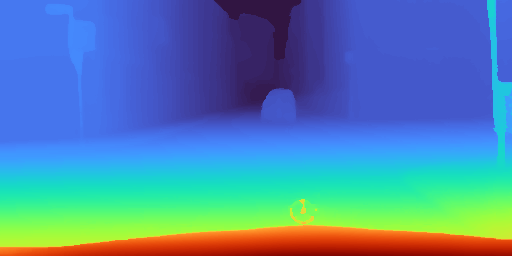} & 
    \resultsfignew{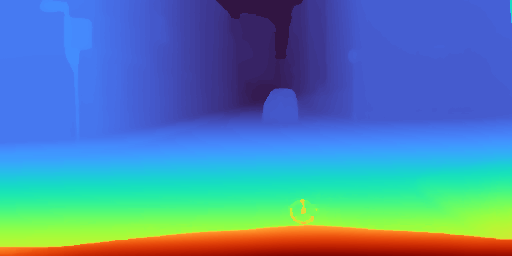} & 
    \resultsfignew{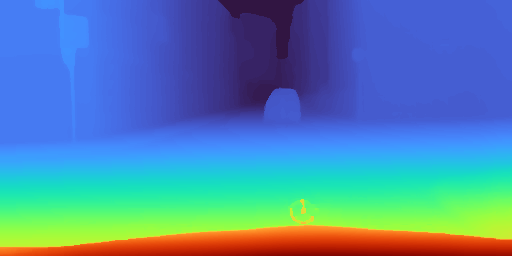}  \\

    \mc{5}{\vspace{-1.5ex}}\\
    &
     $X_{t+27}  (1.62s)$ & $X_{t+30}  (1.8s)$& $X_{t+33}  (1.98s)$ & $X_{t+36}  (2.16s)$ \\

    &
    \resultsfignew{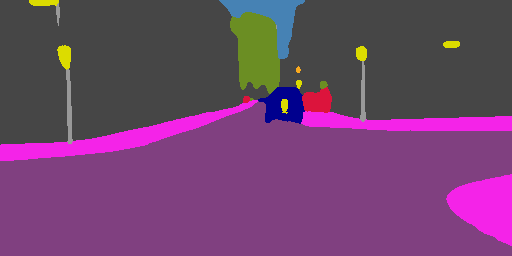} & 
    \resultsfignew{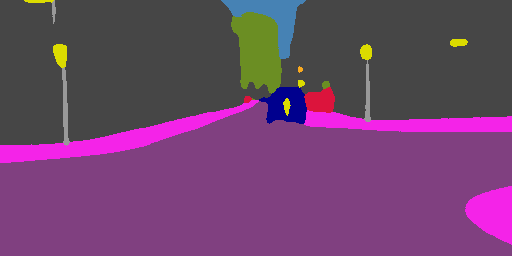} & 
    \resultsfignew{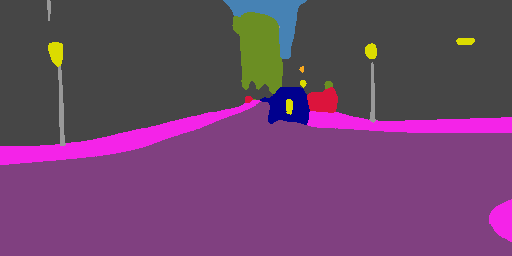} & 
    \resultsfignew{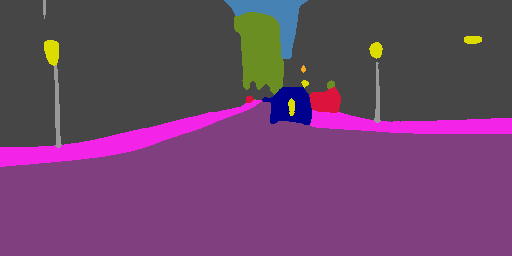}  \\

    \mc{5}{\vspace{-2.0ex}}\\
    
    &
    \resultsfignew{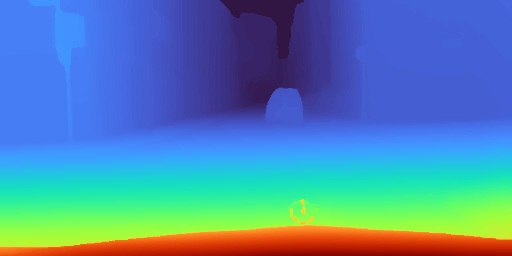} & 
    \resultsfignew{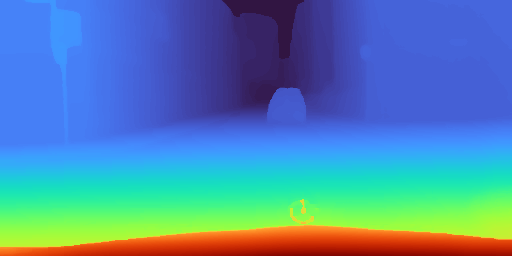} & 
    \resultsfignew{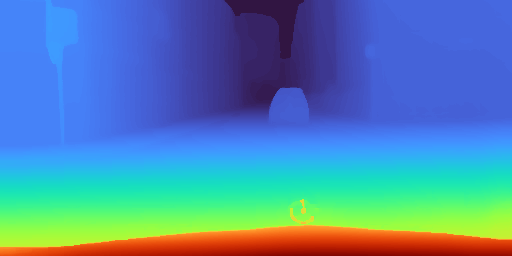} & 
    \resultsfignew{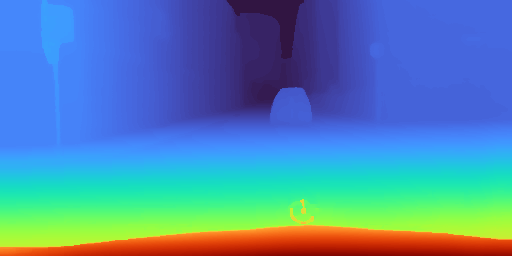}  \\

    \mc{5}{\vspace{-1.5ex}}\\
    &
     $X_{t+39} (2.34s)$ & $X_{t+42} (2.52s)$& $X_{t+45} (2.7s)$ & $X_{t+48} (2.88s)$ \\

     &
    \resultsfignew{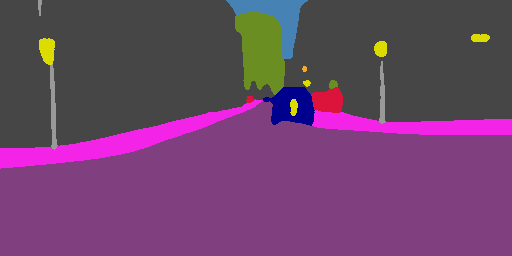} & 
    \resultsfignew{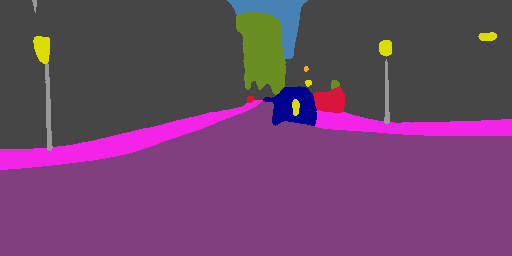} & 
    \resultsfignew{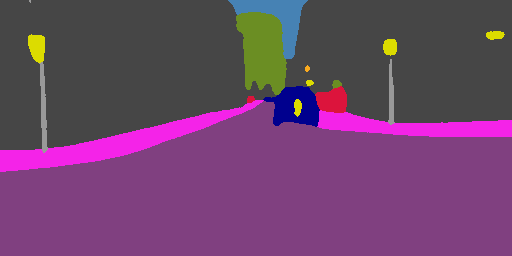} & 
    \resultsfignew{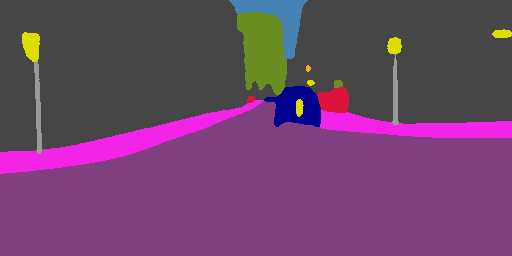}\\

    \mc{5}{\vspace{-2.0ex}}\\
     &
    \resultsfignew{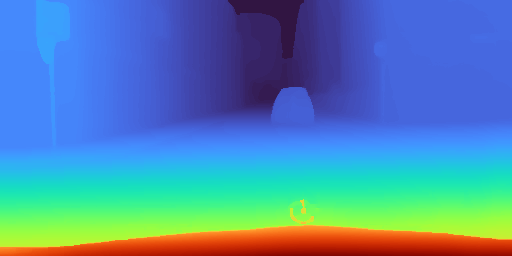} & 
    \resultsfignew{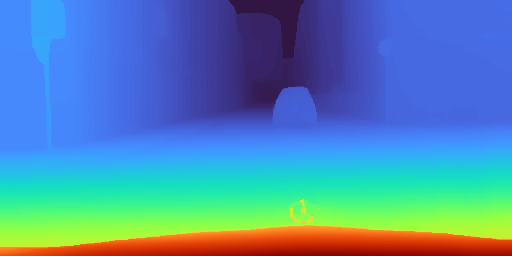} & 
    \resultsfignew{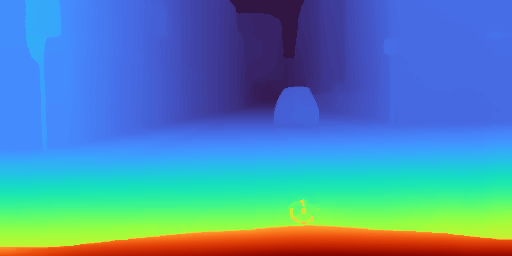} & 
    \resultsfignew{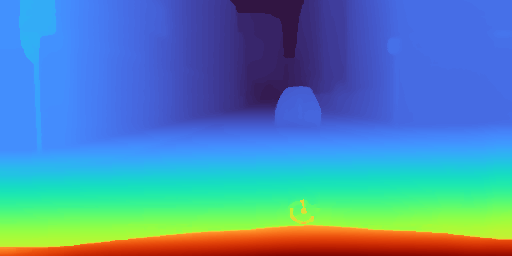}\\
    
\end{tabular}
    };

    \begin{scope}[on background layer]
        \draw[green, thick, rounded corners] 
            ([xshift=1pt, yshift=-2pt]table.north west) 
            rectangle ([xshift=-3pt, yshift=474pt]table.south east);

        \draw[blue, thick, rounded corners] 
            ([xshift=1pt, yshift=-120pt]table.north west) 
            rectangle ([xshift=-3pt, yshift=1pt]table.south east);
    \end{scope}

\end{tikzpicture}
}

%% file: sec/supp_fig_qualitative_rollout_frankfurt_000000_001236.tex
{
\small
\centering
\newcommand{\resultsfignew}[1]{\includegraphics[width=0.2\textwidth,valign=c]{#1}}
\setlength{\tabcolsep}{1pt}

\pgfsetlayers{background,main}

\begin{tikzpicture}
    \node (table) {
\begin{tabular}{@{}ccccc@{}}
    
    \multirow{3}{*}{\raisebox{-84pt}{\rotatebox{90}{Context Frames}}} &
    $X_{t-9}$ & $X_{t-6}$& $X_{t-3}$ & $X_{t}$ \\

    &
    \resultsfignew{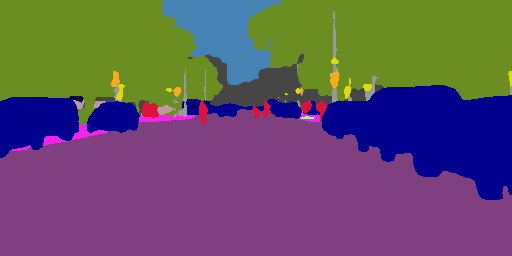} & 
    \resultsfignew{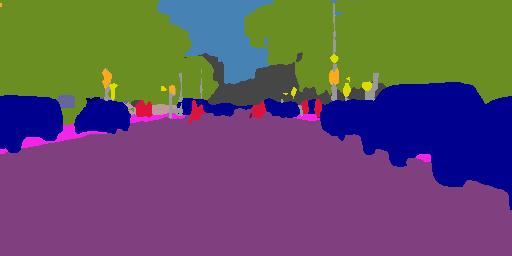} & 
    \resultsfignew{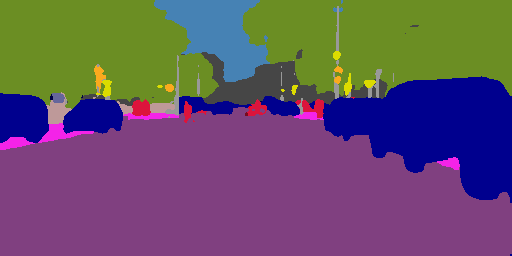} & 
    \resultsfignew{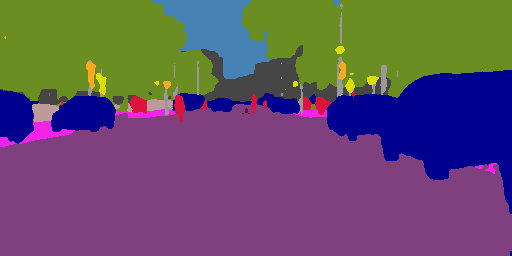}  \\

    \mc{5}{\vspace{-2.0ex}}\\

     &
    \resultsfignew{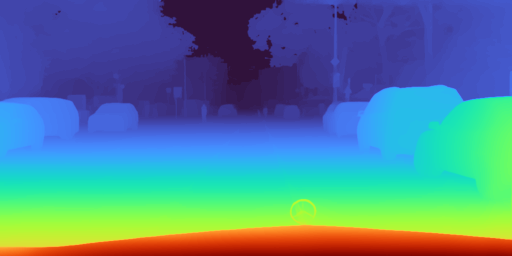} & 
    \resultsfignew{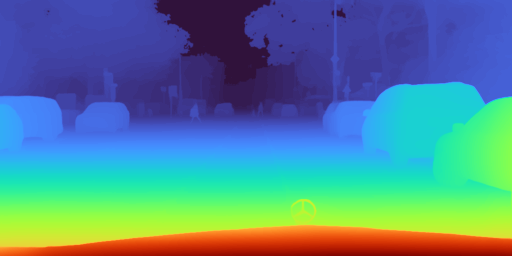} & 
    \resultsfignew{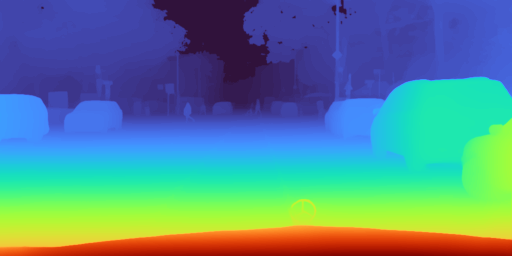} & 
    \resultsfignew{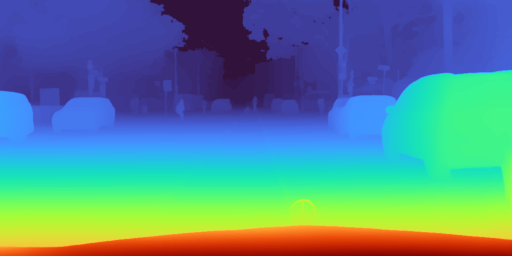}  \\

    \mc{5}{\vspace{-1.5ex}}\\
    
    &
    $X_{t+3} (0.18s)$ & $X_{t+6} (0.36s)$& $X_{t+9} (0.54s)$ & $X_{t+12} (0.72s)$ \\

    &
    \resultsfignew{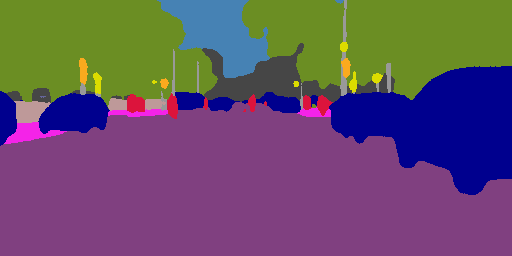} & 
    \resultsfignew{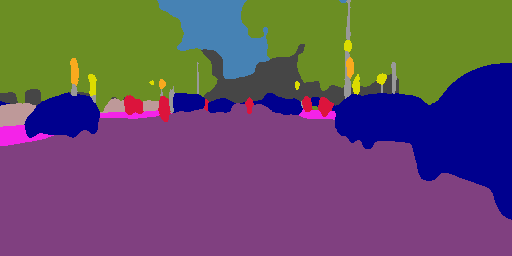} & 
    \resultsfignew{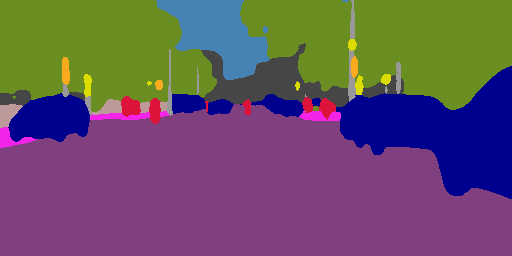} & 
    \resultsfignew{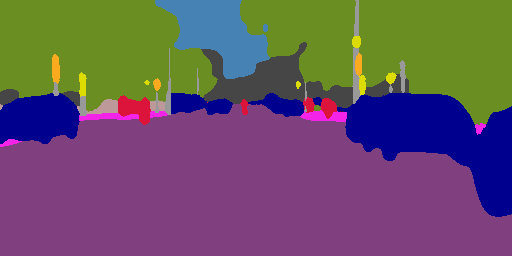}  \\

    \mc{5}{\vspace{-2.0ex}}\\
    &
    \resultsfignew{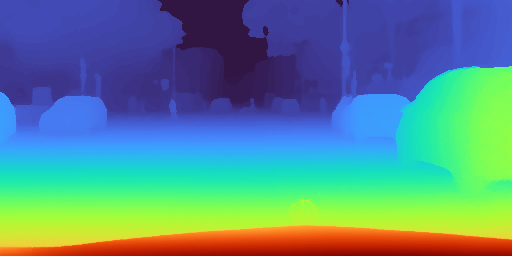} & 
    \resultsfignew{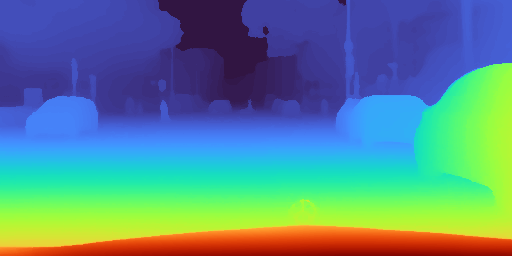} & 
    \resultsfignew{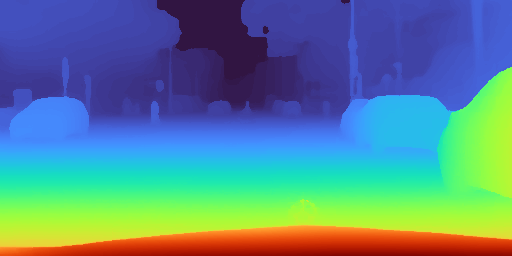} & 
    \resultsfignew{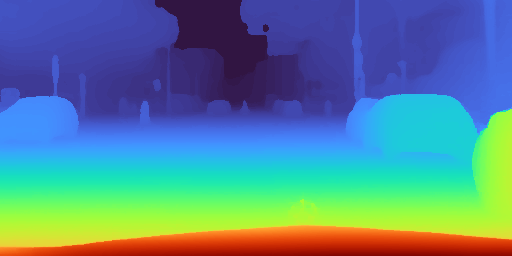}   \\

    \mc{5}{\vspace{-1.5ex}}\\
    
    \multirow{3}{*}{\raisebox{-147pt}{\rotatebox{90}{Predicted Frames}}} &
    $X_{t+15} (0.9s)$ & $X_{t+18}(1.08s)$& $X_{t+21} (1.26s)$ & $X_{t+24} (1.44s)$ \\
       &
    \resultsfignew{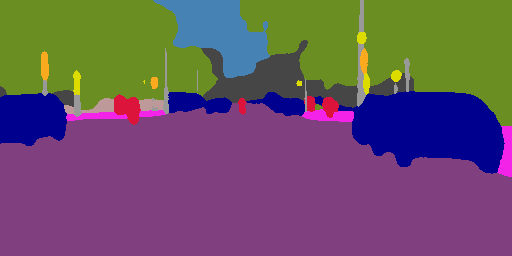} & 
    \resultsfignew{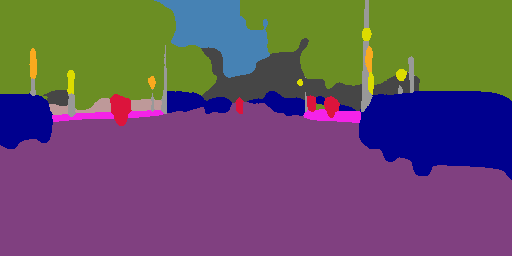} & 
    \resultsfignew{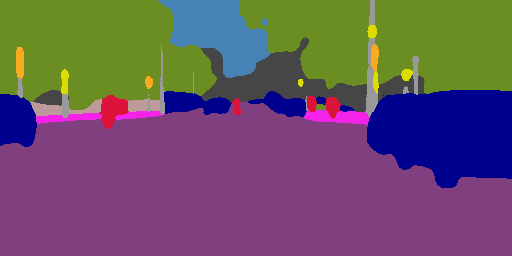} & 
    \resultsfignew{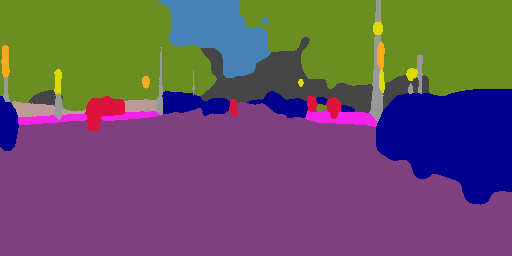}   \\

    \mc{5}{\vspace{-2.0ex}}\\
    &
    \resultsfignew{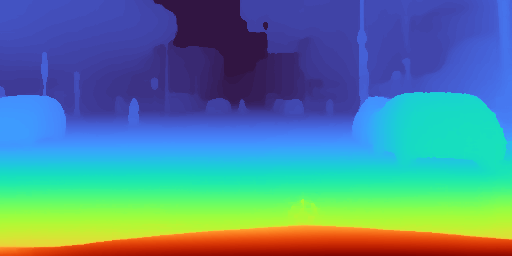} & 
    \resultsfignew{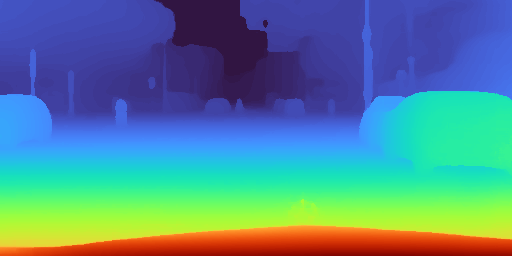} & 
    \resultsfignew{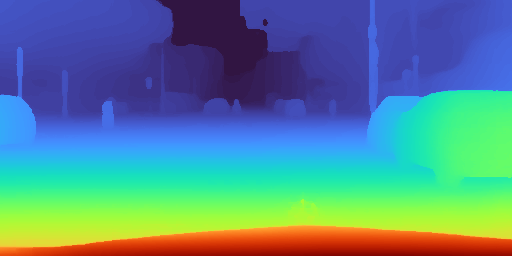} & 
    \resultsfignew{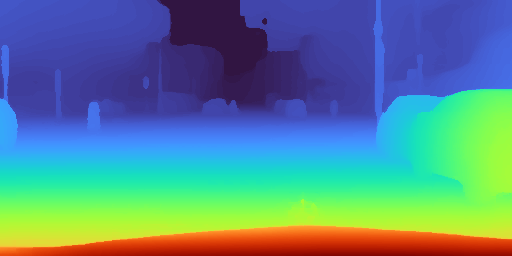}  \\

    \mc{5}{\vspace{-1.5ex}}\\
    &
     $X_{t+27}  (1.62s)$ & $X_{t+30}  (1.8s)$& $X_{t+33}  (1.98s)$ & $X_{t+36}  (2.16s)$ \\

    &
    \resultsfignew{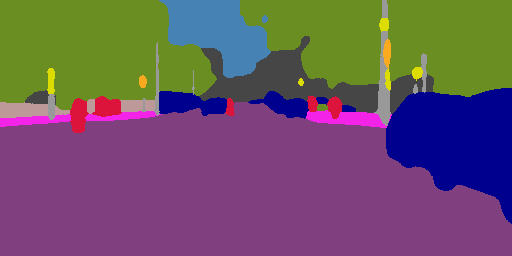} & 
    \resultsfignew{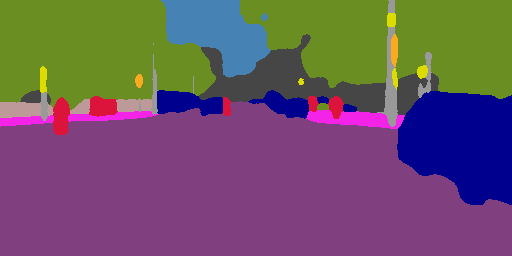} & 
    \resultsfignew{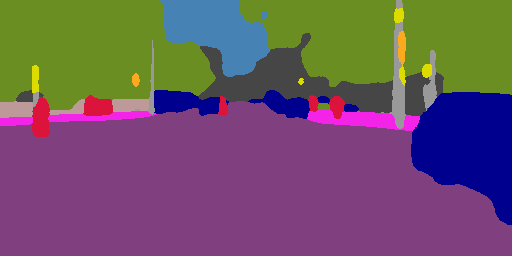} & 
    \resultsfignew{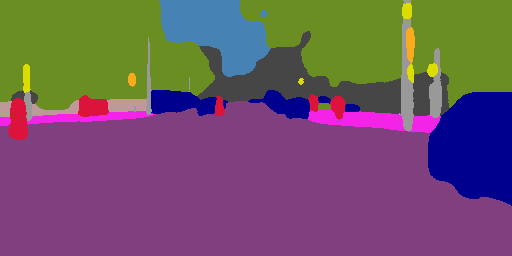}  \\

    \mc{5}{\vspace{-2.0ex}}\\
    
    &
    \resultsfignew{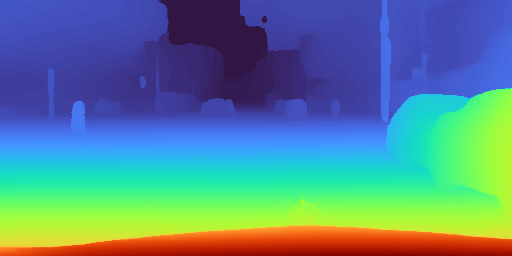} & 
    \resultsfignew{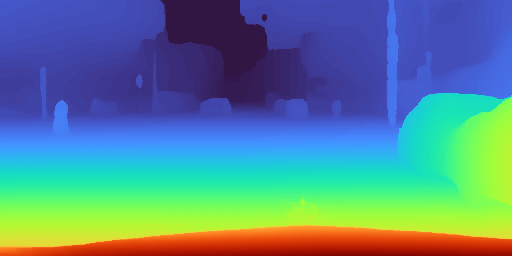} & 
    \resultsfignew{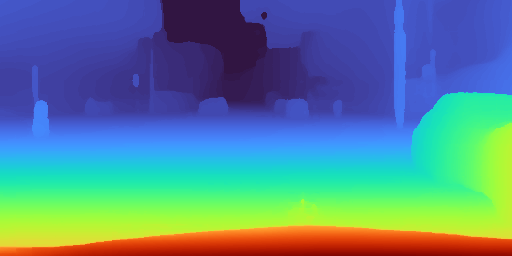} & 
    \resultsfignew{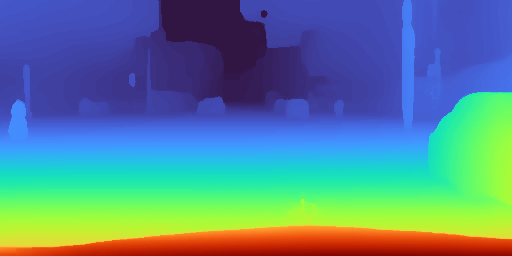}  \\

    \mc{5}{\vspace{-1.5ex}}\\
    &
     $X_{t+39} (2.34s)$ & $X_{t+42} (2.52s)$& $X_{t+45} (2.7s)$ & $X_{t+48} (2.88s)$ \\

     &
    \resultsfignew{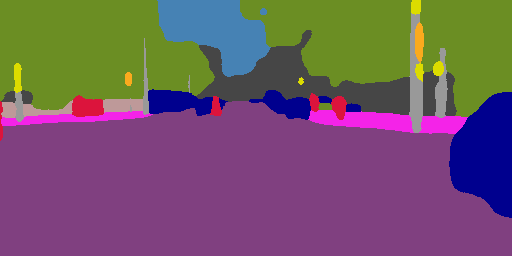} & 
    \resultsfignew{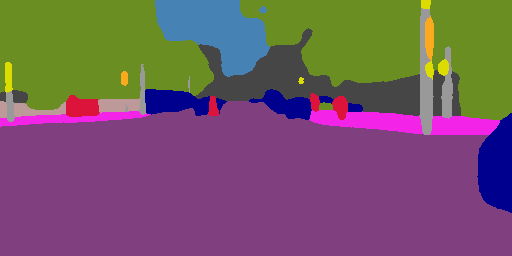} & 
    \resultsfignew{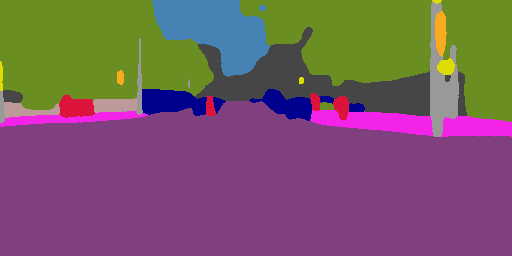} & 
    \resultsfignew{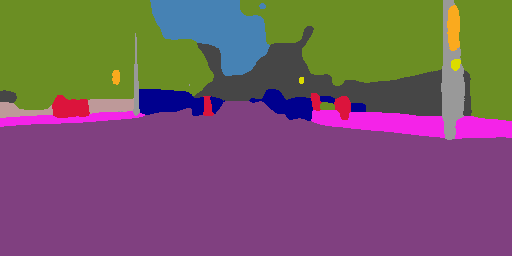}\\

    \mc{5}{\vspace{-2.0ex}}\\
     &
    \resultsfignew{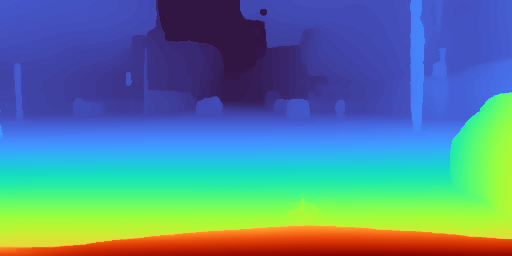} & 
    \resultsfignew{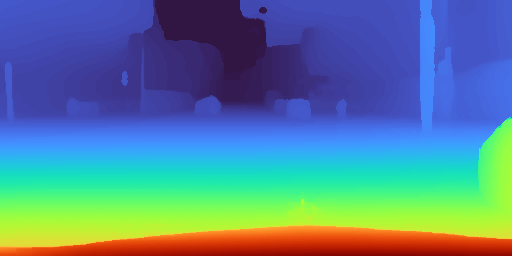} & 
    \resultsfignew{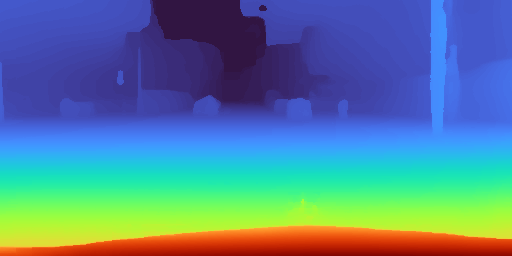} & 
    \resultsfignew{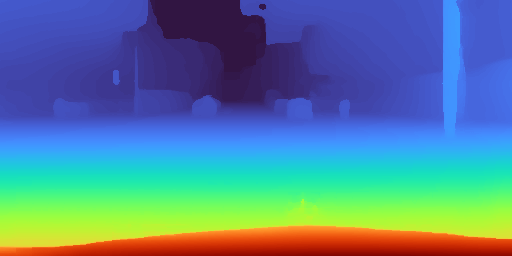}\\
    
\end{tabular}
    };

    \begin{scope}[on background layer]
        \draw[green, thick, rounded corners] 
            ([xshift=1pt, yshift=-2pt]table.north west) 
            rectangle ([xshift=-3pt, yshift=474pt]table.south east);

        \draw[blue, thick, rounded corners] 
            ([xshift=1pt, yshift=-120pt]table.north west) 
            rectangle ([xshift=-3pt, yshift=1pt]table.south east);
    \end{scope}

\end{tikzpicture}
}

%% file: sec/supp_fig_qualitative_rollout_lindau_000037_000019.tex
{
\small
\centering
\newcommand{\resultsfignew}[1]{\includegraphics[width=0.2\textwidth,valign=c]{#1}}
\setlength{\tabcolsep}{1pt}

\pgfsetlayers{background,main}

\begin{tikzpicture}
    \node (table) {
\begin{tabular}{@{}ccccc@{}}
    
    \multirow{3}{*}{\raisebox{-84pt}{\rotatebox{90}{Context Frames}}} &
    $X_{t-9}$ & $X_{t-6}$& $X_{t-3}$ & $X_{t}$ \\

    &
    \resultsfignew{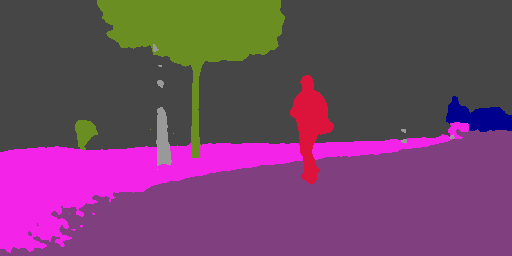} & 
    \resultsfignew{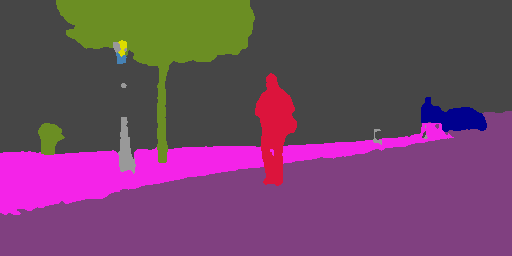} & 
    \resultsfignew{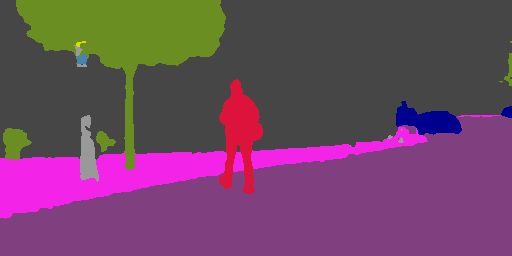} & 
    \resultsfignew{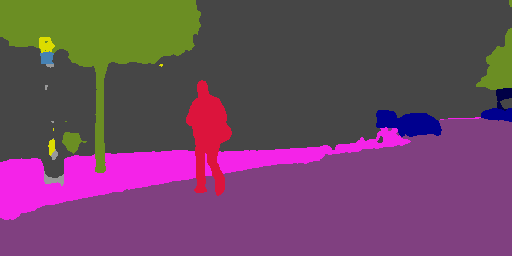}  \\

    \mc{5}{\vspace{-2.0ex}}\\

     &
    \resultsfignew{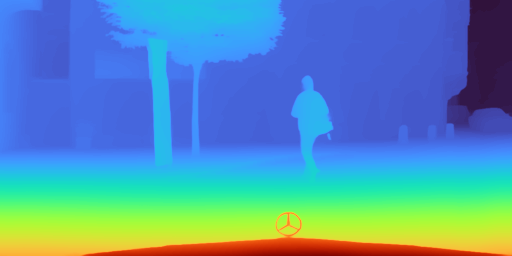} & 
    \resultsfignew{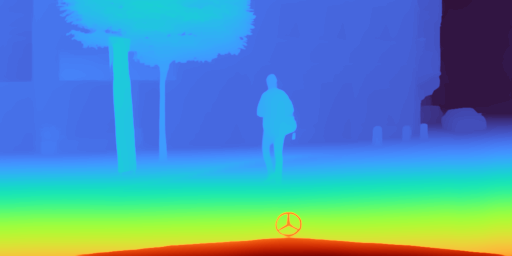} & 
    \resultsfignew{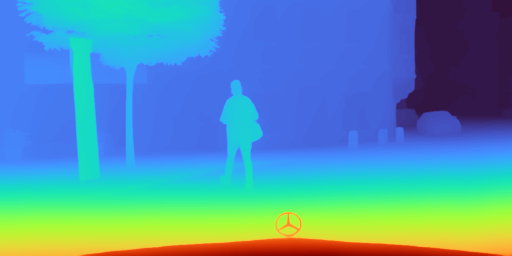} & 
    \resultsfignew{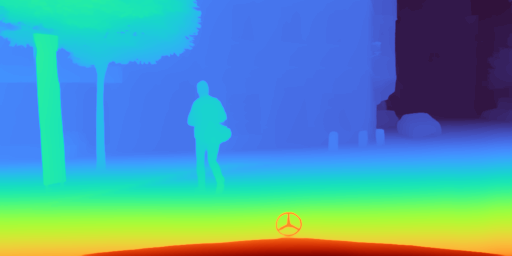}  \\

    \mc{5}{\vspace{-1.5ex}}\\
    
    &
    $X_{t+3} (0.18s)$ & $X_{t+6} (0.36s)$& $X_{t+9} (0.54s)$ & $X_{t+12} (0.72s)$ \\

    &
    \resultsfignew{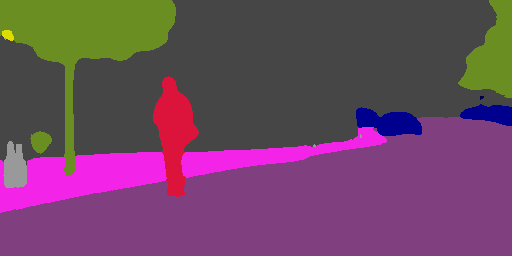} & 
    \resultsfignew{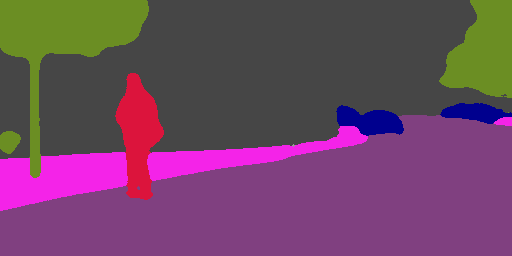} & 
    \resultsfignew{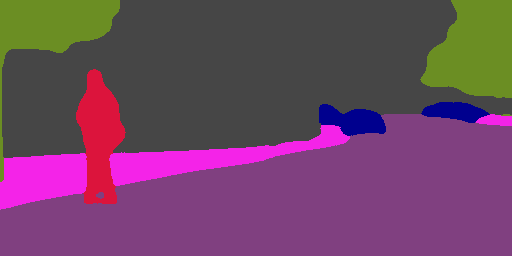} & 
    \resultsfignew{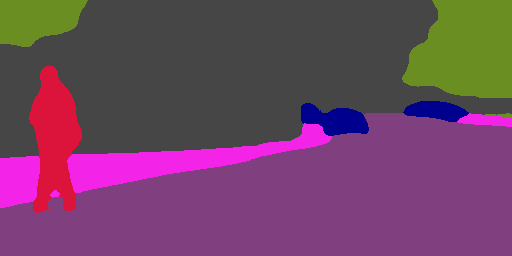}  \\

    \mc{5}{\vspace{-2.0ex}}\\
    &
    \resultsfignew{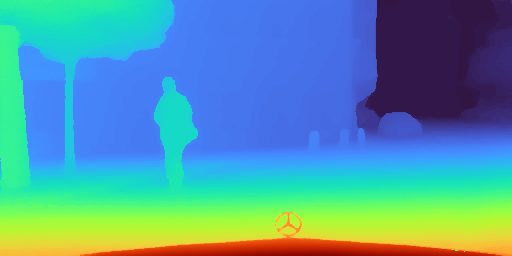} & 
    \resultsfignew{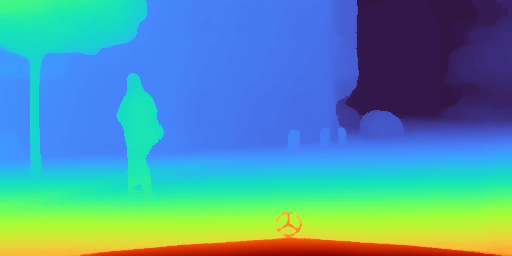} & 
    \resultsfignew{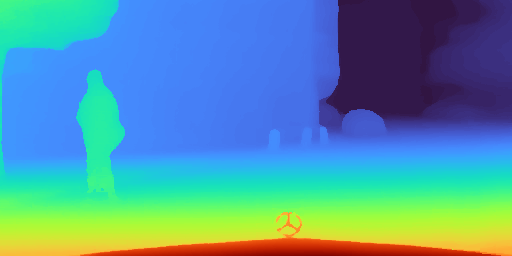} & 
    \resultsfignew{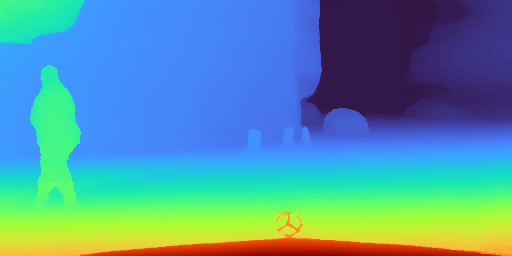}   \\

    \mc{5}{\vspace{-1.5ex}}\\
    
    \multirow{3}{*}{\raisebox{-147pt}{\rotatebox{90}{Predicted Frames}}} &
    $X_{t+15} (0.9s)$ & $X_{t+18}(1.08s)$& $X_{t+21} (1.26s)$ & $X_{t+24} (1.44s)$ \\
       &
    \resultsfignew{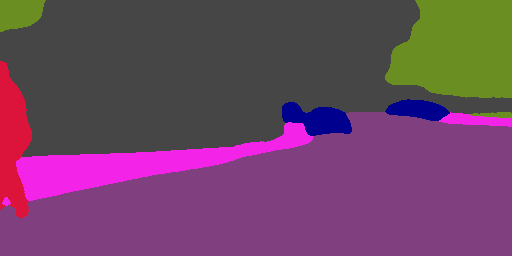} & 
    \resultsfignew{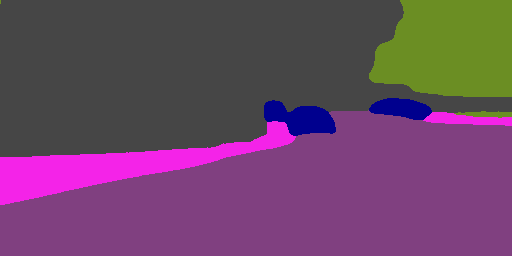} & 
    \resultsfignew{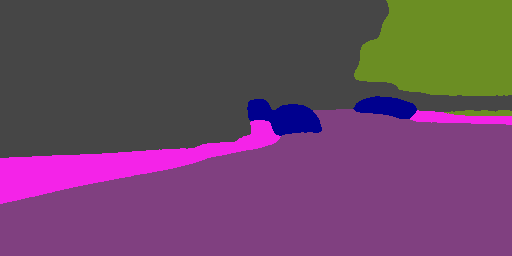} & 
    \resultsfignew{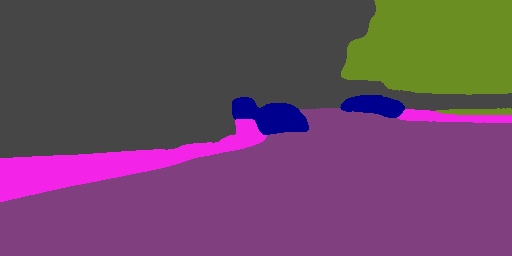}   \\

    \mc{5}{\vspace{-2.0ex}}\\
    &
    \resultsfignew{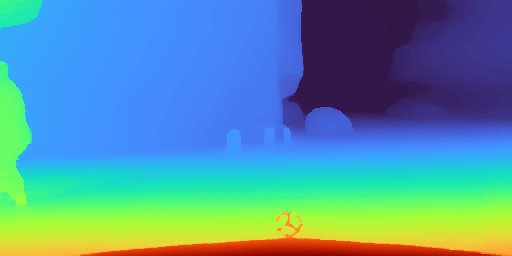} & 
    \resultsfignew{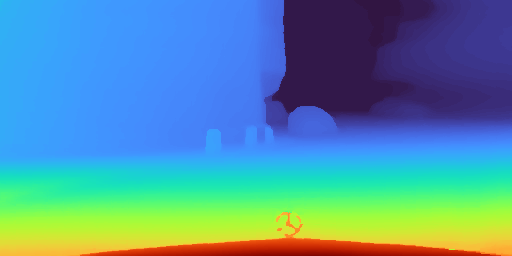} & 
    \resultsfignew{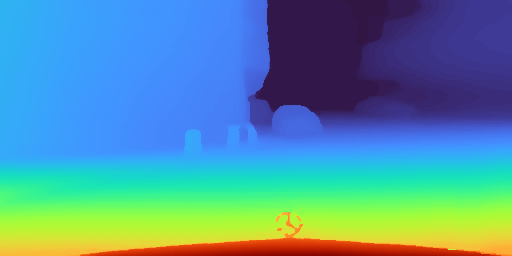} & 
    \resultsfignew{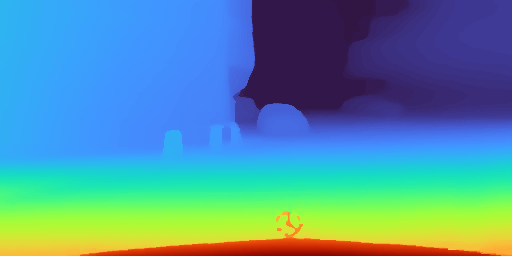}  \\

    \mc{5}{\vspace{-1.5ex}}\\
    &
     $X_{t+27}  (1.62s)$ & $X_{t+30}  (1.8s)$& $X_{t+33}  (1.98s)$ & $X_{t+36}  (2.16s)$ \\

    &
    \resultsfignew{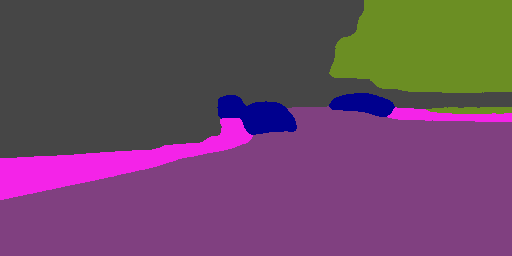} & 
    \resultsfignew{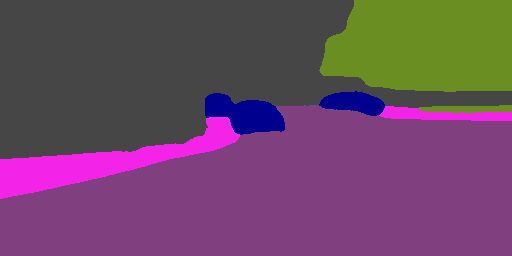} & 
    \resultsfignew{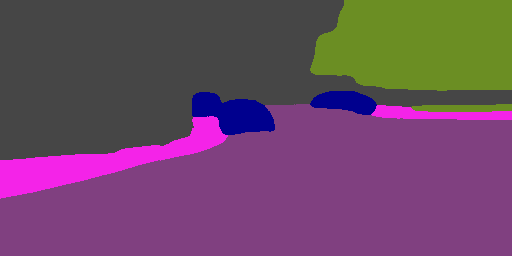} & 
    \resultsfignew{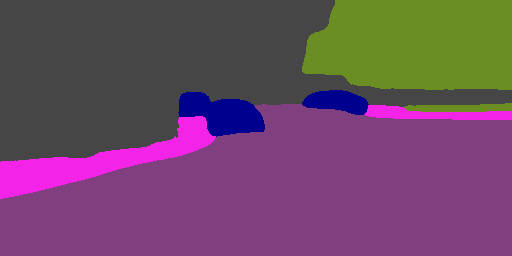}  \\

    \mc{5}{\vspace{-2.0ex}}\\
    
    &
    \resultsfignew{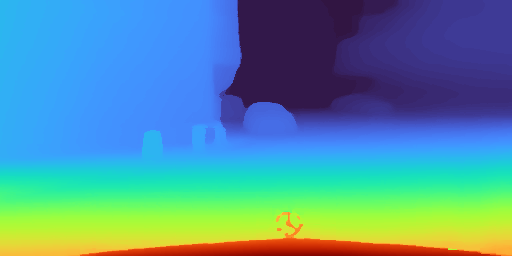} & 
    \resultsfignew{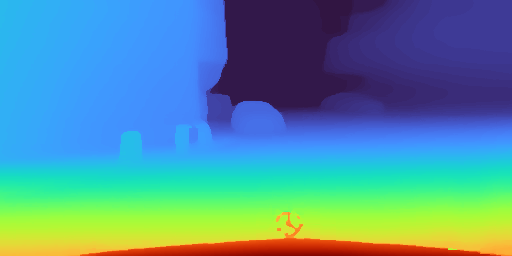} & 
    \resultsfignew{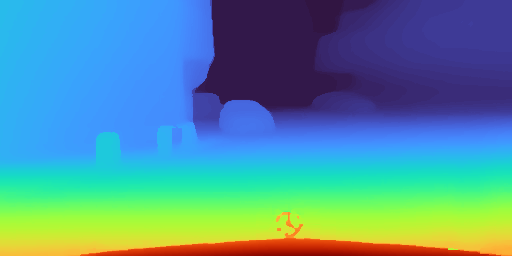} & 
    \resultsfignew{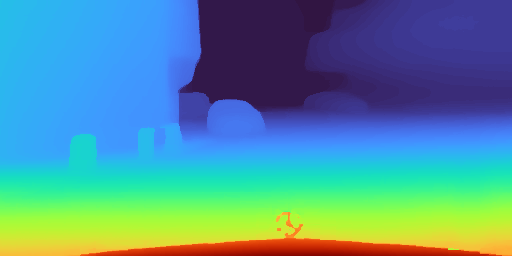}  \\

    \mc{5}{\vspace{-1.5ex}}\\
    &
     $X_{t+39} (2.34s)$ & $X_{t+42} (2.52s)$& $X_{t+45} (2.7s)$ & $X_{t+48} (2.88s)$ \\

     &
    \resultsfignew{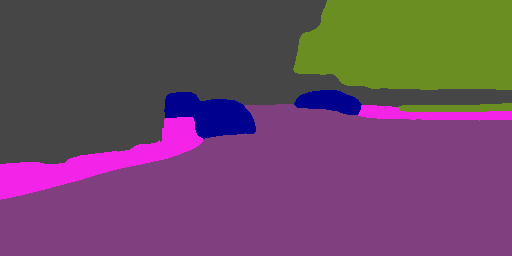} & 
    \resultsfignew{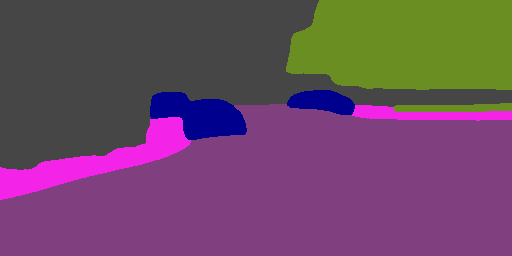} & 
    \resultsfignew{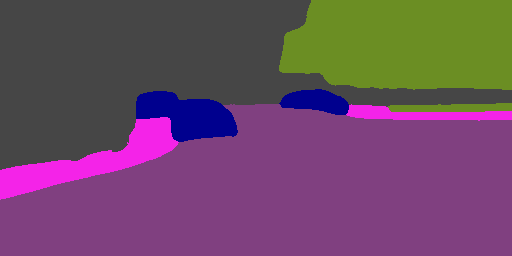} & 
    \resultsfignew{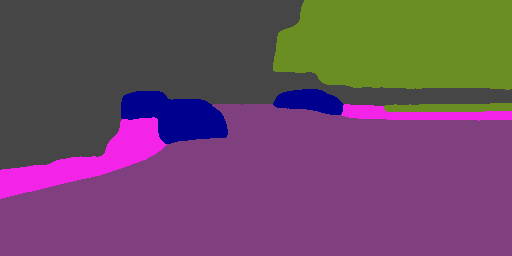}\\

    \mc{5}{\vspace{-2.0ex}}\\
     &
    \resultsfignew{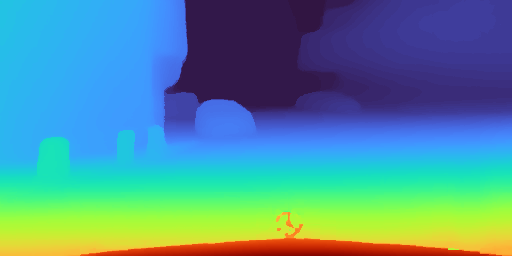} & 
    \resultsfignew{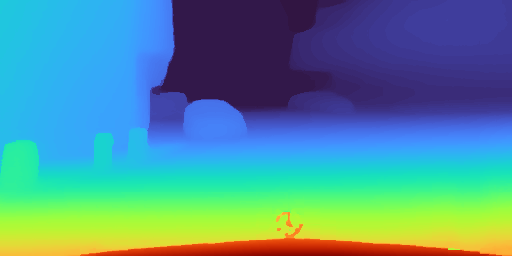} & 
    \resultsfignew{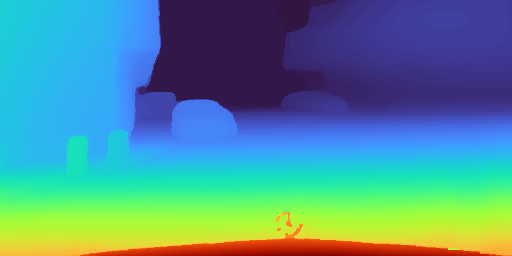} & 
    \resultsfignew{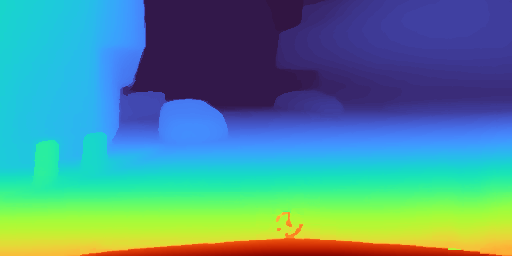}\\
    
\end{tabular}
    };

    \begin{scope}[on background layer]
        \draw[green, thick, rounded corners] 
            ([xshift=1pt, yshift=-2pt]table.north west) 
            rectangle ([xshift=-3pt, yshift=474pt]table.south east);

        \draw[blue, thick, rounded corners] 
            ([xshift=1pt, yshift=-120pt]table.north west) 
            rectangle ([xshift=-3pt, yshift=1pt]table.south east);
    \end{scope}

\end{tikzpicture}
}

%% file: sec/supp_fig_qualitative_rollout_munster_000160_000019.tex
{
\small
\centering
\newcommand{\resultsfignew}[1]{\includegraphics[width=0.2\textwidth,valign=c]{#1}}
\setlength{\tabcolsep}{1pt}

\pgfsetlayers{background,main}

\begin{tikzpicture}
    \node (table) {
\begin{tabular}{@{}ccccc@{}}
    
    \multirow{3}{*}{\raisebox{-84pt}{\rotatebox{90}{Context Frames}}} &
    $X_{t-9}$ & $X_{t-6}$& $X_{t-3}$ & $X_{t}$ \\

    &
    \resultsfignew{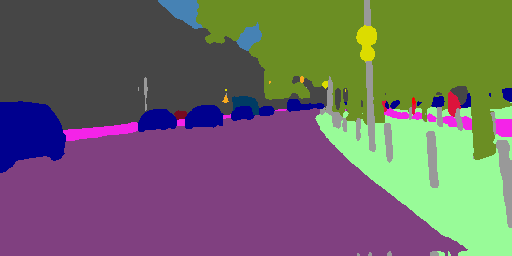} & 
    \resultsfignew{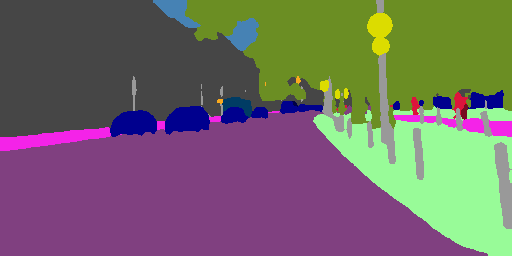} & 
    \resultsfignew{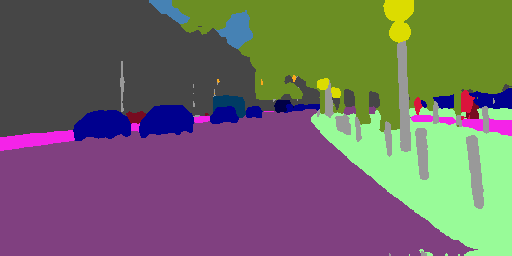} & 
    \resultsfignew{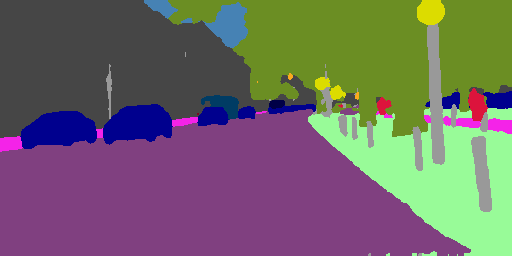}  \\

    \mc{5}{\vspace{-2.0ex}}\\

     &
    \resultsfignew{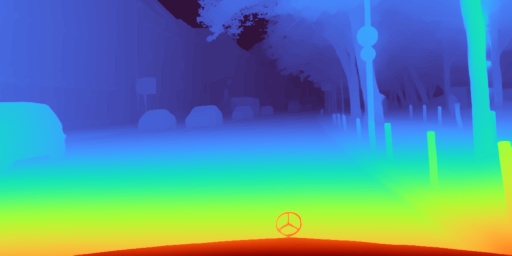} & 
    \resultsfignew{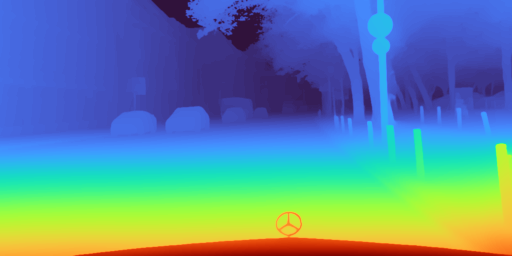} & 
    \resultsfignew{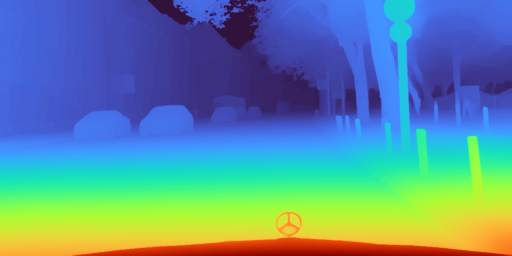} & 
    \resultsfignew{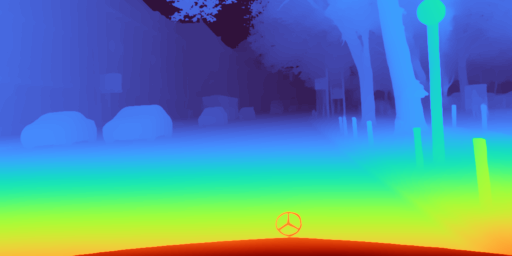}  \\

    \mc{5}{\vspace{-1.5ex}}\\
    
    &
    $X_{t+3} (0.18s)$ & $X_{t+6} (0.36s)$& $X_{t+9} (0.54s)$ & $X_{t+12} (0.72s)$ \\

    &
    \resultsfignew{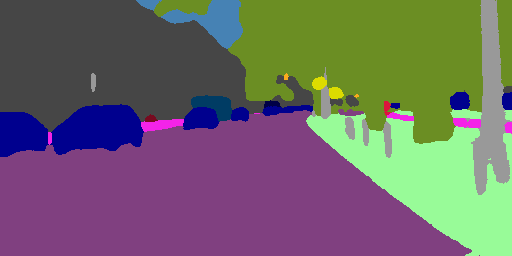} & 
    \resultsfignew{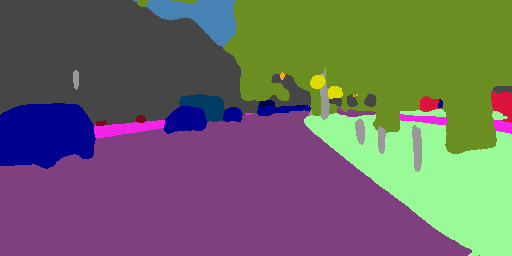} & 
    \resultsfignew{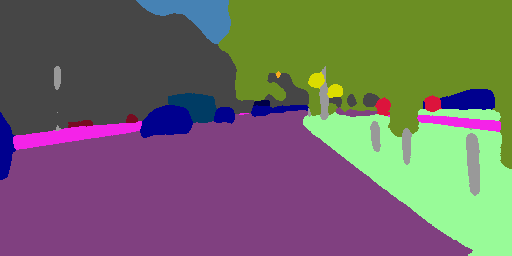} & 
    \resultsfignew{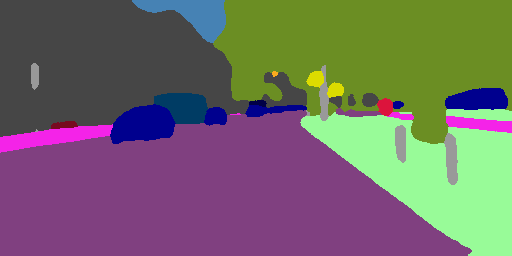}  \\

    \mc{5}{\vspace{-2.0ex}}\\
    &
    \resultsfignew{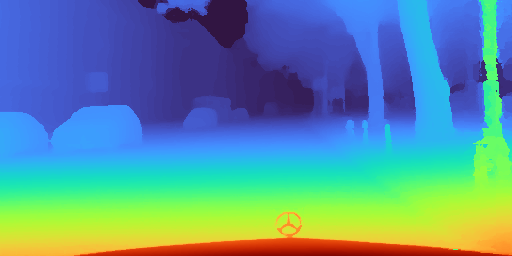} & 
    \resultsfignew{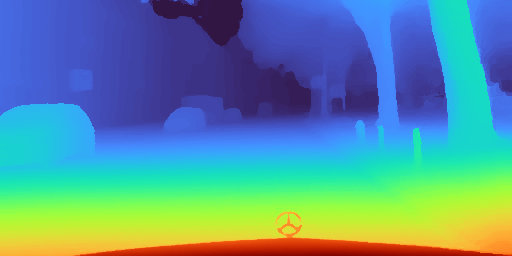} & 
    \resultsfignew{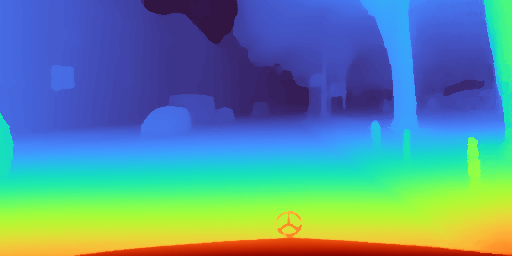} & 
    \resultsfignew{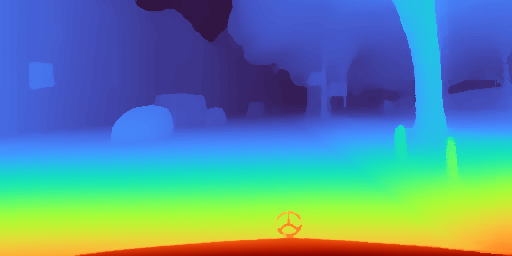}   \\

    \mc{5}{\vspace{-1.5ex}}\\
    
    \multirow{3}{*}{\raisebox{-147pt}{\rotatebox{90}{Predicted Frames}}} &
    $X_{t+15} (0.9s)$ & $X_{t+18}(1.08s)$& $X_{t+21} (1.26s)$ & $X_{t+24} (1.44s)$ \\
       &
    \resultsfignew{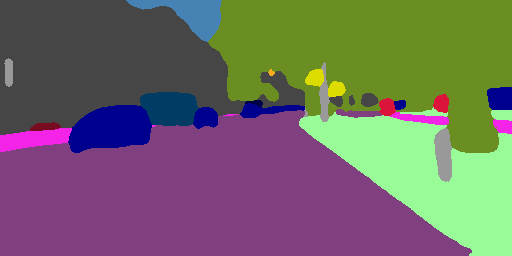} & 
    \resultsfignew{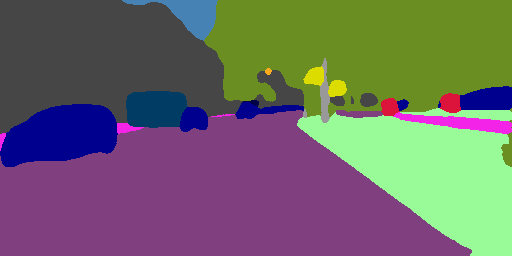} & 
    \resultsfignew{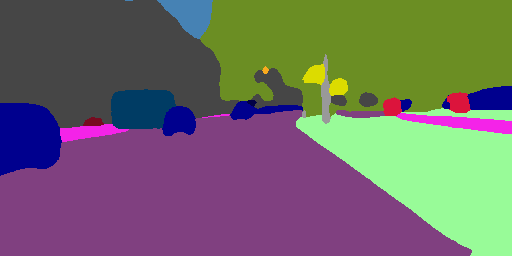} & 
    \resultsfignew{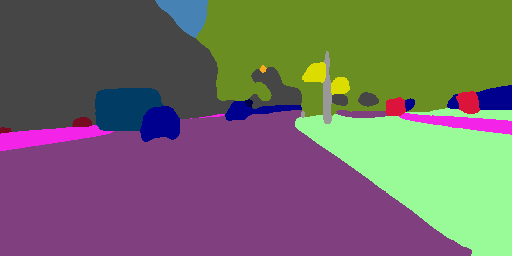}   \\

    \mc{5}{\vspace{-2.0ex}}\\
    &
    \resultsfignew{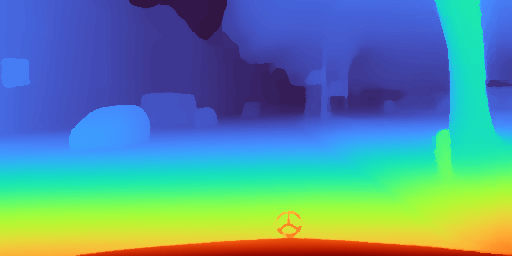} & 
    \resultsfignew{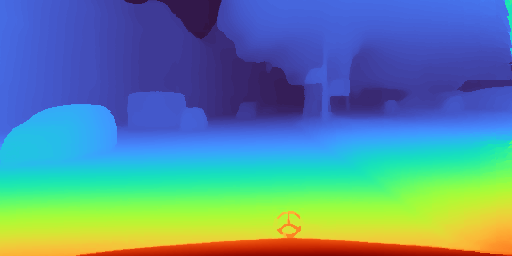} & 
    \resultsfignew{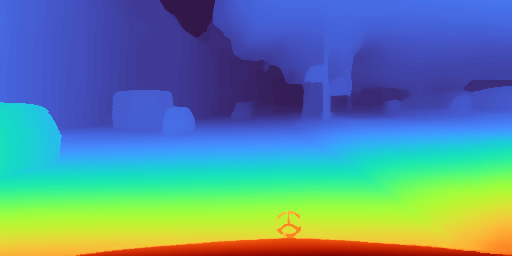} & 
    \resultsfignew{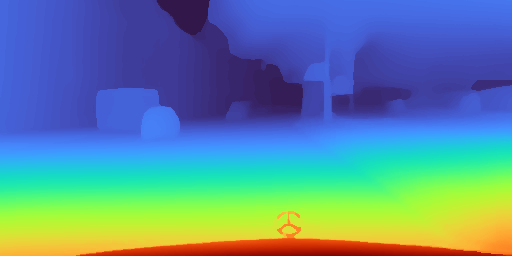}  \\

    \mc{5}{\vspace{-1.5ex}}\\
    &
     $X_{t+27}  (1.62s)$ & $X_{t+30}  (1.8s)$& $X_{t+33}  (1.98s)$ & $X_{t+36}  (2.16s)$ \\

    &
    \resultsfignew{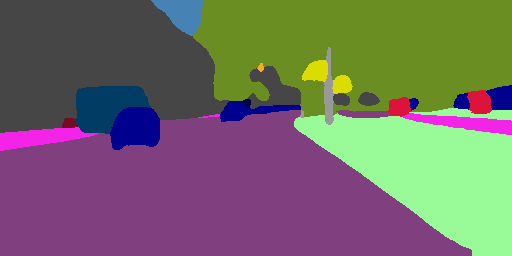} & 
    \resultsfignew{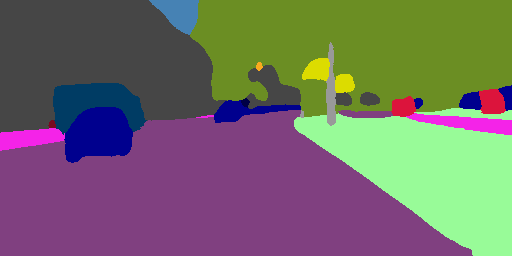} & 
    \resultsfignew{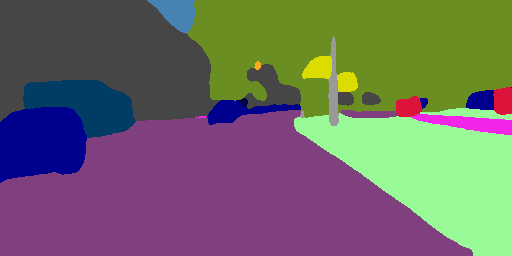} & 
    \resultsfignew{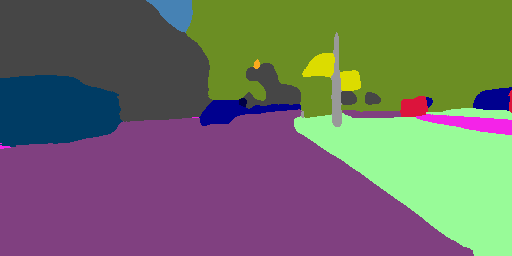}  \\

    \mc{5}{\vspace{-2.0ex}}\\
    
    &
    \resultsfignew{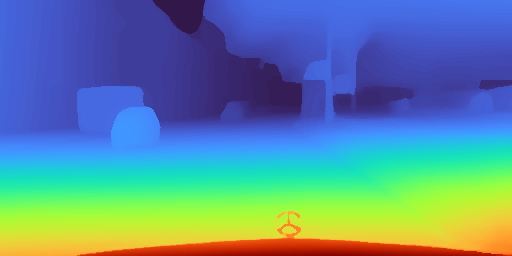} & 
    \resultsfignew{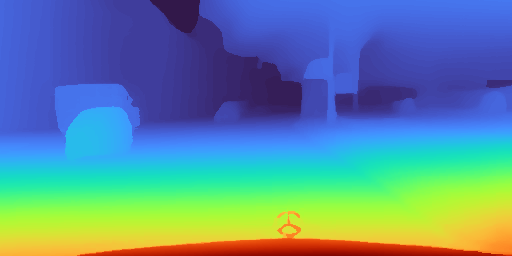} & 
    \resultsfignew{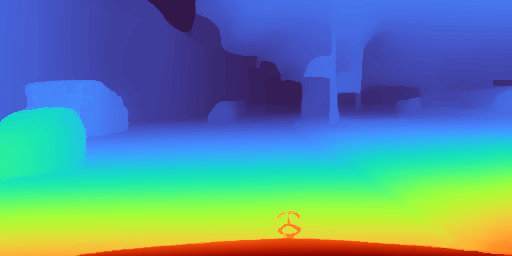} & 
    \resultsfignew{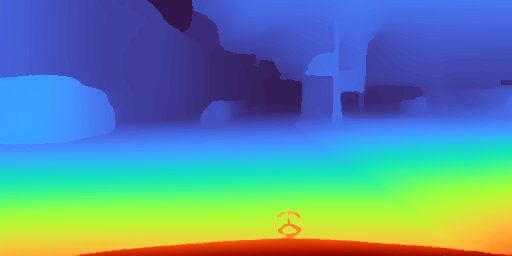}  \\

    \mc{5}{\vspace{-1.5ex}}\\
    &
     $X_{t+39} (2.34s)$ & $X_{t+42} (2.52s)$& $X_{t+45} (2.7s)$ & $X_{t+48} (2.88s)$ \\

     &
    \resultsfignew{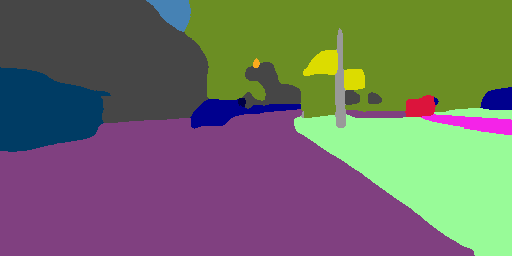} & 
    \resultsfignew{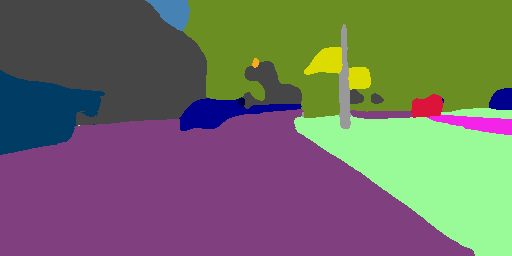} & 
    \resultsfignew{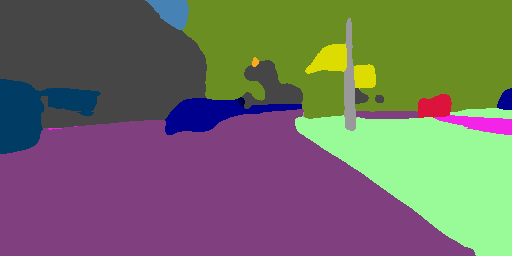} & 
    \resultsfignew{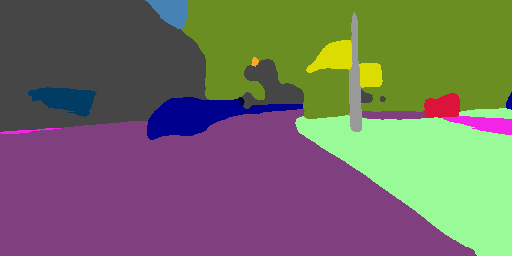}\\

    \mc{5}{\vspace{-2.0ex}}\\
     &
    \resultsfignew{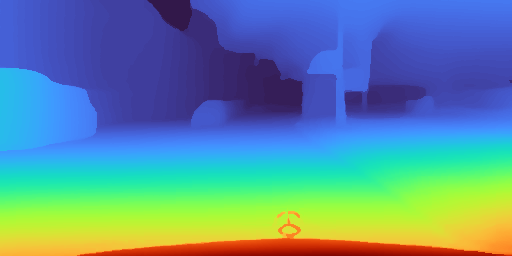} & 
    \resultsfignew{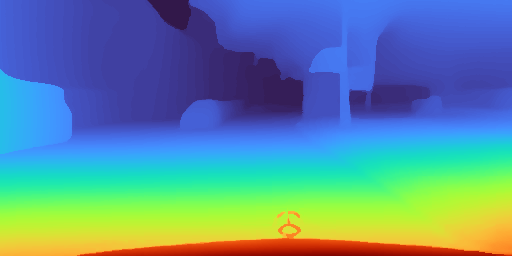} & 
    \resultsfignew{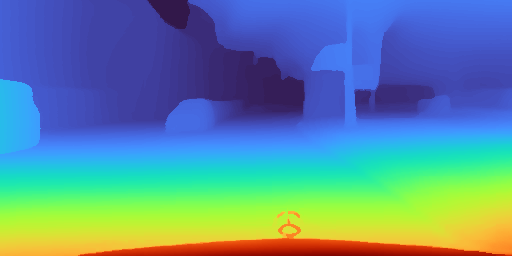} & 
    \resultsfignew{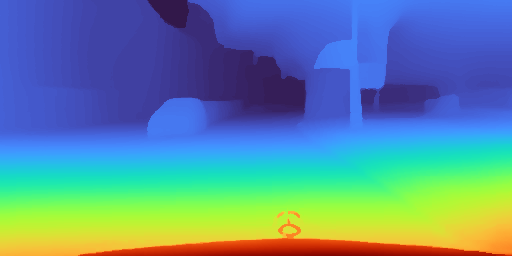}\\
    
\end{tabular}
    };

    \begin{scope}[on background layer]
        \draw[green, thick, rounded corners] 
            ([xshift=1pt, yshift=-2pt]table.north west) 
            rectangle ([xshift=-3pt, yshift=474pt]table.south east);

        \draw[blue, thick, rounded corners] 
            ([xshift=1pt, yshift=-120pt]table.north west) 
            rectangle ([xshift=-3pt, yshift=1pt]table.south east);
    \end{scope}

\end{tikzpicture}
}